\documentclass{article}

\usepackage{iclr2027_conference,times}
\usepackage{wrapfig}

\usepackage{microtype}
\usepackage{graphicx}
\usepackage{subcaption}
\usepackage{booktabs}
\usepackage{placeins}
\usepackage{hyperref}
\usepackage{amsmath}
\usepackage{amssymb}
\usepackage{mathtools}
\usepackage{amsthm}
\usepackage[capitalize,noabbrev]{cleveref}
\usepackage{tikz}
\usepackage{enumitem}
\usetikzlibrary{
    positioning,
    calc,
    backgrounds,
    fit,
    arrows.meta,
    decorations.pathreplacing
}

\usepackage{tikz-3dplot,cancel}

\usepackage{multirow}
\usepackage{bm}

\theoremstyle{plain}

\theoremstyle{definition}

\theoremstyle{remark}

\title{Surrogate Fidelity: When Can Open LLMs\\Explain Closed Ones?}

\author{Philippe Chlenski, Zachariah Carmichael, Ayush Warikoo, Chia-Tse Shao, Yingxiao Ye, \\
\textbf{Aobo Yang, Vivek Miglani, Nehal Bandi} \\
Meta}

\iclrfinalcopy
\begin{document}

\newcommand{\D}{\ensuremath{\mathcal{D}}}

\newcommand{\lp}[2]{\log P (#1\ |\ #2)}

\newcommand{\logit}{\mathrm{logit}}

\newcommand{\Fpred}{F_\mathrm{pred}}
\newcommand{\Fattr}{F_\mathrm{attr}}

\newcommand{\Frepr}{F_\mathrm{mech}}
\newcommand{\Fattn}[1]{F_\mathrm{attn}^\mathrm{#1}}
\newcommand{\Fmag}{F_\mathrm{mag}}
\newcommand{\Falign}{F_\mathrm{align}}

\newcommand{\Fcross}{F_\mathrm{cross}}

\renewcommand{\r}{r^2}

\maketitle
\lhead{Preprint. Under review.}

\begin{abstract}
    Mechanistic interpretability (MI) requires full access to model internals, yet the APIs for most widely deployed language models at best expose log-probabilities over output tokens.
    This creates a surrogate problem: when do measurements made on open models allow us to make claims about a closed model?
    We evaluate \textit{surrogate fidelity} at the prediction, attribution, and mechanistic levels.
    For classification tasks, log-odds provide an API-compatible scalar readout of the model's representation space, and leave-one-out attributions provide insight into model behavior.
    Across eleven models spanning four families (Llama, Qwen, GPT, and Gemini), pooled prediction fidelity substantially exceeds attribution fidelity: models that agree on \emph{what} the answer is often disagree on \emph{why}.
    This gap narrows when conditioning on the ground-truth answer, even reversing when both models answer correctly, suggesting pooled prediction fidelity partly reflects agreement on answers and not fine-grained confidence.
    We document an \textit{access--validity inversion:} white-box signals like attention patterns are highly stable across models but only weakly predictive of causal attributions, which black-box input ablations capture by design. 
    Code and results are available at \url{https://github.com/facebookresearch/surrogate}.
\end{abstract}

\begin{figure}[b]
    \centering
    \resizebox{\textwidth}{!}{\definecolor{lv1}{HTML}{584486}
\definecolor{lv2}{HTML}{0091A1}
\definecolor{lv3}{HTML}{BA7C43}

\begin{tikzpicture}[
  node distance=6mm and 20mm,
  box/.style={draw, rounded corners=2pt, minimum height=6mm, 
              minimum width=16mm, align=center, inner sep=2pt},
  fid/.style={draw, rounded corners=2pt, minimum height=6mm,
              minimum width=16mm, align=center, inner sep=2pt,
              fill=white, font=\small\rmfamily},
  arr/.style={->, >=stealth, thick},
  every node/.append style={font=\small\rmfamily},
  shadedbox/.style={
    rounded corners=4pt, 
    inner sep=4pt,
    opacity=1.0
  }
]

\node[minimum width=24mm] (hS) {$M_S$ (surrogate)};
\node[minimum width=24mm, below=12mm of hS] (hT) {$M_T$ (target)};

\node[left=6mm of $(hS)!0.5!(hT)$] (corpus) {$\D = \{x_1, \ldots, x_n\}$};

\node[box, right=6mm of hS] (lS) {$\ell_{M_S}$};
\node[box, below=14mm of lS] (lT) {$\ell_{M_T}$};
\node[fid] at ($(lS)!0.5!(lT)$) (fp) {$F_\text{pred}$};

\node[box, right=10mm of lS] (dS) {$\Delta\ell_{M_S}$};
\node[box, below=14mm of dS] (dT) {$\Delta\ell_{M_T}$};
\node[fid] at ($(dS)!0.5!(dT)$) (fa) {$F_\text{attr}$};

\node[box, right=32mm of dS] (aS) {$\Delta z_{M_S}$};
\node[box, below=14mm of aS] (aT) {$\Delta z_{M_T}$};
\node[fid] at ($(aS)!0.5!(aT)$) (fattn) {$F_\text{mech}$};

\node[fid] at ($(dT)!0.5!(aS)$) (fc) {$F_\text{cross}$};

\draw[arr] (hS) -- (lS);
\draw[arr] (hT) -- (lT);
\draw[arr] (lS) -- (dS);
\draw[arr] (lT) -- (dT);
\draw[arr] (dS) -- (aS);
\draw[arr] (dT) -- (aT);

\draw[arr, densely dashed] (lS) -- (fp);
\draw[arr, densely dashed] (lT) -- (fp);
\draw[arr, densely dashed] (dS) -- (fa);
\draw[arr, densely dashed] (dT) -- (fa);
\draw[arr, densely dashed] (aS) -- (fattn);
\draw[arr, densely dashed] (aT) -- (fattn);

\draw[arr, densely dashed] (aS) -- (fc);
\draw[arr, densely dashed] (dT) -- (fc);

\draw[arr] (corpus) |- (hS);
\draw[arr] (corpus) |- (hT);

\node[above=2mm of lS] (l1) {Prediction};
\node[above=2mm of dS] (l2) {Attribution};
\node[above=2mm of aS] (l3) {Mechanistic};

\begin{pgfonlayer}{background}
  \node[shadedbox, fill=lv1, opacity=0.3, fit=(lS)(lT)(fp)(l1)] {};
  \node[shadedbox, fill=lv2, opacity=0.3, fit=(dS)(dT)(fa)(l2)] {};
  \node[shadedbox, fill=lv3, opacity=0.3, fit=(aS)(aT)(fattn)(l3)] {};
\end{pgfonlayer}

\end{tikzpicture}}
    \vspace{-0.5cm}
    \caption{
        The surrogate fidelity evaluation pipeline. 
        For each input corpus $\mathcal{D}$, we extract three signals from each model: prediction log-odds $\ell_M$, ablation-based attributions $\Delta\ell_M$, and perturbation responses $\Delta z$.
        Prediction fidelity ($F_{\mathrm{pred}}$) and attribution fidelity ($F_{\mathrm{attr}}$) can be computed for any model pair, including closed-source targets accessible only via API. 
        Mechanistic fidelity ($F_{\mathrm{mech}}$) and cross-level fidelity ($F_{\mathrm{cross}}$) additionally require access to model internals.
    }
    \label{fig:eval_arch}
\end{figure}

\section{Introduction}
    The models that most urgently need mechanistic interpretability (MI) are precisely those least amenable to it.
    Most classical MI tools, such as circuit analysis, sparse autoencoders, and activation patching, require full access to model internals.
    Meanwhile, most users of frontier large language models (LLMs) access text completions from behind APIs that, at best, offer log-probabilities for a handful of high-likelihood tokens.
    This creates a bifurcation: MI research is either conducted by frontier labs within the walled gardens of their own models, or by academic groups studying small open models and hoping their findings generalize.
    The broader community of practitioners, safety researchers, and auditors is left to wonder what can be learned about a model we cannot open from studying one we can.

    Whether mechanistic insights transfer across models is a known open problem \citep{sharkey2025openproblemsmechanisticinterpretability}, but it is rarely measured directly.
    If transfer is weak, then open-source MI work may be telling us less than we think about the frontier models we most need to understand. 
    We argue that the field needs an open dialogue about cross-model fidelity, grounded in a shared convention for measurement that is realistic about the distinction between closed and open models.
    
    We define a \textit{surrogate fidelity metric} as a measure of how faithfully a measurement in a surrogate model predicts a measurement in a target model.
    A surrogate fidelity metric should be \textit{appropriately scoped} (dataset- and claim-dependent), \textit{discriminative} between different model pairs, and governed by the \textit{principle of least privilege}, i.e. it should require no more access to the target model than the claim being evaluated demands.
    In particular, at least some surrogate fidelity measurements should be computable using top-$K$ log-probabilities alone: a constraint consistent with the APIs of many frontier model providers. 

    To this end, we propose a hierarchy of increasingly strict tests as a concrete starting point for measuring surrogate fidelity, drawing on established MI intuitions and practices.
    \textit{Prediction fidelity} for outputs;
    \textit{attribution fidelity} for output-level responses to causal interventions (input ablations);
    \textit{mechanistic fidelity} for internals-level responses to the same interventions;
    and \textit{cross-level fidelity} for predicting the target's attributions from the surrogate's internal representations.
    This progression from observational to interventional to mechanistic agreement echoes the causal hierarchy of \citet{pearl2009causality}, grounding each level in a concrete, measurable test of surrogate faithfulness.

\begin{figure*}[t]
    \centering
    \resizebox{\textwidth}{!}{\definecolor{cvec}{HTML}{C62828}
\definecolor{cdz}{HTML}{0091A1}
\definecolor{cplane}{HTML}{E0E0E0}

\begin{tikzpicture}[
  vec/.style={-{Stealth[length=5pt,width=3.5pt]}, line width=1.2pt},
  vecbold/.style={vec, line width=1.6pt},
  vecd/.style={vec, densely dashed},
  proj/.style={densely dotted, line width=0.8pt},
  fade/.style={opacity=0.18},
  lbl/.style={font=\footnotesize},
  plbl/.style={font=\bfseries\small},
  rlbl/.style={font=\small, anchor=north east, align=left, text width=4cm},
  brc/.style={decorate, decoration={brace, mirror, amplitude=3pt}, line width=1pt},
  brclbl/.style={font=\footnotesize, midway, below=3pt},
]

\pgfmathsetmacro{\pA}{30}
\pgfmathsetmacro{\pF}{0.4}
\pgfmathsetmacro{\pCos}{\pF*cos(\pA)}
\pgfmathsetmacro{\pSin}{\pF*sin(\pA)}

\pgfmathsetmacro{\wpx}{2}  \pgfmathsetmacro{\wpy}{2}
\pgfmathsetmacro{\wmx}{-2}  \pgfmathsetmacro{\wmy}{2}
\pgfmathsetmacro{\zx}{3}   \pgfmathsetmacro{\zy}{0}   \pgfmathsetmacro{\zz}{2}
\pgfmathsetmacro{\zpx}{2}  \pgfmathsetmacro{\zpy}{0}  \pgfmathsetmacro{\zpz}{1}

\pgfmathsetmacro{\vvx}{\wpx - \wmx}
\pgfmathsetmacro{\vvy}{\wpy - \wmy}
\pgfmathsetmacro{\dzx}{\zx - \zpx}
\pgfmathsetmacro{\dzy}{\zy - \zpy}
\pgfmathsetmacro{\dzz}{\zz - \zpz}

\pgfmathsetmacro{\vdv}{\vvx*\vvx + \vvy*\vvy}
\pgfmathsetmacro{\szv}{(\zx*\vvx + \zy*\vvy) / \vdv}
\pgfmathsetmacro{\pvzx}{\szv*\vvx}    \pgfmathsetmacro{\pvzy}{\szv*\vvy}
\pgfmathsetmacro{\szpv}{(\zpx*\vvx + \zpy*\vvy) / \vdv}
\pgfmathsetmacro{\pvzpx}{\szpv*\vvx}  \pgfmathsetmacro{\pvzpy}{\szpv*\vvy}
\pgfmathsetmacro{\sdzv}{(\dzx*\vvx + \dzy*\vvy) / \vdv}
\pgfmathsetmacro{\pvdzx}{\sdzv*\vvx}  \pgfmathsetmacro{\pvdzy}{\sdzv*\vvy}

\pgfmathsetmacro{\wpdwp}{\wpx*\wpx + \wpy*\wpy}
\pgfmathsetmacro{\swp}{(\zx*\wpx + \zy*\wpy) / \wpdwp}
\pgfmathsetmacro{\lwpx}{\swp*\wpx}    \pgfmathsetmacro{\lwpy}{\swp*\wpy}
\pgfmathsetmacro{\wmdwm}{\wmx*\wmx + \wmy*\wmy}
\pgfmathsetmacro{\swm}{(\zx*\wmx + \zy*\wmy) / \wmdwm}
\pgfmathsetmacro{\lwmx}{\swm*\wmx}    \pgfmathsetmacro{\lwmy}{\swm*\wmy}

\pgfmathsetmacro{\wpu}{\wpx + \wpy*\pCos}   \pgfmathsetmacro{\wpv}{\wpy*\pSin}
\pgfmathsetmacro{\wmu}{\wmx + \wmy*\pCos}    \pgfmathsetmacro{\wmv}{\wmy*\pSin}
\pgfmathsetmacro{\vu}{\vvx + \vvy*\pCos}     \pgfmathsetmacro{\vv}{\vvy*\pSin}
\pgfmathsetmacro{\zu}{\zx + \zy*\pCos}       \pgfmathsetmacro{\zv}{\zz + \zy*\pSin}
\pgfmathsetmacro{\zpu}{\zpx + \zpy*\pCos}    \pgfmathsetmacro{\zpv}{\zpz + \zpy*\pSin}
\pgfmathsetmacro{\dzu}{\dzx + \dzy*\pCos}    \pgfmathsetmacro{\dzv}{\dzz + \dzy*\pSin}
\pgfmathsetmacro{\pvzu}{\pvzx + \pvzy*\pCos}      \pgfmathsetmacro{\pvzv}{\pvzy*\pSin}
\pgfmathsetmacro{\pvzpu}{\pvzpx + \pvzpy*\pCos}    \pgfmathsetmacro{\pvzpv}{\pvzpy*\pSin}
\pgfmathsetmacro{\pvdzu}{\pvdzx + \pvdzy*\pCos}    \pgfmathsetmacro{\pvdzv}{\pvdzy*\pSin}
\pgfmathsetmacro{\lwpu}{\lwpx + \lwpy*\pCos}       \pgfmathsetmacro{\lwpv}{\lwpy*\pSin}
\pgfmathsetmacro{\lwmu}{\lwmx + \lwmy*\pCos}       \pgfmathsetmacro{\lwmv}{\lwmy*\pSin}

\pgfmathsetmacro{\plxlo}{-2}  \pgfmathsetmacro{\plxhi}{4}
\pgfmathsetmacro{\plylo}{-3}  \pgfmathsetmacro{\plyhi}{3}
\pgfmathsetmacro{\plAu}{\plxlo + \plylo*\pCos}  \pgfmathsetmacro{\plAv}{\plylo*\pSin}
\pgfmathsetmacro{\plBu}{\plxhi + \plylo*\pCos}  \pgfmathsetmacro{\plBv}{\plylo*\pSin}
\pgfmathsetmacro{\plCu}{\plxhi + \plyhi*\pCos}  \pgfmathsetmacro{\plCv}{\plyhi*\pSin}
\pgfmathsetmacro{\plDu}{\plxlo + \plyhi*\pCos}  \pgfmathsetmacro{\plDv}{\plyhi*\pSin}

\def\panelW{6}
\def\panelH{-3.5}
\def\panelS{0.8}
\def\lblX{-2.5}
\def\brcA{-0.2}
\def\brcB{-0.9}

\node[rlbl] at (\lblX, 1.5)
  {\small \textbf{Prediction level:}\\
   Log-probs are nonlinear in $z$; log-odds are linear.};
\node[rlbl] at (\lblX, \panelH+1.5)
  {\small \textbf{Attribution level:}\\
   Differences in projections are projections of differences.};

\begin{scope}[shift={(0,0)}, scale=\panelS, every node/.style={scale=\panelS}]
  \fill[cplane, opacity=0.5]
    (\plAu,\plAv) -- (\plBu,\plBv) -- (\plCu,\plCv) -- (\plDu,\plDv) -- cycle;
  \draw[vec] (0,0) -- (\wpu,\wpv) node[lbl, right] {$u_+$};
  \draw[vec] (0,0) -- (\wmu,\wmv) node[lbl, left] {$u_-$};
  \draw[vecd, cvec] (\wmu,\wmv) -- (\wpu,\wpv);
  \draw[vecbold] (0,0) -- (\zu,\zv) node[lbl, above right] {$z$};
  \draw[proj] (\zu,\zv) -- (\lwpu,\lwpv);
  \fill[black] (\lwpu,\lwpv) circle(2pt);
  \node[lbl, below right] at (\lwpu,\lwpv) {$\logit_+$};
  \draw[proj] (0,0) -- (\lwmu,\lwmv);
  \draw[proj] (\zu,\zv) -- (\lwmu,\lwmv);
  \fill[black] (\lwmu,\lwmv) circle(2pt);
  \node[lbl, below right] at (\lwmu,\lwmv) {$\logit_-$};
  \node[plbl] at (-1.5, \zv) {(a)};
\end{scope}

\begin{scope}[shift={(\panelW,0)}, scale=\panelS, every node/.style={scale=\panelS}]
  \fill[cplane, opacity=0.5]
    (\plAu,\plAv) -- (\plBu,\plBv) -- (\plCu,\plCv) -- (\plDu,\plDv) -- cycle;
  \draw[vec, fade] (0,0) -- (\wpu,\wpv);
  \draw[vec, fade] (0,0) -- (\wmu,\wmv);
  \draw[vecbold, cvec] (0,0) -- (\vu,\vv) node[lbl, right] {$v$};
  \draw[vecbold] (0,0) -- (\zu,\zv) node[lbl, above right] {$z$};
  \draw[proj] (\zu,\zv) -- (\pvzu,\pvzv);
  \fill[black] (\pvzu,\pvzv) circle(2pt);
  \draw[brc] ($(0,0)+(0,\brcA)$) -- ($(\pvzu,\pvzv)+(0,\brcA)$)
    node[brclbl, align=center] {$\begin{aligned} \ell(z) &= \logit_+ - \logit_- \\ &= z \cdot v \end{aligned}$};
  \node[plbl] at (-1.5, \zv) {(b)};
\end{scope}

\begin{scope}[shift={(0,\panelH)}, scale=\panelS, every node/.style={scale=\panelS}]
  \fill[cplane, opacity=0.5]
    (\plAu,\plAv) -- (\plBu,\plBv) -- (\plCu,\plCv) -- (\plDu,\plDv) -- cycle;
  \draw[vecbold, cvec] (0,0) -- (\vu,\vv) node[lbl, right] {$v$};
  \draw[vecbold] (0,0) -- (\zu,\zv) node[lbl, above right] {$z$};
  \draw[vecbold] (0,0) -- (\zpu,\zpv) node[lbl, right] {$z'$};
  \draw[vecd, cdz] (\zpu,\zpv) -- (\zu,\zv);
  \draw[proj] (\zu,\zv) -- (\pvzu,\pvzv);
  \fill[black] (\pvzu,\pvzv) circle(2pt);
  \draw[proj] (\zpu,\zpv) -- (\pvzpu,\pvzpv);
  \fill[black] (\pvzpu,\pvzpv) circle(2pt);
  \draw[brc] ($(0,0)+(0,\brcA)$) -- ($(\pvzpu,\pvzpv)+(0-0.1,\brcA)$)
    node[brclbl] {$\ell(z')$};
  \draw[brc] ($(\pvzpu,\pvzpv)+(0.1,\brcA)$) -- ($(\pvzu,\pvzv)+(0.1,\brcA)$)
    node[brclbl] {$\Delta\ell$};

  \draw[brc] ($(0,0)+(0,\brcB)$) -- ($(\pvzu,\pvzv)+(-0.1,\brcB)$)
    node[brclbl] {$\ell(z)$};
  \node[plbl] at (-1.5, \zv) {(c)};
\end{scope}

\begin{scope}[shift={(\panelW,\panelH)}, scale=\panelS, every node/.style={scale=\panelS}]
  \fill[cplane, opacity=0.5]
    (\plAu,\plAv) -- (\plBu,\plBv) -- (\plCu,\plCv) -- (\plDu,\plDv) -- cycle;
  \draw[vec, fade] (0,0) -- (\zu,\zv);
  \draw[vec, fade] (0,0) -- (\zpu,\zpv);
  \draw[vecbold, cvec] (0,0) -- (\vu,\vv) node[lbl, right] {$v$};
  \draw[vecbold, cdz] (0,0) -- (\dzu,\dzv) node[lbl, above right] {$\Delta z$};
  \draw[proj] (\dzu,\dzv) -- (\pvdzu,\pvdzv);
  \fill[black] (\pvdzu,\pvdzv) circle(2pt);
  \draw[brc] ($(0,0)+(0,\brcA)$) -- ($(\pvdzu,\pvdzv)+(0,\brcA)$)
    node[brclbl, align=center] {$\begin{aligned} \Delta\ell &= z \cdot v - z' \cdot v \\&= \Delta z \cdot v \\ &= \|\Delta z\|\,\|v\|\,\cos(\Delta z, v) \end{aligned}$};
  \node[plbl] at (-1.5, \zv) {(d)};
\end{scope}

\end{tikzpicture}}
    \vspace{-0.5cm}
    \caption{
        The geometric intuition behind our evaluations. 
        \textbf{(a)} Logits are computed from a representation $z$ by projection onto unembedding vectors $u_-$ and $u_+$; however, these projections are not recoverable from the top-$K$ log-probabilities exposed by most LLM inference providers.
        \textbf{(b)} The log-odds are equivalent to the difference in logits, which in turn equals the projection of $z$ onto the difference vector $v = u_+ - u_-$.
        This quantity can be derived from log-probabilities alone.
        \textbf{(c)} Given a base representation $z$ and a modified representation $z'$, we define the attribution as the difference in log-odds derived from the two representations.
        This is equal to the difference in projection lengths.
        \textbf{(d)} The difference in projections can also be thought of as a projection of the difference, i.e. $\Delta z \cdot v$.
        Thus, for any $\Delta z$, the attribution depends on its magnitude, $\|\Delta z\|$, and its alignment with $v$, $\cos(\Delta z, v)$.
    }
    \label{fig:representation_geometry}
\end{figure*}

    \textbf{Our contributions.} Our contributions are as follows:
    \begin{enumerate}[leftmargin=*,nosep]
        \item We argue that the question of whether mechanistic findings transfer across models deserves direct measurement, and propose \textit{surrogate fidelity} as a conceptual framework for it.
        \item We propose a hierarchy of surrogate fidelity metrics---prediction, attribution, mechanistic, and cross-level---for classification benchmarks.
        Prediction and attribution require only top-$K$ log-probabilities, while mechanistic- and cross-level fidelity require open-weight model access.
        \item Evaluating two open-weight families (Qwen 2.5, Llama 3) and two closed-API families (GPT, Gemini) across multiple scales (0.5B--70B parameters) on four classification benchmarks, we find a striking dissociation: models that share similar representations and predictions can nonetheless disagree sharply on causal attributions.
        \item We identify an \textit{access--validity inversion}: high-agreement white-box signals like attention fail to predict more causally relevant black-box attributions.
    \end{enumerate}

\section{Preliminaries}
\label{sec:preliminaries}

    All quantities are implicitly conditioned on a single input $x$ and its last-layer normalized representation $z$ in model $M$; we make these explicit in \cref{sec:method} when comparing across models and inputs.

    \textbf{Static geometry.}
        We review the geometric relationship between a transformer's residual stream and its output probabilities, establishing the quantities on which our fidelity metrics operate.
        
        Given a normalized last-layer representation $z \in \mathbb{R}^d$, the logit for token $t$ is the dot product
            $\logit_t = z \cdot u_t$,
        where $u_t \in \mathbb{R}^d$ is the column of the unembedding matrix $W_U \in \mathbb{R}^{d \times V}$ corresponding to token $t$.
        The output probability is computed via softmax:
            $P(t) = \exp(z \cdot u_t) / \sum_{i=1}^{V} \exp(z \cdot u_i).$
        Thus, the log-probability of any token is $\log P(t) = z \cdot u_t - C$, where $C = \log \sum_{s=1}^{V} \exp(z \cdot u_s)$ is the log of the softmax normalization constant.
        Without access to the full vocabulary of logits, $C$ is not recoverable from a handful of top-$K$ log-probabilities.
        
        Since $C$ is shared across all logits, it cancels when we take the \textit{difference} of two log-probabilities:
        \begin{equation}
            \log P(t) - \log P(s)
            = (z \cdot u_t - C) - (z \cdot u_s - C)
            = z \cdot (u_t - u_s).
            \label{eq:log_odds_linear}
        \end{equation}
        Defining the \textit{readout (log-odds) direction} $v = u_+ - u_-$ for a binary classification task with positive and negative class tokens, the log-odds reduce to a single dot product,
            $\ell(x) = z \cdot v.$
        While individual log-probabilities are nonlinear functions of $z$, the log-odds are \textit{exactly linear} in the representation.
        This is the key observation that makes our fidelity framework possible: log-odds are both API-accessible (as a difference of two log-probabilities) and geometrically interpretable (as a projection onto a readout direction in representation space).
        \cref{fig:representation_geometry}(a--b) illustrates this progression from individual logits to the log-odds direction.%

        \paragraph{Residual stream decomposition.}
        The residual stream of a transformer is a running sum: after layer $l$, the state is $h^{(l)} = h^{(0)} + \sum_{k=1}^{l}(a^{(k)} + m^{(k)})$, where $h^{(0)}$ is the token embedding and $a^{(k)}, m^{(k)} \in \mathbb{R}^d$ are the outputs of the attention and MLP sublayers at layer $k$.
        Because log-odds are linear in the normalized last layer $z$, this additivity extends directly to the prediction: unembedding any intermediate state $\operatorname{RMSNorm}(h^{(l)}) \cdot v$ yields a valid log-odds estimate at depth $l$ --- the \textit{logit lens} \citep{nostalgebraist2020logitlens} --- while decomposing the final projection term by term yields \textit{direct logit attribution} (DLA) \citep{elhage2021mathematical}.
        Writing out both relationships:
        \begin{equation}
            h^{(L)} = h^{(0)} + \sum_{l=1}^{L} \left(a^{(l)} + m^{(l)}\right), \quad
            \ell(x)
            = \operatorname{RMSNorm}\left(
            h^{(L)} \right) \cdot v.
            \label{eq:dla}
        \end{equation}
        The logit lens reads the renormalized partial sums $\operatorname{RMSNorm}(h^{(l)}) \cdot v$ cumulatively across depth; DLA reads the individual terms $a^{(l)} \cdot v$ and $m^{(l)} \cdot v$ as per-component contributions to the final log-odds.
        We investigate the DLA of predictions and attributions in \Cref{sec:per_layer}.
        
    \textbf{Intervention and attribution.}
        For an intervention\footnote{%
            Our framework is intervention-agnostic. 
            For our experiments, we use prompt-level leave-one-out ablations because they are compatible with black-box API access.
        } that changes the representation from $z$ to $z'$, the resulting change in log-odds is
        \begin{equation}
            \Delta\ell = \ell(z) - \ell(z') = \Delta z \cdot v,
            \label{eq:attribution}
        \end{equation}
        where $\Delta z = z - z'$ is the perturbation in the residual stream.
        We call $\Delta\ell$ the \textit{attribution} of the intervention.
        Concretely, attribution measures the causal effect of changing the input on $\ell$; it is not a complete internal mechanism. 
        
        Because attribution is a dot product, it decomposes into three factors:
        \begin{equation}
            \Delta\ell
            = \|\Delta z\|\; \|v\|\; \cos(\Delta z, v).
            \label{eq:factorize_attribution}
        \end{equation}
        Since $\|v\|$ is fixed for a given model and task, attribution is governed by two properties of the perturbation: its \textit{magnitude} $\|\Delta z\|$ and its \textit{alignment} $\cos(\Delta z, v)$ with the log-odds direction.
        A large perturbation orthogonal to $v$ produces zero attribution; a small perturbation aligned with $v$ can shift the prediction substantially.
        \cref{fig:representation_geometry}(c--d) illustrates this geometry.

        \textbf{The role of normalization.} 
        In closed models, $\Delta \ell$ only reveals information about the  post-normalization representations $z$ and $z'$.
        If we think of perturbations as acting on the \textit{pre-normalization} residual stream (written as $h$), then
        for a normalization
        $h \to h\,/\,f(h)$ that acts by scalar rescaling\footnote{%
            For RMSNorm (used in Qwen 2.5 and Llama 3), $f(h) = \|h\|/\sqrt{d}$.
        }, the post-normalization perturbation decomposes further as
        \begin{equation}
            \Delta z
            = \underbrace{\frac{\Delta h}{f(h)}}_{\text{direct}}
              \;+\;
              \underbrace{h'\,\frac{f(h') - f(h)}{f(h)\,f(h')}}_{\text{indirect}}.
            \label{eq:norm_decomp}
        \end{equation}
        The \textit{direct} term is the pre-normalization perturbation $\Delta h$ rescaled by the baseline normalization factor, i.e. the contribution that would remain if $h'$ was normalized by the original norm $\|h\|$.
        The \textit{indirect} term arises because changing $h \to h'$ also changes the normalization denominator, effectively rescaling the entire post-normalization representation $z'$.
        This term is nonzero whenever $f(h) \neq f(h')$ and $h' \cdot v \neq 0$, even if $\Delta h$ itself is orthogonal to $v$.

\begin{figure*}[t]
\resizebox{\textwidth}{!}{
    \begin{minipage}[t]{2.8in}
        \textbf{(a)}\\
        \includegraphics{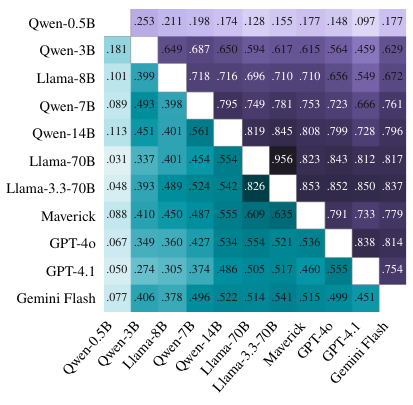}
    \end{minipage}%
    \hfill
    \begin{minipage}[t]{2.7in}
        \textbf{(b)}\\
        \includegraphics{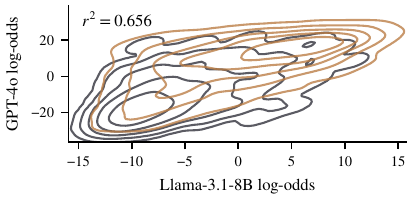}
        \vspace{-1em}\\
        \textbf{(c)}\\
        \includegraphics{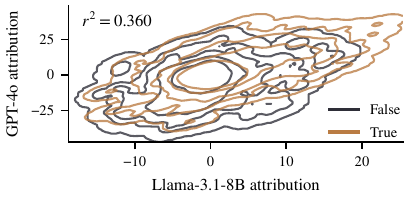}
    \end{minipage}%
    \hfill
    \begin{minipage}[t]{1.25in}
        \textbf{(d)}\\
        \input{figures/prompt_callout.tex}
    \end{minipage}%
    }
    \caption{
        \textbf{(a)} Pairwise {\color[HTML]{584486}$\bm\Fpred$} and {\color[HTML]{0091A1} $\bm\Fattr$} heatmaps for eleven models across four families.
        \textbf{(b)} Log-odds contour plot for a representative cross-family pair (Llama 3-8B vs.\ GPT-4o), stratified by ground truth answer ({\color[HTML]{BA7C43} \textbf{True}} vs {\color[HTML]{2E2E38} \textbf{False}}).
        Log-odds for this pair fall into two clusters by label: pooled $r^2$ is .66, but drops to .42 and .35 within a class, so much of the prediction agreement is agreement on the answer.
        \textbf{(c)} Ablation contour plot for the same pair.
        Here ground-truth labels do not reveal additional structure.
        \textbf{(d)} Example BoolQ prompt with a sentence-level ablation (greyed-out text).
    }
    \label{fig:dashboard}
\end{figure*}

\section{Method}
\label{sec:method}
    We formalize surrogate fidelity at three levels of increasing depth: do two models make similar \textit{predictions}, produce similar \textit{attributions} under the same perturbations, and share \textit{internal mechanisms}?
    Each level imposes a stricter test of whether mechanistic insights transfer across model families.

    Prediction and attribution fidelity require only log-probabilities and are therefore computable for any model pair;
    mechanistic fidelity additionally requires access to model internals.

    Let $M_S$ and $M_T$ denote surrogate and target models, and $\mathcal{D} = \{x_1, \ldots, x_n\}$ an evaluation corpus.
    For each model $M$, the quantities defined in \cref{sec:preliminaries} --- log-odds $\ell_M(x)$, attributions $\Delta\ell_M(x, i)$, perturbation norms $\|\Delta z_M(x, i)\|$, and alignments $\gamma_M(x, i) = \cos(\Delta z_M, v_M)$ --- are stacked over all prompts and segments to form vectors (denoted in boldface).
    We additionally write $\alpha_M(x, i)$ for attention-based importance scores.
    Agreement is measured by row-pooled $\r$ (scalar signals) or RV coefficients (multivariate signals).
    Confidence intervals use 1,000 prompt-level bootstrap resamples, and tables summarize scores by their median.

    \textit{Prediction fidelity} measures agreement on outputs:
    \begin{equation}
        F_\text{pred}(M_S, M_T; \mathcal{D})
        = \r\!\left(
            \boldsymbol{\ell}_{M_S},\
            \boldsymbol{\ell}_{M_T}
        \right).
        \label{eq:f_pred}
    \end{equation}

    \textit{Attribution fidelity} measures agreement on causal importance under leave-one-out ablation:
    \begin{equation}
        F_\text{attr}(M_S, M_T; \mathcal{D})
        = \r\!\left(
            \boldsymbol{\Delta\ell}_{M_S},\ 
            \boldsymbol{\Delta\ell}_{M_T}
        \right).
        \label{eq:f_attr}
    \end{equation}
    High $F_\text{pred}$ with low $F_\text{attr}$ signals that models agree on outputs while disagreeing on which inputs drive them, a critical failure mode for surrogate-based interpretability.

    For open-weight models, we can additionally define mechanistic metrics, referred to generically as $F_\textrm{mech}$.
    The decomposition in \cref{eq:factorize_attribution} motivates two mechanistic fidelity metrics measuring agreement in perturbation magnitude and alignment respectively:
    \begin{align}
        \label{eq:f_mag}
        F_\text{mag}(M_S, M_T; \mathcal{D})
        &= \r\!\left(
            \|\boldsymbol{\Delta z}_{M_S}\|,\ 
            \|\boldsymbol{\Delta z}_{M_T}\|
        \right), \\
        \label{eq:f_align}
        F_\text{align}(M_S, M_T; \mathcal{D})
        &= \r\!\left(
            \boldsymbol{\gamma}_{M_S},\ 
            \boldsymbol{\gamma}_{M_T}
        \right).
    \end{align}
    Attention patterns have also been used as a proxy for mechanistic importance~\citep{abnar2020quantifying}; we include \textit{attention fidelity} as a third mechanistic metric:
    \begin{equation}
        F_\text{attn}(M_S, M_T; \mathcal{D})
        = \r\!\left(
            \boldsymbol{\alpha}_{M_S},\ 
            \boldsymbol{\alpha}_{M_T}
        \right).
        \label{eq:f_attn}
    \end{equation}
    For attention, we consider several attention aggregation strategies (mean, max, rollout; see \cref{sec:experiments}).

    Finally, \textit{cross-level fidelity} asks whether the surrogate's internals predict the target's attributions:
    \begin{equation}
        F_\text{cross}(M_S, M_T; \mathcal{D})
        = \r\!\left(
            \boldsymbol{\alpha}_{M_S},\ 
            \boldsymbol{\Delta\ell}_{M_T}
        \right).
        \label{eq:f_cross}
    \end{equation}
    For closed $M_T$, $F_\text{cross}$ quantifies whether mechanistic analysis of an open $M_S$ yields actionable insight about $M_T$.

    \textbf{Readout-compatible linear transfer.}
    Our use of Pearson $\r$ is motivated by a readout-compatible notion of representation transfer.
    Concretely, for idealized models $M_S, M_T$ related through a readout-compatible linear transformation, i.e. $z_T \approx A z_S$ and $A^T v_T \approx cv_S$,
    \begin{equation}
        \label{eq:linear_transform}
        z_T \cdot v_T \approx A z_S \cdot v_T
        = z_S \cdot (A^T v_T)
        \approx c(z_S \cdot v_S)
        = c\ell_S,
    \end{equation}
    i.e. the log-odds of $M_S$ and $M_T$ are scalar multiples.
    For attributions, $\Delta \ell_T \approx c \Delta \ell_S$ analogously.
    
    Thus, $\r$ measures the fraction of variation explained by $M_S$ under the simplest readout-compatible linear calibration.
    Low $\r$ in a single dimension rules out high-fidelity transfer under such a readout-compatible $A$, though it leaves open weaker forms of representational similarity.

    Analogous correlation-based measures exist in higher dimensions: the RV coefficient \citep{robert1976unifying} for multivariate readouts and centered kernel alignment (CKA) \citep{kornblith2019similarity} for full representations.
    In multivariate settings, we therefore report centered RV scores over all pairwise class log-odds. 
    This lineage grounds our use of $\r$ as a readout-level probe into overall representational similarity and motivates our use of RV for multivariate classification.

\begin{table*}[t]
    \centering
    \caption{
        Cross-model agreement (%
        Pearson $\r$%
        ) on BoolQ sentence-level ($n=27{,}516$ segments / $3{,}270$ prompts). 
        Each cell is min$/$\textbf{median}$/$max over the relevant model-pair set. 
        $F_{\mathrm{xxx}}$ measures agreement on a single signal; 
        $F_{\mathrm{cross}}$ measures how well a signal from model $M_S$ predicts ablation response in model $M_T$. 
        Open$\to$Open = within the 5 open instruct models; Open$\to$Closed = open$\times$closed-source pairs; All$\to$All = all 11 models. 
        Cells marked --- cannot be filled because mechanistic signals are unavailable for closed models.
    }
    \scriptsize
    \begin{tabular}{llccc}
\toprule
Group & Metric & Open$\to$Open & Open$\to$Closed & All$\rightarrow$All\\
\midrule
Black-box & $F_{\mathrm{pred}}$ & .174 / .650 / .795 & .097 / .669 / .845 & .097 / .718 / .956 \\
 & $F_{\mathrm{attr}}$ & .089 / .399 / .561 & .031 / .404 / .555 & .031 / .451 / .826 \\
\midrule
Mechanistic & $F_{\mathrm{attn}}^{\mathrm{mean}}$ & .747 / .904 / .984 & --- & --- \\
 & $F_{\mathrm{attn}}^{\mathrm{max}}$ & .643 / .859 / .906 & --- & --- \\
 & $F_{\mathrm{attn}}^{\mathrm{rollout}}$ & .512 / .861 / .938 & --- & --- \\
 & $F_{\mathrm{mag}}$ & .639 / .843 / .904 & --- & --- \\
 & $F_{\mathrm{align}}$ & .050 / .201 / .296 & --- & --- \\
\midrule
Mechanistic-to-causal & $F_{\mathrm{mag}\to|\mathrm{attr}|}$ & .000 / .065 / .231 & .000 / .112 / .283 & .000 / .108 / .283 \\
 & $F_{\mathrm{align}\to\mathrm{attr}}$ & .048 / .254 / .381 & .033 / .267 / .362 & .033 / .257 / .381 \\
 & $F_{\mathrm{attn}\to\mathrm{attr}}^{\mathrm{mean}}$ & .000 / .005 / .042 & .000 / .001 / .024 & .000 / .002 / .042 \\
 & $F_{\mathrm{attn}\to\mathrm{attr}}^{\mathrm{max}}$ & .000 / .003 / .032 & .000 / .002 / .019 & .000 / .002 / .032 \\
 & $F_{\mathrm{attn}\to\mathrm{attr}}^{\mathrm{rollout}}$ & .000 / .003 / .063 & .000 / .004 / .014 & .000 / .003 / .063 \\
\bottomrule
\end{tabular}

    \label{tab:f-summary}
\end{table*}

\section{Evaluation Setting}
\label{sec:experiments}

    This is a summary of our experimental setup;
    full details can be found in the Appendix, \cref{sec:experimental-details}.

    \textbf{Benchmarks.}
        We focus our initial investigation on classification tasks: when a model's output space is restricted, the log-odds $\ell_M(x)$ collapse to one or more scalars per prompt, providing a compact and interpretable summary of the model's internal decision boundary.
        Classification is practical under API constraints: the top-$K$ most likely tokens are typically constrained to a small, predictable set of label variants.
        In more open-ended settings, we expect significantly more missing values.
        
        We evaluate scalar fidelity on
        BoolQ \citep{clark2019boolq} and  WinoGrande \citep{sakaguchi2021winogrande},
        three-class fidelity on ANLI R1--R3 \citep{nie2020adversarial},
        and four-class fidelity on RACE \citep{lai-etal-2017-race}.
        For multi-class benchmarks, we report centered RV over pairwise log-odds readouts. 

    \textbf{Models.}
        We compare models across four families and multiple scales:
        \begin{itemize}[leftmargin=*,nosep]
            \item \textbf{Llama} \citep{grattafiori2024llama3herdmodels}: Llama~3.1-8B, Llama~3.1-70B, Llama~3.3-70B, Llama~4 Maverick
            \item \textbf{Qwen} \citep{qwen2025qwen25technicalreport}: Qwen 2.5-0.5B, Qwen 2.5-3B, Qwen 2.5-7B, Qwen 2.5-14B
            \item \textbf{GPT} \citep{achiam2023gpt}: GPT-4o, GPT-4.1
            \item \textbf{Gemini} \citep{comanici2025gemini}: Gemini~2.5 Flash Lite
        \end{itemize}
        
        The Llama and Qwen families provide multiscale open-weight models, enabling representation-level extraction (for $F_\text{mech}$ and $F_\text{cross}$) and investigation into how surrogate fidelity scales with model size; we evaluate hosted variants of 70B Llama models due to compute constraints.
        GPT and Gemini models are accessible only through black-box APIs exposing top-$K$ log-probabilities, restricting us to $F_\text{pred}$ and $F_\text{attr}$ for these targets.
        Notably, the Claude and GPT-5 families, which do not expose any log-probabilities to end users, are excluded from this evaluation.
    
    \textbf{Ablation protocol.}
        For attribution fidelity, we segment each prompt's user and system component and remove individual segments to produce ablated inputs $\tilde{x}_i$.
        We evaluate ablations at sentence- and word-level granularity. 
        We evaluate all sentence-level ablations and an $n=10{,}000$ subset of word-level ablations.
        We segment the complete dialog identically for attention and ablation.

\section{Results}
    \label{sec:results}
    We first evaluate prediction and sentence-level attribution fidelity on the BoolQ validation set across all eleven models.
    We report pairwise $\r$ values in \cref{tab:f-summary} and show the full pairwise fidelities and cross-family contour plots for $\Fpred$ and $\Fattr$ in \cref{fig:dashboard}, as well as an example BoolQ prompt.
    
    \subsection{Prediction fidelity}
    \label{sec:prediction-fidelity}
        Prediction fidelity varies dramatically across model pairs, and family membership does not guarantee higher fidelity.
        Predictably, Llama 3.1-70B and Llama 3.3-70B have high mutual $\Fpred$ ($0.956$) and exhibit similar $\Fpred$ profiles generally.
        The same is true for Qwen-7B and Qwen-14B ($0.795$) and GPT-4o and GPT-4.1 $(0.838)$.
        By contrast, the highest cross-family $\Fpred$ is between Llama 3.3-70B and GPT-4o at $0.852$; cross-family ranges are generally slightly lower.
        
        Qwen-0.5B is an outlier on all fronts, with low $\Fpred$ to all other models, including larger Qwens.
        This is consistent with its uniquely small size, and is contextualized by the generally low confidence and poor log-probability separation reported in \cref{fig:histograms} in the Appendix. 

    \subsection{Attribution fidelity}
    \label{sec:attribution-fidelity}
        For each model and each BoolQ prompt, we computed sentence-level ablations and measured the attribution fidelity across model pairs.
        
        At this level, attribution agreement is lower than prediction fidelity on all BoolQ model pairs.
        Median $\Fattr$ is $0.451$ overall, and $0.399$ among open models.
        The Qwen-7B--Qwen-14B pair reached $\Fattr = 0.561$, and the Llama 3.1-70B--Llama 3.3-70B pair attained the overall maximum of $0.826$.

        \textbf{Word-level validation.}
        To rule out that the low ablation agreement is an artifact of sentence-level coarseness, we repeat the analysis at word level, using leave-one-out \emph{single deletion}~\citep{Nauta_2023} of individual whitespace tokens. 
        We subsample 10{,}000 (prompt, word) ablation pairs per model from a pool of 476{,}154 possible BoolQ word ablations using a shared random seed. 
        
        Per Table \ref{tab:cross_benchmark_results}, switching from sentence- to word-level ablation leaves mean and rollout attention fidelity mostly unchanged but sharply reduces $F_{\mathrm{attr}}$ (from $.451$ to $.168$), driven almost entirely by the collapse of $F_{\mathrm{align}}$ ($.201 \to .019$). 

\begin{figure*}[t]
    \resizebox{\textwidth}{!}{
    \includegraphics{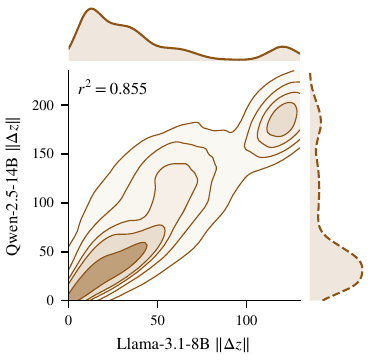}%
    \includegraphics{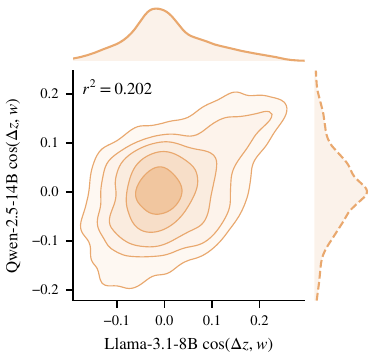}%
    \includegraphics{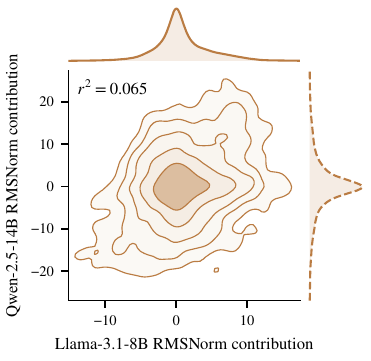}%
    }
    \caption{
        Perturbations affect attribution through their norm (\textbf{left}), alignment with the log-odds direction $v$ (\textbf{middle}), and indirectly via RMSNorm (\textbf{right}).
        We plot joint and marginal distributions of each, for a representative model pair.
        Across models, the magnitude of an ablation response agrees in the residual stream (left, $r^2 = .86$), but which way it moves relative to the readout does not (middle, $r^2 = .20$); indirect effects through RMSNorm are small and nearly uncorrelated (right, $r^2=.07$).
    }
    \label{fig:attribution-decomposition}
\end{figure*}

    \subsection{Mechanistic fidelity}
    \label{sec:representational-fidelity}
        In this section, we focus both on decomposing $\Fattr$ according to the components of attribution (norm of perturbation, angle between perturbation and readout direction, and normalization-induced rescaling) and investigating the role attention could play in surrogate model predictions.

        \textbf{Decomposing attributions.}
        \cref{fig:attribution-decomposition} shows that models' responses to perturbations are stable in magnitude (high $\Fmag$) but not direction (low $\Falign$).
        Low $\Fattr$ is accompanied by disagreement in the direction of the perturbation-response vector $\Delta z$.
        The maximum cosine similarity between $\Delta z$ and $v$ tops out around 0.4, with most cosine similarities landing near zero.
        Using unsigned cosine similarity does not improve $\Falign$.

        While most perturbations have negligible effects on log-odds from the perspective of the projection of $\Delta z$ onto $v$, it is also possible for perturbations to affect a model's log-odds through the RMSNorm's rescaling of the overall vector, as rescaling a vector by $\alpha$ also rescales $\ell$ by $\alpha$.

        We find that the absolute magnitude of the RMSNorm contribution is 6.0--16.2\% of the direct residual-change term.
        Interestingly, the RMSNorm rescaling opposes the direct term in approximately half of cases, meaning that while this component of the model architecture can contribute to lower $\Fattr$ scores, it is unlikely to bias it in a specific direction. 
        For models $\geq$3B parameters, there is no clear trend in terms of the RMSNorm's contribution to $\Delta \ell$, with Llama 3-8B sitting in the middle of the Qwen results across all metrics.

        \textbf{Attention: stable but limited.}
        In contrast to ablation attributions, attention-based scores exhibit strong agreement across the board.  
        Mean attention correlations range from $\r = 0.747$ to $0.984$, and attention rollout, which accounts for residual connections across layers~\citep{abnar2020quantifying}, achieves median $\r = 0.861$.
        Even Qwen-0.5B, which was an outlier for prediction and attribution fidelity, attends to the same tokens as its larger siblings.
        
        Despite high within-method consistency, the three attention aggregation strategies are not interchangeable. 
        Mean and max attention correlate moderately ($\r \approx 0.670$--$0.957$), but rollout attention is almost uncorrelated with these ($\r \leq 0.054$).
        This methodological sensitivity underscores the importance of specifying the aggregation scheme when reporting attention-based results.

        \begin{wrapfigure}{r}{0.5\textwidth}
            \centering
            \resizebox{0.5\textwidth}{!}{\includegraphics{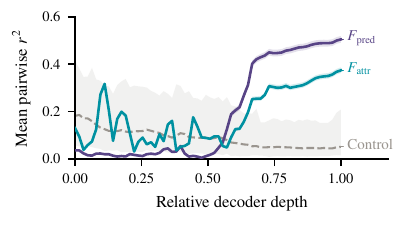}}
            \caption{Bootstrap-CI pair-averaged per-layer {\color[HTML]{584486} $\Fpred^{(l)}$} and
            {\color[HTML]{0091A1} $\Fattr^{(l)}$} versus relative depth, with a random-unembedding control overlay (gray dashed).
            $\Fattr$ is unstable in early layers, and rises analogously to $\Fpred$ in the second half of the forward pass.
            Control readouts stay low, suggesting a task-specific aspect to this measurement.
            }
            \label{fig:per_layer:random_dira}
        \end{wrapfigure}

        The contrast between high mechanistic fidelity and low attribution fidelity reveals a dissociation: models can share attention patterns while diverging in causal structure. 
        Attention consistency (often used as informal evidence that ``models are doing the same thing'') is a weak proxy for the mechanistic agreement that MI practitioners actually need.

        \textbf{Per-layer fidelity.}
        We track the evolution of BoolQ $\Fpred$ and $\Fattr$ across layers using the logit lens in
        \cref{fig:per_layer:random_dira}. 
        Both metrics start off low and co-emerge approximately halfway through the forward pass across ten open-model pairs, with $\Fpred > \Fattr$ throughout late layers;
        in early layers, $\Fattr$ shows pronounced non-monotonic variation.
        As a control, we evaluate 256 paired contrasts for random tokens: for instance, a ``fun'' vs ``new'' direction, aggregated across lexical variants.
        Agreement between these readouts actually \textit{decreases} with depth, suggesting a task-specific cause.
    
        See \cref{sec:per_layer} for extended experiments, analysis, and experiment setup.

\begin{table*}[t]
    \centering
    \caption{
        Median agreement on all benchmarks: BoolQ and WinoGrande report Pearson $\r$; ANLI and RACE report centered RV.
        Absolute magnitudes are not comparable across RV and $r^2$, but their ordering is meaningful.
        The first group of rows aggregates over open and closed models; the second group over open models only; and in the final group, open models predict all models' attributions.
    }
    \scriptsize
    \begin{tabular}{lccccccc}
\toprule
Metric & BoolQ & ANLI R1 & ANLI R2 & ANLI R3 & WinoGrande & BoolQ word & RACE RV \\
\midrule
$F_{\mathrm{pred}}$ & .718 & .531 & .443 & .520 & .224 & .719 & .791 \\
$F_{\mathrm{attr}}$ & .451 & .308 & .265 & .291 & .152 & .168 & .334 \\
\midrule
$F_{\mathrm{attn}}^{\mathrm{mean}}$ & .904 & .949 & .948 & .950 & .852 & .801 & .905 \\
$F_{\mathrm{attn}}^{\mathrm{max}}$ & .859 & .939 & .938 & .936 & .817 & .602 & .832 \\
$F_{\mathrm{attn}}^{\mathrm{rollout}}$ & .861 & .885 & .885 & .884 & .877 & .816 & .834 \\
$F_{\mathrm{mag}}$ & .843 & .864 & .868 & .881 & .668 & .538 & .777 \\
$F_{\mathrm{align}}$ & .201 & .093 & .095 & .080 & .016 & .019 & .094 \\
\midrule
$F_{\mathrm{mag}\to|\mathrm{attr}|}$ & .108 & .244 & .215 & .236 & .227 & .205 & .280 \\
$F_{\mathrm{align}\to\mathrm{attr}}$ & .257 & .066 & .068 & .068 & .059 & .035 & .153 \\
$F_{\mathrm{attn}\to\mathrm{attr}}^{\mathrm{mean}}$ & .002 & .016 & .011 & .015 & .018 & .002 & .044 \\
$F_{\mathrm{attn}\to\mathrm{attr}}^{\mathrm{max}}$ & .002 & .015 & .013 & .010 & .025 & .003 & .053 \\
$F_{\mathrm{attn}\to\mathrm{attr}}^{\mathrm{rollout}}$ & .003 & .008 & .011 & .009 & .020 & .001 & .015 \\
\bottomrule
\end{tabular}

    \label{tab:cross_benchmark_results}
\end{table*}

    \subsection{Additional experiments}
        \textbf{Robustness.}
        We carry out a number of robustness experiments in \cref{app:robustness}. 
        On BoolQ, we measured a noise floor of 0.175 by shuffling attributions and a noise ceiling of at least 0.913 by resampling five times; the median $\Fattr$ of $0.451$ is quite far from either value. 
        Centering attributions per prompt, ablating only passage sentences, and controlling for segment length affect $\Fattr$ by less than .05.
        Passage-only perturbation magnitude explains .425 of the size of another model's attribution, compared to .108 over all segments.
        For pooled BoolQ prompts, $\Fpred$ and $\Fattr$ medians are .718 and .451, respectively.
        Stratifying by ground-truth label largely closes this gap: .498/.428 for True, and .407/.429 for False.
        Further restricting to single-label prompts both models get right reverses the trend (.214/.559, .197/.508).
        This last comparison selects a same-sign quadrant of the models' joint log-odds on a subset of prompts; we present it as a useful diagnostic alongside our pooled results.

        \textbf{Case study: when Qwen 2.5-14B explains GPT-4o.}
        Average fidelity hides large differences between prompts; we therefore study GPT-4o through its best open surrogate, Qwen 2.5-14B, prompt-by-prompt using our framework in \cref{app:case-study}.
        Qwen is a much better surrogate for 4o when one segment strongly supports its answer: for the top third of ``high support'' prompts, Qwen names 4o's most influential passage sentence 82\% of the time, compared to 29\% by chance or 63\% using a simple word-overlap baseline.
        By comparison, the overlap rate is a mere 47\% for the other two-thirds.
        This pattern generalizes to other targets (76--86\%/43--53\% across six closed target models) and surrogates (70--84\%/42--48\% with Llama 3.1-8B). 
        Ranking prompts by the surrogate’s $|\ell|$, on the other hand, does not raise this rate for \textit{any} of the twelve pairs.

    \textbf{Cross-benchmark results.}
        \cref{tab:cross_benchmark_results} lists results on ANLI, WinoGrande, and RACE benchmarks.
        While $\r$ and RV scores are not directly comparable, the qualitative ordering is stable across all benchmarks: $\Fpred$ generally exceeds $\Fattr$, attention is stable, and $\Fmag$ is more stable than $\Falign$.
        WinoGrande fidelity is lower across $\Fpred$ and $\Fattr$, and shows higher $\Fcross$ for mean- and max-pooled attention. 
        Surprisingly, most fidelity statistics do not decrease monotonically with ANLI difficulty (R1 $<$ R2 $<$ R3), suggesting they may not reflect task difficulty.

\section{Discussion}
\label{sec:discussion}

    \textbf{The access-validity inversion.} 
    Across seven evaluation settings, pooled $\Fpred > \Fattr$ in 360 out of 385 model-pair comparisons ($93.5\%$).
    This does not imply an invariant ranking of prediction and attribution agreement.
    Pooled prediction fidelity reflects both within-label covariance and inter-label separation; in input ablations, inter-label offsets cancel. 
    
    We find most mechanistic properties do not predict attributions well, and are outperformed by 
    API-compatible attribution metrics alone.
    In fact, median $\Frepr$ and $\Fcross$ are very poorly aligned (see \cref{fig:metric-calibration}), implying mechanistic transfer is largely uncorrelated with causal prediction.
    These negative results on attention add to the growing literature reexamining its explanatory value \citep{jain2019attention, wiegreffe2019attention, bibal2022attention}. 
    
    A plausible explanation for the access-validity inversion is that training forces models to align representations \textit{in high-probability subspaces}, but the null space of the top-$K$ readouts is less constrained.
    Consequently, models may agree on their overall predictions and even gross structural response ($\Fmag$) while pointing in very different directions.
    Interestingly, this divergence is detectable with only API-level access: you can't see \textit{how} these representations differ, but you can see \textit{that} they do.
    (Since we observe that log-probs diverge between true and false prompts only in instruction fine-tuned models (see \cref{sec:scaling}), it's even possible that the access-validity inversion is posttraining-specific.)
    
    \textbf{Surrogate fidelity as local linear geometry.} 
    The Platonic Representation Hypothesis \citep{huh2024platonic} holds that models converge to a shared representation, citing evidence such as model stitching, which unifies two models' representations via a learned linear transformation (similar to the $A$ discussed in \cref{eq:linear_transform}).
    Under this framing, our results are ironic: while \textit{appearances} do indeed converge (high $\Fpred$), the \textit{Forms} themselves actually fan out across different models (low $\Falign$). 
    
    The polytope lens relaxation \citep{black2022interpreting} permits models to shuffle operations within closed consistency regions, so low surrogate fidelity need not imply deep representational divergence. 
    This more closely matches how \citet{huh2024platonic} actually demonstrate their claims, eschewing the strict orthogonal transformation assumptions underpinning CKA in favor of nearest-neighbor overlaps.

    \textbf{Practical guidance.}
    We find that surrogate validity should be measured rather than assumed, and suggest a concrete way for practitioners to make surrogate validity claims.
    Prediction fidelity is a useful screen, but it can overstate agreement at the level needed for explanation:
    on BoolQ with word-level ablations, $\Fpred = .719$ whereas $\Fattr = .168$.
    Mechanistic similarities can also be misleading.
    For instance, $\Fmag$ and $\Fattn{}$ are high across model pairs, suggesting high similarity more generally, but $\Falign$ is consistently low, and so is $\Fattr$.

    The appropriate fidelity test depends on the level of access and the claim being made.
    Both $\Fpred$ and $\Fattr$ can be measured using log-probabilities alone, and should be reported whenever comparing models to make claims about surrogate explanations.
    For directly testing open-to-closed mechanism transfer, $\Fcross$ requires open-weight access to the surrogate.
    Validating at the level of the explanation being made and the behavior being tested will help ground claims about generalization to frontier models.
    Note that the intervention can and should be chosen to match the behavior of interest: for instance, leave-one-out ablations are useful for claims about input evidence or saliency, but may not be appropriate for circuit-level claims.
    
    \textbf{Limitations and future work.} 
    We offer a first step towards standardized surrogate fidelity evaluation and expect practices to evolve with input from the MI community.
    We focus on constrained classification tasks where log-odds give a reliable readout; open-ended text generation, which represents the vast majority of LLM use cases, is a natural next direction requiring new scoring functions.
    Our focus on leave-one-out attribution trades mechanistic depth for API compatibility;
    moreover, ablations shifting inputs off-distribution may have unintended side effects.
    It is therefore natural to extend our work to more commonly used \textit{mechanistic} interventions like activation patching, richer mechanistic signals (steering vectors, SAE features), and meaning-preserving input perturbations \citep{romanou2026brittlebenchquantifyingllmrobustness}.
    Low cross-seed consistency between SAE features \citep{chanin2024absorption, leask2025sparse} might imply similarly low mechanistic fidelity.

\section{Related Work}
\label{sec:related_work}
    \textbf{Feature attribution for LLMs.} 
    Leave-one-out attribution \citep{koh2017understanding}, ContextCite \citep{cohen2024contextcite}, and local surrogate methods such as MExGen \citep{paes2025multilevelexplanationsgenerativelanguage} and gSmILE \citep{dehghani2025explaininglargelanguagemodels} estimate input-level importance for generative models. 
    AttriBoT \citep{liu2025attribotbagtricksefficiently} scales attribution to long contexts via efficient approximation. 
    These methods only explain \emph{what} a model's prediction depends on; we connect attributions to internal mechanism by exploiting the linearity of log-odds. 
    In this regard, \citet{carlini2024stealing} is also influential to our work in its use of log-probabilities to probe model mechanisms. 
    
    \textbf{Proxy and surrogate models.} 
    Prior work asks whether proxy models can cheaply approximate a target's predictions \citep{hinton2015distilling} or attributions \citep{ribeiro2016should, koh2026predictingllmreasoningperformance}, often by training or fine-tuning a surrogate. 
    On the other hand, we ask when a surrogate-derived explanation is valid evidence about a target model's \emph{mechanism}.
    
    \textbf{Mechanistic interpretability.} 
    Circuit analysis \citep{olah2020zoom, elhage2021mathematical} has been scaled \citep{wang2022interpretability, conmy2023towards} and automated \citep{goldowsky2023localizing}; sparse autoencoders offer a complementary feature-level decomposition \citep{bricken2023towards, cunningham2023sparse, templeton2024scaling}. 
    Causally relevant directions---steering vectors \citep{turner2023steering, rimsky2024steering, li2023inference}, linear probes \citep{belinkov2022probing, hewitt2019structural}, crosscoder latents \citep{dunefsky2024transcoders, lindsey2024sparse, ameisen2025circuit}---identify features that influence behavior but require full model access. 
    We relax this requirement and test whether surrogate-derived features transfer to closed targets. 
    
    \textbf{Attention as explanation.} 
    Attention weights are widely used as explanations but their validity is contested \citep{jain2019attention, wiegreffe2019attention, serrano2019attention, bastings2020elephant, bibal2022attention}. 
    Our $F_{\mathrm{cross}}$ metric provides direct evidence: attention patterns are highly consistent across scales yet uncorrelated with causal attributions, so attention stability is not evidence of mechanistic agreement. 
    
    \textbf{Representation similarity.} CKA \citep{kornblith2019similarity}, SVCCA \citep{raghu2017svcca}, and RSA \citep{kriegeskorte2008representational} compare hidden representations across models. 
    The linear representation hypothesis \citep{park2024linearrepresentationhypothesisgeometry} and polytope lens \citep{black2022interpreting} characterize how models encode concepts geometrically.
    Our results partially agree: attention converges across scales, but its dissociation from attribution fidelity shows representational convergence does not imply functional equivalence.
    This split mirrors the work of \citet{klabunde2025similarity} and \citep{ding2021grounding}.

\section{Conclusion}
    We introduce a hierarchy of surrogate fidelity metrics grounded in the distinction between observational, causal, and structural agreement. 
    We find that neither prediction nor mechanistic fidelity implies attribution fidelity.
    When interpreting open surrogates as proxies for closed systems, we recommend that practitioners validate surrogates at the level of analysis they intend to perform, and release our evaluation framework to support this practice.
    More broadly, we offer an opinionated first proposal and invite the MI community to refine it: cross-model fidelity deserves to be discussed explicitly, and the right framework to measure it may still need to evolve.

\subsection*{AI use statement}
Most code was LLM-generated based on a human-written spec to carry out experiments scoped by humans. All code was reviewed for correctness and edited by humans. All figures were generated by LLM-based coding assistants, with significant user input and iteration. In particular, the authors directly verified that HuggingFace-based code for open models matched API call outputs for Llama models and verified all mathematical claims (e.g. attribution decomposition into magnitude, alignment, and normalization components) numerically to within floating-point precision.

At the conceptualization level, LLM input was used to select appropriate benchmark datasets and models, as well as to refine some of the mathematical intuitions presented in the text. 

LLMs were used to refine and edit text, but all of the text in the main body of the paper is human-generated and AI-edited. Some sections of the appendix include more LLM-generated text summarizing experimental procedures and results, but they have been read and verified by the authors.
Appendices \ref{app:robustness} and \ref{app:case-study} are primarily LLM-generated; they are human-checked and edited. 

\bibliography{references}

@book{pearl2009causality,
  title={Causality},
  author={Pearl, Judea},
  year={2009},
  publisher={Cambridge university press}
}

@inproceedings{abnar2020quantifying,
  title={Quantifying attention flow in transformers},
  author={Abnar, Samira and Zuidema, Willem},
  booktitle={Proceedings of the 58th annual meeting of the association for computational linguistics},
  pages={4190--4197},
  year={2020}
}

@inproceedings{clark2019boolq,
  title={Boolq: Exploring the surprising difficulty of natural yes/no questions},
  author={Clark, Christopher and Lee, Kenton and Chang, Ming-Wei and Kwiatkowski, Tom and Collins, Michael and Toutanova, Kristina},
  booktitle={Proceedings of the 2019 conference of the north American chapter of the association for computational linguistics: Human language technologies, volume 1 (long and short papers)},
  pages={2924--2936},
  year={2019}
}

@inproceedings{nie2020adversarial,
  title={Adversarial NLI: A new benchmark for natural language understanding},
  author={Nie, Yixin and Williams, Adina and Dinan, Emily and Bansal, Mohit and Weston, Jason and Kiela, Douwe},
  booktitle={Proceedings of the 58th annual meeting of the association for computational linguistics},
  pages={4885--4901},
  year={2020}
}

@article{sakaguchi2021winogrande,
  title={Winogrande: An adversarial winograd schema challenge at scale},
  author={Sakaguchi, Keisuke and Bras, Ronan Le and Bhagavatula, Chandra and Choi, Yejin},
  journal={Communications of the ACM},
  volume={64},
  number={9},
  pages={99--106},
  year={2021},
  publisher={ACM New York, NY, USA}
}

@inproceedings{jain2019attention,
  title={Attention is not explanation},
  author={Jain, Sarthak and Wallace, Byron C},
  booktitle={Proceedings of the 2019 Conference of the North American Chapter of the Association for Computational Linguistics: Human Language Technologies, Volume 1 (Long and Short Papers)},
  pages={3543--3556},
  year={2019}
}

@inproceedings{wiegreffe2019attention,
  title={Attention is not not explanation},
  author={Wiegreffe, Sarah and Pinter, Yuval},
  booktitle={Proceedings of the 2019 conference on empirical methods in natural language processing and the 9th international joint conference on natural language processing (EMNLP-IJCNLP)},
  pages={11--20},
  year={2019}
}

@article{templeton2024scaling,
   title={Scaling Monosemanticity: Extracting Interpretable Features from Claude 3 Sonnet},
   author={Templeton, Adly and Conerly, Tom and Marcus, Jonathan and Lindsey, Jack and Bricken, Trenton and Chen, Brian and Pearce, Adam and Citro, Craig and Ameisen, Emmanuel and Jones, Andy and Cunningham, Hoagy and Turner, Nicholas L and McDougall, Callum and MacDiarmid, Monte and Freeman, C. Daniel and Sumers, Theodore R. and Rees, Edward and Batson, Joshua and Jermyn, Adam and Carter, Shan and Olah, Chris and Henighan, Tom},
   year={2024},
   journal={Transformer Circuits Thread},
   url={https://transformer-circuits.pub/2024/scaling-monosemanticity/index.html}
}

@inproceedings{bibal2022attention,
  title={Is attention explanation? an introduction to the debate},
  author={Bibal, Adrien and Cardon, R{\'e}mi and Alfter, David and Wilkens, Rodrigo and Wang, Xiaoou and Fran{\c{c}}ois, Thomas and Watrin, Patrick},
  booktitle={Proceedings of the 60th Annual Meeting of the Association for Computational Linguistics (volume 1: long papers)},
  pages={3889--3900},
  year={2022}
}

@article{olah2020zoom,
  title={Zoom in: An introduction to circuits},
  author={Olah, Chris and Cammarata, Nick and Schubert, Ludwig and Goh, Gabriel and Petrov, Michael and Carter, Shan},
  journal={Distill},
  volume={5},
  number={3},
  pages={e00024--001},
  year={2020}
}

@article{elhage2021mathematical,
  title={A mathematical framework for transformer circuits},
  author={Elhage, Nelson and Nanda, Neel and Olsson, Catherine and Henighan, Tom and Joseph, Nicholas and Mann, Ben and Askell, Amanda and Bai, Yuntao and Chen, Anna and Conerly, Tom and others},
  journal={Transformer Circuits Thread},
  volume={1},
  number={1},
  pages={12},
  year={2021}
}

@article{wang2022interpretability,
  title={Interpretability in the wild: a circuit for indirect object identification in gpt-2 small},
  author={Wang, Kevin and Variengien, Alexandre and Conmy, Arthur and Shlegeris, Buck and Steinhardt, Jacob},
  journal={arXiv preprint arXiv:2211.00593},
  year={2022}
}

@article{conmy2023towards,
  title={Towards automated circuit discovery for mechanistic interpretability},
  author={Conmy, Arthur and Mavor-Parker, Augustine and Lynch, Aengus and Heimersheim, Stefan and Garriga-Alonso, Adri{\`a}},
  journal={Advances in Neural Information Processing Systems},
  volume={36},
  pages={16318--16352},
  year={2023}
}

@article{goldowsky2023localizing,
  title={Localizing model behavior with path patching},
  author={Goldowsky-Dill, Nicholas and MacLeod, Chris and Sato, Lucas and Arora, Aryaman},
  journal={arXiv preprint arXiv:2304.05969},
  year={2023}
}

@article{bricken2023towards,
  title={Towards monosemanticity: Decomposing language models with dictionary learning},
  author={Bricken, Trenton and Templeton, Adly and Batson, Joshua and Chen, Brian and Jermyn, Adam and Conerly, Tom and Turner, Nick and Anil, Cem and Denison, Carson and Askell, Amanda and others},
  journal={Transformer Circuits Thread},
  volume={2},
  number={5},
  pages={6},
  year={2023}
}

@article{cunningham2023sparse,
  title={Sparse autoencoders find highly interpretable features in language models},
  author={Cunningham, Hoagy and Ewart, Aidan and Riggs, Logan and Huben, Robert and Sharkey, Lee},
  journal={arXiv preprint arXiv:2309.08600},
  year={2023}
}

@inproceedings{serrano2019attention,
  title={Is attention interpretable?},
  author={Serrano, Sofia and Smith, Noah A},
  booktitle={Proceedings of the 57th annual meeting of the association for computational linguistics},
  pages={2931--2951},
  year={2019}
}

@inproceedings{bastings2020elephant,
  title={The elephant in the interpretability room: Why use attention as explanation when we have saliency methods?},
  author={Bastings, Jasmijn and Filippova, Katja},
  booktitle={Proceedings of the Third BlackboxNLP Workshop on Analyzing and Interpreting Neural Networks for NLP},
  pages={149--155},
  year={2020}
}

@inproceedings{ribeiro2016should,
  title={" Why should i trust you?" Explaining the predictions of any classifier},
  author={Ribeiro, Marco Tulio and Singh, Sameer and Guestrin, Carlos},
  booktitle={Proceedings of the 22nd ACM SIGKDD international conference on knowledge discovery and data mining},
  pages={1135--1144},
  year={2016}
}

@inproceedings{kornblith2019similarity,
  title={Similarity of neural network representations revisited},
  author={Kornblith, Simon and Norouzi, Mohammad and Lee, Honglak and Hinton, Geoffrey},
  booktitle={International conference on machine learning},
  pages={3519--3529},
  year={2019},
  organization={PMLR}
}

@article{raghu2017svcca,
  title={Svcca: Singular vector canonical correlation analysis for deep learning dynamics and interpretability},
  author={Raghu, Maithra and Gilmer, Justin and Yosinski, Jason and Sohl-Dickstein, Jascha},
  journal={Advances in neural information processing systems},
  volume={30},
  year={2017}
}

@article{kriegeskorte2008representational,
  title={Representational similarity analysis-connecting the branches of systems neuroscience},
  author={Kriegeskorte, Nikolaus and Mur, Marieke and Bandettini, Peter A},
  journal={Frontiers in systems neuroscience},
  volume={2},
  pages={249},
  year={2008},
  publisher={Frontiers}
}

@article{huh2024platonic,
  title={The platonic representation hypothesis},
  author={Huh, Minyoung and Cheung, Brian and Wang, Tongzhou and Isola, Phillip},
  journal={arXiv preprint arXiv:2405.07987},
  year={2024}
}

@article{hinton2015distilling,
  title={Distilling the knowledge in a neural network},
  author={Hinton, Geoffrey and Vinyals, Oriol and Dean, Jeff},
  journal={arXiv preprint arXiv:1503.02531},
  year={2015}
}

@article{turner2023steering,
  title={Steering language models with activation engineering},
  author={Turner, Alexander Matt and Thiergart, Lisa and Leech, Gavin and Udell, David and Vazquez, Juan J and Mini, Ulisse and MacDiarmid, Monte},
  journal={arXiv preprint arXiv:2308.10248},
  year={2023}
}

@inproceedings{rimsky2024steering,
  title={Steering llama 2 via contrastive activation addition},
  author={Rimsky, Nina and Gabrieli, Nick and Schulz, Julian and Tong, Meg and Hubinger, Evan and Turner, Alexander},
  booktitle={Proceedings of the 62nd Annual Meeting of the Association for Computational Linguistics (Volume 1: Long Papers)},
  pages={15504--15522},
  year={2024}
}

@article{li2023inference,
  title={Inference-time intervention: Eliciting truthful answers from a language model},
  author={Li, Kenneth and Patel, Oam and Vi{\'e}gas, Fernanda and Pfister, Hanspeter and Wattenberg, Martin},
  journal={Advances in Neural Information Processing Systems},
  volume={36},
  pages={41451--41530},
  year={2023}
}

@article{belinkov2022probing,
  title={Probing classifiers: Promises, shortcomings, and advances},
  author={Belinkov, Yonatan},
  journal={Computational Linguistics},
  volume={48},
  number={1},
  pages={207--219},
  year={2022}
}

@inproceedings{hewitt2019structural,
  title={A structural probe for finding syntax in word representations},
  author={Hewitt, John and Manning, Christopher D},
  booktitle={Proceedings of the 2019 Conference of the North American Chapter of the Association for Computational Linguistics: Human Language Technologies, Volume 1 (Long and Short Papers)},
  pages={4129--4138},
  year={2019}
}

@article{chanin2024absorption,
  title={A is for absorption: Studying feature splitting and absorption in sparse autoencoders},
  author={Chanin, David and Wilken-Smith, James and Dulka, Tom{\'a}{\v{s}} and Bhatnagar, Hardik and Golechha, Satvik and Bloom, Joseph},
  journal={arXiv preprint arXiv:2409.14507},
  year={2024}
}

@article{leask2025sparse,
  title={Sparse autoencoders do not find canonical units of analysis},
  author={Leask, Patrick and Bussmann, Bart and Pearce, Michael and Bloom, Joseph and Tigges, Curt and Moubayed, Noura Al and Sharkey, Lee and Nanda, Neel},
  journal={arXiv preprint arXiv:2502.04878},
  year={2025}
}

@article{dunefsky2024transcoders,
  title={Transcoders find interpretable llm feature circuits},
  author={Dunefsky, Jacob and Chlenski, Philippe and Nanda, Neel},
  journal={Advances in Neural Information Processing Systems},
  volume={37},
  pages={24375--24410},
  year={2024}
}

@article{kokhlikyan2020captum,
  title={Captum: A unified and generic model interpretability library for pytorch},
  author={Kokhlikyan, Narine and Miglani, Vivek and Martin, Miguel and Wang, Edward and Alsallakh, Bilal and Reynolds, Jonathan and Melnikov, Alexander and Kliushkina, Natalia and Araya, Carlos and Yan, Siqi and others},
  journal={arXiv preprint arXiv:2009.07896},
  year={2020}
}

@article{lindsey2024sparse,
  title={Sparse crosscoders for cross-layer features and model diffing},
  author={Lindsey, Jack and Templeton, Adly and Marcus, Jonathan and Conerly, Thomas and Batson, Joshua and Olah, Christopher},
  journal={Transformer Circuits Thread},
  pages={3982--3992},
  year={2024}
}

@article{ameisen2025circuit,
  author={Ameisen, Emmanuel and Lindsey, Jack and Pearce, Adam and Gurnee, Wes and Turner, Nicholas L. and Chen, Brian and Citro, Craig and Abrahams, David and Carter, Shan and Hosmer, Basil and Marcus, Jonathan and Sklar, Michael and Templeton, Adly and Bricken, Trenton and McDougall, Callum and Cunningham, Hoagy and Henighan, Thomas and Jermyn, Adam and Jones, Andy and Persic, Andrew and Qi, Zhenyi and Ben Thompson, T. and Zimmerman, Sam and Rivoire, Kelley and Conerly, Thomas and Olah, Chris and Batson, Joshua},
  title={Circuit Tracing: Revealing Computational Graphs in Language Models},
  journal={Transformer Circuits Thread},
  year={2025},
  url={https://transformer-circuits.pub/2025/attribution-graphs/methods.html}
}

@article{Nauta_2023,
   title={From Anecdotal Evidence to Quantitative Evaluation Methods: A Systematic Review on Evaluating Explainable AI},
   volume={55},
   ISSN={1557-7341},
   url={http://dx.doi.org/10.1145/3583558},
   DOI={10.1145/3583558},
   number={13s},
   journal={ACM Computing Surveys},
   publisher={Association for Computing Machinery (ACM)},
   author={Nauta, Meike and Trienes, Jan and Pathak, Shreyasi and Nguyen, Elisa and Peters, Michelle and Schmitt, Yasmin and Schlötterer, Jörg and van Keulen, Maurice and Seifert, Christin},
   year={2023},
   month=jul, pages={1–42}
}

@article{bai2023qwen,
  title={Qwen technical report},
  author={Bai, Jinze and Bai, Shuai and Chu, Yunfei and Cui, Zeyu and Dang, Kai and Deng, Xiaodong and Fan, Yang and Ge, Wenbin and Han, Yu and Huang, Fei and others},
  journal={arXiv preprint arXiv:2309.16609},
  year={2023}
}

@inproceedings{wolf2020transformers,
  title={Transformers: State-of-the-art natural language processing},
  author={Wolf, Thomas and Debut, Lysandre and Sanh, Victor and Chaumond, Julien and Delangue, Clement and Moi, Anthony and Cistac, Pierric and Rault, Tim and Louf, R{\'e}mi and Funtowicz, Morgan and others},
  booktitle={Proceedings of the 2020 conference on empirical methods in natural language processing: system demonstrations},
  pages={38--45},
  year={2020}
}

@misc{nostalgebraist2020logitlens,
  author = {nostalgebraist},
  title = {Interpreting {GPT}: The {Logit Lens}},
  year = {2020},
  url = {https://www.lesswrong.com/posts/AcKRB8wDpdaN6v6ru/interpreting-gpt-the-logit-lens},
  note = {Accessed: 2026-04-23}
}

@misc{belrose2025elicitinglatentpredictionstransformers,
      title={Eliciting Latent Predictions from Transformers with the Tuned Lens}, 
      author={Nora Belrose and Igor Ostrovsky and Lev McKinney and Zach Furman and Logan Smith and Danny Halawi and Stella Biderman and Jacob Steinhardt},
      year={2025},
      eprint={2303.08112},
      archivePrefix={arXiv},
      primaryClass={cs.LG},
      url={https://arxiv.org/abs/2303.08112}, 
}

@article{achiam2023gpt,
  title={Gpt-4 technical report},
  author={Achiam, Josh and Adler, Steven and Agarwal, Sandhini and Ahmad, Lama and Akkaya, Ilge and Aleman, Florencia Leoni and Almeida, Diogo and Altenschmidt, Janko and Altman, Sam and Anadkat, Shyamal and others},
  journal={arXiv preprint arXiv:2303.08774},
  year={2023}
}

@article{comanici2025gemini,
  title={Gemini 2.5: Pushing the frontier with advanced reasoning, multimodality, long context, and next generation agentic capabilities},
  author={Comanici, Gheorghe and Bieber, Eric and Schaekermann, Mike and Pasupat, Ice and Sachdeva, Noveen and Dhillon, Inderjit and Blistein, Marcel and Ram, Ori and Zhang, Dan and Rosen, Evan and others},
  journal={arXiv preprint arXiv:2507.06261},
  year={2025}
}

@article{carlini2024stealing,
  title={Stealing part of a production language model},
  author={Carlini, Nicholas and Paleka, Daniel and Dvijotham, Krishnamurthy Dj and Steinke, Thomas and Hayase, Jonathan and Cooper, A Feder and Lee, Katherine and Jagielski, Matthew and Nasr, Milad and Conmy, Arthur and others},
  journal={arXiv preprint arXiv:2403.06634},
  year={2024}
}

@article{black2022interpreting,
  title={Interpreting neural networks through the polytope lens},
  author={Black, Sid and Sharkey, Lee and Grinsztajn, Leo and Winsor, Eric and Braun, Dan and Merizian, Jacob and Parker, Kip and Guevara, Carlos Ram{\'o}n and Millidge, Beren and Alfour, Gabriel and others},
  journal={arXiv preprint arXiv:2211.12312},
  year={2022}
}

@inproceedings{koh2017understanding,
  title={Understanding black-box predictions via influence functions},
  author={Koh, Pang Wei and Liang, Percy},
  booktitle={International conference on machine learning},
  pages={1885--1894},
  year={2017},
  organization={PMLR}
}

@article{cohen2024contextcite,
  title={Contextcite: Attributing model generation to context},
  author={Cohen-Wang, Benjamin and Shah, Harshay and Georgiev, Kristian and Madry, Aleksander},
  journal={Advances in Neural Information Processing Systems},
  volume={37},
  pages={95764--95807},
  year={2024}
}

@misc{agrawal2026gepareflectivepromptevolution,
      title={GEPA: Reflective Prompt Evolution Can Outperform Reinforcement Learning}, 
      author={Lakshya A Agrawal and Shangyin Tan and Dilara Soylu and Noah Ziems and Rishi Khare and Krista Opsahl-Ong and Arnav Singhvi and Herumb Shandilya and Michael J Ryan and Meng Jiang and Christopher Potts and Koushik Sen and Alexandros G. Dimakis and Ion Stoica and Dan Klein and Matei Zaharia and Omar Khattab},
      year={2026},
      eprint={2507.19457},
      archivePrefix={arXiv},
      primaryClass={cs.CL},
      url={https://arxiv.org/abs/2507.19457}, 
}

@misc{paes2025multilevelexplanationsgenerativelanguage,
      title={Multi-Level Explanations for Generative Language Models}, 
      author={Lucas Monteiro Paes and Dennis Wei and Hyo Jin Do and Hendrik Strobelt and Ronny Luss and Amit Dhurandhar and Manish Nagireddy and Karthikeyan Natesan Ramamurthy and Prasanna Sattigeri and Werner Geyer and Soumya Ghosh},
      year={2025},
      eprint={2403.14459},
      archivePrefix={arXiv},
      primaryClass={cs.CL},
      url={https://arxiv.org/abs/2403.14459}, 
}

@misc{dehghani2025explaininglargelanguagemodels,
      title={Explaining Large Language Models with gSMILE}, 
      author={Zeinab Dehghani and Mohammed Naveed Akram and Koorosh Aslansefat and Adil Khan and Yiannis Papadopoulos},
      year={2025},
      eprint={2505.21657},
      archivePrefix={arXiv},
      primaryClass={cs.CL},
      url={https://arxiv.org/abs/2505.21657}, 
}

@misc{liu2025attribotbagtricksefficiently,
      title={AttriBoT: A Bag of Tricks for Efficiently Approximating Leave-One-Out Context Attribution}, 
      author={Fengyuan Liu and Nikhil Kandpal and Colin Raffel},
      year={2025},
      eprint={2411.15102},
      archivePrefix={arXiv},
      primaryClass={cs.LG},
      url={https://arxiv.org/abs/2411.15102}, 
}

@misc{koh2026predictingllmreasoningperformance,
      title={Predicting LLM Reasoning Performance with Small Proxy Model}, 
      author={Woosung Koh and Juyoung Suk and Sungjun Han and Se-Young Yun and Jamin Shin},
      year={2026},
      eprint={2509.21013},
      archivePrefix={arXiv},
      primaryClass={cs.LG},
      url={https://arxiv.org/abs/2509.21013}, 
}

@misc{park2024linearrepresentationhypothesisgeometry,
      title={The Linear Representation Hypothesis and the Geometry of Large Language Models}, 
      author={Kiho Park and Yo Joong Choe and Victor Veitch},
      year={2024},
      eprint={2311.03658},
      archivePrefix={arXiv},
      primaryClass={cs.CL},
      url={https://arxiv.org/abs/2311.03658}, 
}

@article{robert1976unifying,
  title={A unifying tool for linear multivariate statistical methods: the RV-coefficient},
  author={Robert, Paul and Escoufier, Yves},
  journal={Journal of the Royal Statistical Society Series C: Applied Statistics},
  volume={25},
  number={3},
  pages={257--265},
  year={1976},
  publisher={Oxford University Press}
}

@inproceedings{lai-etal-2017-race,
    title = "{RACE}: Large-scale {R}e{A}ding Comprehension Dataset From Examinations",
    author = "Lai, Guokun  and
      Xie, Qizhe  and
      Liu, Hanxiao  and
      Yang, Yiming  and
      Hovy, Eduard",
    editor = "Palmer, Martha  and
      Hwa, Rebecca  and
      Riedel, Sebastian",
    booktitle = "Proceedings of the 2017 Conference on Empirical Methods in Natural Language Processing",
    month = sep,
    year = "2017",
    address = "Copenhagen, Denmark",
    publisher = "Association for Computational Linguistics",
    url = "https://aclanthology.org/D17-1082/",
    doi = "10.18653/v1/D17-1082",
    pages = "785--794"
}

@article{Johnson1984ExtensionsOL,
  title={Extensions of Lipschitz mappings into Hilbert space},
  author={William B. Johnson and Joram Lindenstrauss},
  journal={Contemporary mathematics},
  year={1984},
  volume={26},
  pages={189-206},
  url={https://api.semanticscholar.org/CorpusID:117819162}
}

@misc{romanou2026brittlebenchquantifyingllmrobustness,
      title={Brittlebench: Quantifying LLM robustness via prompt sensitivity}, 
      author={Angelika Romanou and Mark Ibrahim and Candace Ross and Chantal Shaib and Kerem Oktar and Samuel J. Bell and Anaelia Ovalle and Jesse Dodge and Antoine Bosselut and Koustuv Sinha and Adina Williams},
      year={2026},
      eprint={2603.13285},
      archivePrefix={arXiv},
      primaryClass={cs.LG},
      url={https://arxiv.org/abs/2603.13285}, 
}

@article{klabunde2025similarity,
  title={Similarity of neural network models: A survey of functional and representational measures},
  author={Klabunde, Max and Schumacher, Tobias and Strohmaier, Markus and Lemmerich, Florian},
  journal={ACM Computing Surveys},
  volume={57},
  number={9},
  pages={1--52},
  year={2025},
  publisher={ACM New York, NY}
}

@article{ding2021grounding,
  title={Grounding representation similarity through statistical testing},
  author={Ding, Frances and Denain, Jean-Stanislas and Steinhardt, Jacob},
  journal={Advances in neural information processing systems},
  volume={34},
  pages={1556--1568},
  year={2021}
}

@misc{grattafiori2024llama3herdmodels,
      title={The Llama 3 Herd of Models}, 
      author={Aaron Grattafiori and Abhimanyu Dubey and Abhinav Jauhri and Abhinav Pandey and Abhishek Kadian and Ahmad Al-Dahle and Aiesha Letman and Akhil Mathur and Alan Schelten and Alex Vaughan and Amy Yang and Angela Fan and Anirudh Goyal and Anthony Hartshorn and Aobo Yang and Archi Mitra and Archie Sravankumar and Artem Korenev and Arthur Hinsvark and Arun Rao and Aston Zhang and Aurelien Rodriguez and Austen Gregerson and Ava Spataru and Baptiste Roziere and Bethany Biron and Binh Tang and Bobbie Chern and Charlotte Caucheteux and Chaya Nayak and Chloe Bi and Chris Marra and Chris McConnell and Christian Keller and Christophe Touret and Chunyang Wu and Corinne Wong and Cristian Canton Ferrer and Cyrus Nikolaidis and Damien Allonsius and Daniel Song and Danielle Pintz and Danny Livshits and Danny Wyatt and David Esiobu and Dhruv Choudhary and Dhruv Mahajan and Diego Garcia-Olano and Diego Perino and Dieuwke Hupkes and Egor Lakomkin and Ehab AlBadawy and Elina Lobanova and Emily Dinan and Eric Michael Smith and Filip Radenovic and Francisco Guzmán and Frank Zhang and Gabriel Synnaeve and Gabrielle Lee and Georgia Lewis Anderson and Govind Thattai and Graeme Nail and Gregoire Mialon and Guan Pang and Guillem Cucurell and Hailey Nguyen and Hannah Korevaar and Hu Xu and Hugo Touvron and Iliyan Zarov and Imanol Arrieta Ibarra and Isabel Kloumann and Ishan Misra and Ivan Evtimov and Jack Zhang and Jade Copet and Jaewon Lee and Jan Geffert and Jana Vranes and Jason Park and Jay Mahadeokar and Jeet Shah and Jelmer van der Linde and Jennifer Billock and Jenny Hong and Jenya Lee and Jeremy Fu and Jianfeng Chi and Jianyu Huang and Jiawen Liu and Jie Wang and Jiecao Yu and Joanna Bitton and Joe Spisak and Jongsoo Park and Joseph Rocca and Joshua Johnstun and Joshua Saxe and Junteng Jia and Kalyan Vasuden Alwala and Karthik Prasad and Kartikeya Upasani and Kate Plawiak and Ke Li and Kenneth Heafield and Kevin Stone and Khalid El-Arini and Krithika Iyer and Kshitiz Malik and Kuenley Chiu and Kunal Bhalla and Kushal Lakhotia and Lauren Rantala-Yeary and Laurens van der Maaten and Lawrence Chen and Liang Tan and Liz Jenkins and Louis Martin and Lovish Madaan and Lubo Malo and Lukas Blecher and Lukas Landzaat and Luke de Oliveira and Madeline Muzzi and Mahesh Pasupuleti and Mannat Singh and Manohar Paluri and Marcin Kardas and Maria Tsimpoukelli and Mathew Oldham and Mathieu Rita and Maya Pavlova and Melanie Kambadur and Mike Lewis and Min Si and Mitesh Kumar Singh and Mona Hassan and Naman Goyal and Narjes Torabi and Nikolay Bashlykov and Nikolay Bogoychev and Niladri Chatterji and Ning Zhang and Olivier Duchenne and Onur Çelebi and Patrick Alrassy and Pengchuan Zhang and Pengwei Li and Petar Vasic and Peter Weng and Prajjwal Bhargava and Pratik Dubal and Praveen Krishnan and Punit Singh Koura and Puxin Xu and Qing He and Qingxiao Dong and Ragavan Srinivasan and Raj Ganapathy and Ramon Calderer and Ricardo Silveira Cabral and Robert Stojnic and Roberta Raileanu and Rohan Maheswari and Rohit Girdhar and Rohit Patel and Romain Sauvestre and Ronnie Polidoro and Roshan Sumbaly and Ross Taylor and Ruan Silva and Rui Hou and Rui Wang and Saghar Hosseini and Sahana Chennabasappa and Sanjay Singh and Sean Bell and Seohyun Sonia Kim and Sergey Edunov and Shaoliang Nie and Sharan Narang and Sharath Raparthy and Sheng Shen and Shengye Wan and Shruti Bhosale and Shun Zhang and Simon Vandenhende and Soumya Batra and Spencer Whitman and Sten Sootla and Stephane Collot and Suchin Gururangan and Sydney Borodinsky and Tamar Herman and Tara Fowler and Tarek Sheasha and Thomas Georgiou and Thomas Scialom and Tobias Speckbacher and Todor Mihaylov and Tong Xiao and Ujjwal Karn and Vedanuj Goswami and Vibhor Gupta and Vignesh Ramanathan and Viktor Kerkez and Vincent Gonguet and Virginie Do and Vish Vogeti and Vítor Albiero and Vladan Petrovic and Weiwei Chu and Wenhan Xiong and Wenyin Fu and Whitney Meers and Xavier Martinet and Xiaodong Wang and Xiaofang Wang and Xiaoqing Ellen Tan and Xide Xia and Xinfeng Xie and Xuchao Jia and Xuewei Wang and Yaelle Goldschlag and Yashesh Gaur and Yasmine Babaei and Yi Wen and Yiwen Song and Yuchen Zhang and Yue Li and Yuning Mao and Zacharie Delpierre Coudert and Zheng Yan and Zhengxing Chen and Zoe Papakipos and Aaditya Singh and Aayushi Srivastava and Abha Jain and Adam Kelsey and Adam Shajnfeld and Adithya Gangidi and Adolfo Victoria and Ahuva Goldstand and Ajay Menon and Ajay Sharma and Alex Boesenberg and Alexei Baevski and Allie Feinstein and Amanda Kallet and Amit Sangani and Amos Teo and Anam Yunus and Andrei Lupu and Andres Alvarado and Andrew Caples and Andrew Gu and Andrew Ho and Andrew Poulton and Andrew Ryan and Ankit Ramchandani and Annie Dong and Annie Franco and Anuj Goyal and Aparajita Saraf and Arkabandhu Chowdhury and Ashley Gabriel and Ashwin Bharambe and Assaf Eisenman and Azadeh Yazdan and Beau James and Ben Maurer and Benjamin Leonhardi and Bernie Huang and Beth Loyd and Beto De Paola and Bhargavi Paranjape and Bing Liu and Bo Wu and Boyu Ni and Braden Hancock and Bram Wasti and Brandon Spence and Brani Stojkovic and Brian Gamido and Britt Montalvo and Carl Parker and Carly Burton and Catalina Mejia and Ce Liu and Changhan Wang and Changkyu Kim and Chao Zhou and Chester Hu and Ching-Hsiang Chu and Chris Cai and Chris Tindal and Christoph Feichtenhofer and Cynthia Gao and Damon Civin and Dana Beaty and Daniel Kreymer and Daniel Li and David Adkins and David Xu and Davide Testuggine and Delia David and Devi Parikh and Diana Liskovich and Didem Foss and Dingkang Wang and Duc Le and Dustin Holland and Edward Dowling and Eissa Jamil and Elaine Montgomery and Eleonora Presani and Emily Hahn and Emily Wood and Eric-Tuan Le and Erik Brinkman and Esteban Arcaute and Evan Dunbar and Evan Smothers and Fei Sun and Felix Kreuk and Feng Tian and Filippos Kokkinos and Firat Ozgenel and Francesco Caggioni and Frank Kanayet and Frank Seide and Gabriela Medina Florez and Gabriella Schwarz and Gada Badeer and Georgia Swee and Gil Halpern and Grant Herman and Grigory Sizov and Guangyi and Zhang and Guna Lakshminarayanan and Hakan Inan and Hamid Shojanazeri and Han Zou and Hannah Wang and Hanwen Zha and Haroun Habeeb and Harrison Rudolph and Helen Suk and Henry Aspegren and Hunter Goldman and Hongyuan Zhan and Ibrahim Damlaj and Igor Molybog and Igor Tufanov and Ilias Leontiadis and Irina-Elena Veliche and Itai Gat and Jake Weissman and James Geboski and James Kohli and Janice Lam and Japhet Asher and Jean-Baptiste Gaya and Jeff Marcus and Jeff Tang and Jennifer Chan and Jenny Zhen and Jeremy Reizenstein and Jeremy Teboul and Jessica Zhong and Jian Jin and Jingyi Yang and Joe Cummings and Jon Carvill and Jon Shepard and Jonathan McPhie and Jonathan Torres and Josh Ginsburg and Junjie Wang and Kai Wu and Kam Hou U and Karan Saxena and Kartikay Khandelwal and Katayoun Zand and Kathy Matosich and Kaushik Veeraraghavan and Kelly Michelena and Keqian Li and Kiran Jagadeesh and Kun Huang and Kunal Chawla and Kyle Huang and Lailin Chen and Lakshya Garg and Lavender A and Leandro Silva and Lee Bell and Lei Zhang and Liangpeng Guo and Licheng Yu and Liron Moshkovich and Luca Wehrstedt and Madian Khabsa and Manav Avalani and Manish Bhatt and Martynas Mankus and Matan Hasson and Matthew Lennie and Matthias Reso and Maxim Groshev and Maxim Naumov and Maya Lathi and Meghan Keneally and Miao Liu and Michael L. Seltzer and Michal Valko and Michelle Restrepo and Mihir Patel and Mik Vyatskov and Mikayel Samvelyan and Mike Clark and Mike Macey and Mike Wang and Miquel Jubert Hermoso and Mo Metanat and Mohammad Rastegari and Munish Bansal and Nandhini Santhanam and Natascha Parks and Natasha White and Navyata Bawa and Nayan Singhal and Nick Egebo and Nicolas Usunier and Nikhil Mehta and Nikolay Pavlovich Laptev and Ning Dong and Norman Cheng and Oleg Chernoguz and Olivia Hart and Omkar Salpekar and Ozlem Kalinli and Parkin Kent and Parth Parekh and Paul Saab and Pavan Balaji and Pedro Rittner and Philip Bontrager and Pierre Roux and Piotr Dollar and Polina Zvyagina and Prashant Ratanchandani and Pritish Yuvraj and Qian Liang and Rachad Alao and Rachel Rodriguez and Rafi Ayub and Raghotham Murthy and Raghu Nayani and Rahul Mitra and Rangaprabhu Parthasarathy and Raymond Li and Rebekkah Hogan and Robin Battey and Rocky Wang and Russ Howes and Ruty Rinott and Sachin Mehta and Sachin Siby and Sai Jayesh Bondu and Samyak Datta and Sara Chugh and Sara Hunt and Sargun Dhillon and Sasha Sidorov and Satadru Pan and Saurabh Mahajan and Saurabh Verma and Seiji Yamamoto and Sharadh Ramaswamy and Shaun Lindsay and Shaun Lindsay and Sheng Feng and Shenghao Lin and Shengxin Cindy Zha and Shishir Patil and Shiva Shankar and Shuqiang Zhang and Shuqiang Zhang and Sinong Wang and Sneha Agarwal and Soji Sajuyigbe and Soumith Chintala and Stephanie Max and Stephen Chen and Steve Kehoe and Steve Satterfield and Sudarshan Govindaprasad and Sumit Gupta and Summer Deng and Sungmin Cho and Sunny Virk and Suraj Subramanian and Sy Choudhury and Sydney Goldman and Tal Remez and Tamar Glaser and Tamara Best and Thilo Koehler and Thomas Robinson and Tianhe Li and Tianjun Zhang and Tim Matthews and Timothy Chou and Tzook Shaked and Varun Vontimitta and Victoria Ajayi and Victoria Montanez and Vijai Mohan and Vinay Satish Kumar and Vishal Mangla and Vlad Ionescu and Vlad Poenaru and Vlad Tiberiu Mihailescu and Vladimir Ivanov and Wei Li and Wenchen Wang and Wenwen Jiang and Wes Bouaziz and Will Constable and Xiaocheng Tang and Xiaojian Wu and Xiaolan Wang and Xilun Wu and Xinbo Gao and Yaniv Kleinman and Yanjun Chen and Ye Hu and Ye Jia and Ye Qi and Yenda Li and Yilin Zhang and Ying Zhang and Yossi Adi and Youngjin Nam and Yu and Wang and Yu Zhao and Yuchen Hao and Yundi Qian and Yunlu Li and Yuzi He and Zach Rait and Zachary DeVito and Zef Rosnbrick and Zhaoduo Wen and Zhenyu Yang and Zhiwei Zhao and Zhiyu Ma},
      year={2024},
      eprint={2407.21783},
      archivePrefix={arXiv},
      primaryClass={cs.AI},
      url={https://arxiv.org/abs/2407.21783}, 
}

@misc{qwen2025qwen25technicalreport,
      title={Qwen2.5 Technical Report}, 
      author={Qwen and : and An Yang and Baosong Yang and Beichen Zhang and Binyuan Hui and Bo Zheng and Bowen Yu and Chengyuan Li and Dayiheng Liu and Fei Huang and Haoran Wei and Huan Lin and Jian Yang and Jianhong Tu and Jianwei Zhang and Jianxin Yang and Jiaxi Yang and Jingren Zhou and Junyang Lin and Kai Dang and Keming Lu and Keqin Bao and Kexin Yang and Le Yu and Mei Li and Mingfeng Xue and Pei Zhang and Qin Zhu and Rui Men and Runji Lin and Tianhao Li and Tianyi Tang and Tingyu Xia and Xingzhang Ren and Xuancheng Ren and Yang Fan and Yang Su and Yichang Zhang and Yu Wan and Yuqiong Liu and Zeyu Cui and Zhenru Zhang and Zihan Qiu},
      year={2025},
      eprint={2412.15115},
      archivePrefix={arXiv},
      primaryClass={cs.CL},
      url={https://arxiv.org/abs/2412.15115}, 
}

@misc{sharkey2025openproblemsmechanisticinterpretability,
      title={Open Problems in Mechanistic Interpretability}, 
      author={Lee Sharkey and Bilal Chughtai and Joshua Batson and Jack Lindsey and Jeff Wu and Lucius Bushnaq and Nicholas Goldowsky-Dill and Stefan Heimersheim and Alejandro Ortega and Joseph Bloom and Stella Biderman and Adria Garriga-Alonso and Arthur Conmy and Neel Nanda and Jessica Rumbelow and Martin Wattenberg and Nandi Schoots and Joseph Miller and Eric J. Michaud and Stephen Casper and Max Tegmark and William Saunders and David Bau and Eric Todd and Atticus Geiger and Mor Geva and Jesse Hoogland and Daniel Murfet and Tom McGrath},
      year={2025},
      eprint={2501.16496},
      archivePrefix={arXiv},
      primaryClass={cs.LG},
      url={https://arxiv.org/abs/2501.16496}, 
}
\bibliographystyle{iclr2027_conference}

\newpage
\appendix
\onecolumn

\section{Experimental Details}
\label{sec:experimental-details}
    \paragraph{Models.}
    We evaluate the Qwen2.5 model family~\citep{bai2023qwen} at four scales: 0.5B, 3B, 7B, and 14B parameters, using the instruction-tuned variants (Qwen2.5-\{0.5B,3B,7B,14B\}-Instruct).
    For the scaling analysis in Section~\ref{sec:scaling}, we additionally evaluate the base (non-instruction-tuned) variants at the same scales plus 32B.
    All locally loaded models are loaded in \texttt{bfloat16} precision on a single NVIDIA A100 80\,GB GPU using HuggingFace Transformers~\citep{wolf2020transformers} with \texttt{device\_map="auto"}.
    Models are loaded and unloaded sequentially to fit within GPU memory.

    We call hosted models via APIs that expose top-$k$ log-probabilities.
    Due to capacity constraints, we treat Llama 3.1-70B and Llama 3.3-70B as closed models, calling them via such an API.

    \paragraph{Inference.}
    We extract logits from a single forward pass with no sampling---we do not generate text.
    For each prompt, we compute the log-softmax over the \emph{full vocabulary} at the last token position (i.e., the position where the model would begin generating).
    This differs from API-based approaches that return only the top-$k$ log-probabilities; our local inference provides exact log-probabilities for any token.
    Temperature is not applicable as we operate on raw logits without sampling.

    \paragraph{Log-odds scoring.}
    \label{sec:log_odds_details}
    For a prompt $x$ and label classes $\mathcal{P}$ (positive) and $\mathcal{N}$ (negative), we define the log-odds as:
    \begin{equation}
        \ell(x) = \log \sum_{t \in \mathcal{P}} P(t \mid x) - \log \sum_{t \in \mathcal{N}} P(t \mid x)
    \end{equation}
    where $P(t \mid x)$ is the next-token probability for token $t$.
    For BoolQ, $\mathcal{P}$ and $\mathcal{N}$ correspond to the tokens ``true'' and ``false'' respectively.
    To handle tokenizer variation, we expand each label to include case variants (\texttt{true}, \texttt{True}, \texttt{TRUE}) and prefix variants (no prefix, space prefix, underscore prefix), yielding 9 token variants per label.
    Log-probabilities across variants are aggregated via log-sum-exp.
    For WinoGrande, $\mathcal{P}$ is \texttt{1} and $\mathcal{N}$ is \texttt{2}.

    For multivariate classification ($K$ total classes), we collect the scores for every class and construct the vector of all unordered pairwise log-odds.
    Letting each answer's log-sum-exp aggregation be $s_k(x)=\log\sum_{t\in\mathcal{T}_k}P(t\mid x),$ we then collect the answers into log-odds vectors:
    \begin{equation}
        \boldsymbol{\ell}(x) = \left(
            s_a(x) - s_b(x)
        \right)_{1 \leq a < b \leq K}.
    \end{equation}  
    For ANLI, the three categories are ``entailment,'' ``neutral,'' and ``contradiction,'' corresponding to tokens \texttt{ent}, \texttt{neutral}, and \texttt{contr}, expanded analogously to BoolQ above.
    For RACE, the four categories are ``A,'' ``B,'' ``C,'' and ``D,'' with variant tokens considered. 

    For segment $i$, the multivariate attribution is
    \begin{equation}
      \boldsymbol{\Delta\ell}_i(x)
      =
      \boldsymbol{\ell}(x)
      -
      \boldsymbol{\ell}(x_{\setminus i}).
    \end{equation}

    \textbf{Missing values.} 
    A top-$K$ response can omit every variant of a requested label. 
    For the binary benchmarks (BoolQ, WinoGrande) we use pairwise complete cases: a prompt or segment enters a pair’s statistic only when both labels are observed for both models, before and after the deletion, and we report the resulting coverage (\cref{app:rob-missing}). 
    For the multi-class benchmarks (ANLI, RACE), each omitted score is replaced by
    \begin{equation}
        \label{eq:imputed_floor}
        m - (0.01|m| + 0.01),
    \end{equation}
    where $m$ is the minimum log-probability for that model and benchmark. Rows with no observed label are dropped in both cases.

    \paragraph{Prompt templates.}
    Each benchmark uses a task-specific system prompt and a structured user prompt.
    For BoolQ:
    \begin{quote}
    \small
    \textbf{System:} ``You are a general-purpose binary question-answering machine. You will be shown a passage (marked `Context'), followed by a true/false question (marked `Question') about that passage. Answer the question on the basis of the context, and respond only with `true' or `false'.'' \\[4pt]
    \textbf{User:} ``Context: \{passage\}. Question: \{question\}.''
    \end{quote}
    ANLI uses ``Premise'' / ``Hypothesis'' labels with an entailment/neutral/contradiction instruction; WinoGrande uses ``Sentence'' / ``Option 1'' / ``Option 2'' with a blank-filling instruction.
    Prompts are formatted using each model's native chat template via \texttt{tokenizer.apply\_chat\_template()}.

    \paragraph{Segmentation.}
    We segment prompts using Captum's InterpretableInput framework~\citep{kokhlikyan2020captum}.
    For sentence-level experiments, we use the \texttt{"sentence"} TextSegmentInput; for word-level experiments, the \texttt{"word"} TextSegmentInput.
    Segmentation operates on both the system and user message components of the context.

    \paragraph{Ablation scoring.}
    For each prompt, we perform leave-one-out ablation: for each of $N$ segments, we remove that segment and re-score the modified prompt.
    The ablation response for segment $i$ is:
    \begin{equation}
        a_i(x) = \ell(x) - \ell(x_{\setminus i})
    \end{equation}
    where $x_{\setminus i}$ is the prompt with segment $i$ removed.
    This requires $N + 1$ forward passes per prompt (one for the original, one per ablation).

    \paragraph{Attention scoring.}
    We extract attention weights from a single forward pass with \texttt{output\_attentions=True}.
    Each transformer layer $l$ produces an attention matrix $A^{(l)} \in \mathbb{R}^{H \times T \times T}$ where $H$ is the number of heads and $T$ is the sequence length.
    We consider three aggregation strategies:

    \begin{itemize}[nosep]
        \item \textbf{Mean pooling:} Average over all heads and layers, then extract the last row (attention from the prediction position to all input tokens).
        \item \textbf{Max pooling:} Maximum over all heads and layers, then extract the last row.
        \item \textbf{Attention rollout}~\citep{abnar2020quantifying}: At each layer, average over heads, mix 50/50 with the identity matrix (accounting for residual connections), renormalize rows, and multiply into a running product. The result captures attention flow through the full network.
    \end{itemize}

    Token-level attention is aggregated to segment-level scores by summing attention weights over the tokens belonging to each segment (mapped via tokenizer offset positions).

    \paragraph{Agreement estimation and uncertainty.}
    For scalar tasks, prediction fidelity is the squared Pearson correlation between models' prompt-level log-odds:
    \begin{equation}
        r^2(\mathbf{x},\mathbf{y})
        = \operatorname{corr}(\mathbf{x},\mathbf{y})^2.
    \end{equation}
    Letting $\mathbf{x}_c=\mathbf{x}-\bar{x}\mathbf{1}$ and $\mathbf{y}_c=\mathbf{y}-\bar{y}\mathbf{1}$, this expands as
    \begin{equation}
      r^2(\mathbf{x},\mathbf{y})
      =
      \frac{
          \left(\mathbf{x}_c^\top\mathbf{y}_c\right)^2
      }{
          \left(\mathbf{x}_c^\top\mathbf{x}_c\right)
          \left(\mathbf{y}_c^\top\mathbf{y}_c\right)
      }.
    \end{equation}

    Generalizing to multiclass signals with centered $\mathbf{X_c}$ and $\mathbf{Y_c}$, we define the RV coefficient as
    \begin{equation}
    \label{eq:rv}
      \operatorname{RV}(\mathbf{X},\mathbf{Y})
      =
      \frac{\| \mathbf{X_c}^\top \mathbf{Y_c}\|_F^2}
      {\sqrt{
          \| \mathbf{X_c}^\top \mathbf{X_c}\|_F^2
          \| \mathbf{Y_c}^\top \mathbf{Y_c}\|_F^2
      }}.
    \end{equation}
    
    Segment-level metrics pool all jointly observed $(x,i)$ coordinates across the corpus and compute a single global correlation; we do not average correlations computed separately within each prompt. 
    Thus, prompts with more segments contribute proportionally more observations to the point estimate.
    For multiclass tasks, each observation remains a vector of pairwise label margins, and we compute centered RV between the resulting row-by-margin matrices. 
    
    We estimate 95\% confidence intervals using $1{,}000$ prompt-cluster bootstrap resamples: prompts are sampled with replacement and every coordinate belonging to a sampled prompt is retained.
  
    \paragraph{Benchmarks.}
    We evaluate on four benchmarks:
    BoolQ~\citep{clark2019boolq} (3,270 validation examples),
    ANLI~\citep{nie2020adversarial} R1/R2/R3 (1,000/1,000/1,200 test examples),
    WinoGrande~\citep{sakaguchi2021winogrande} (1,267 validation examples), and
    RACE~\citep{lai-etal-2017-race} (4,934 prompts).

    \paragraph{Compute.}
    All experiments are run on a single NVIDIA A100 80\,GB GPU.
    Sentence-level scoring (attention + ablation) for one benchmark $\times$ four models takes approximately 2--4 hours depending on dataset size.
    Word-level ablation is substantially more expensive (${\sim}100\times$ more forward passes) and runs for approximately 24 hours per benchmark.

\FloatBarrier
\section{Additional results}
\label{app:additional_results}

\subsection{Detailed mechanistic agreement heatmap}
\begin{figure}[ht]
    \centering
    \includegraphics[width=0.8\linewidth]{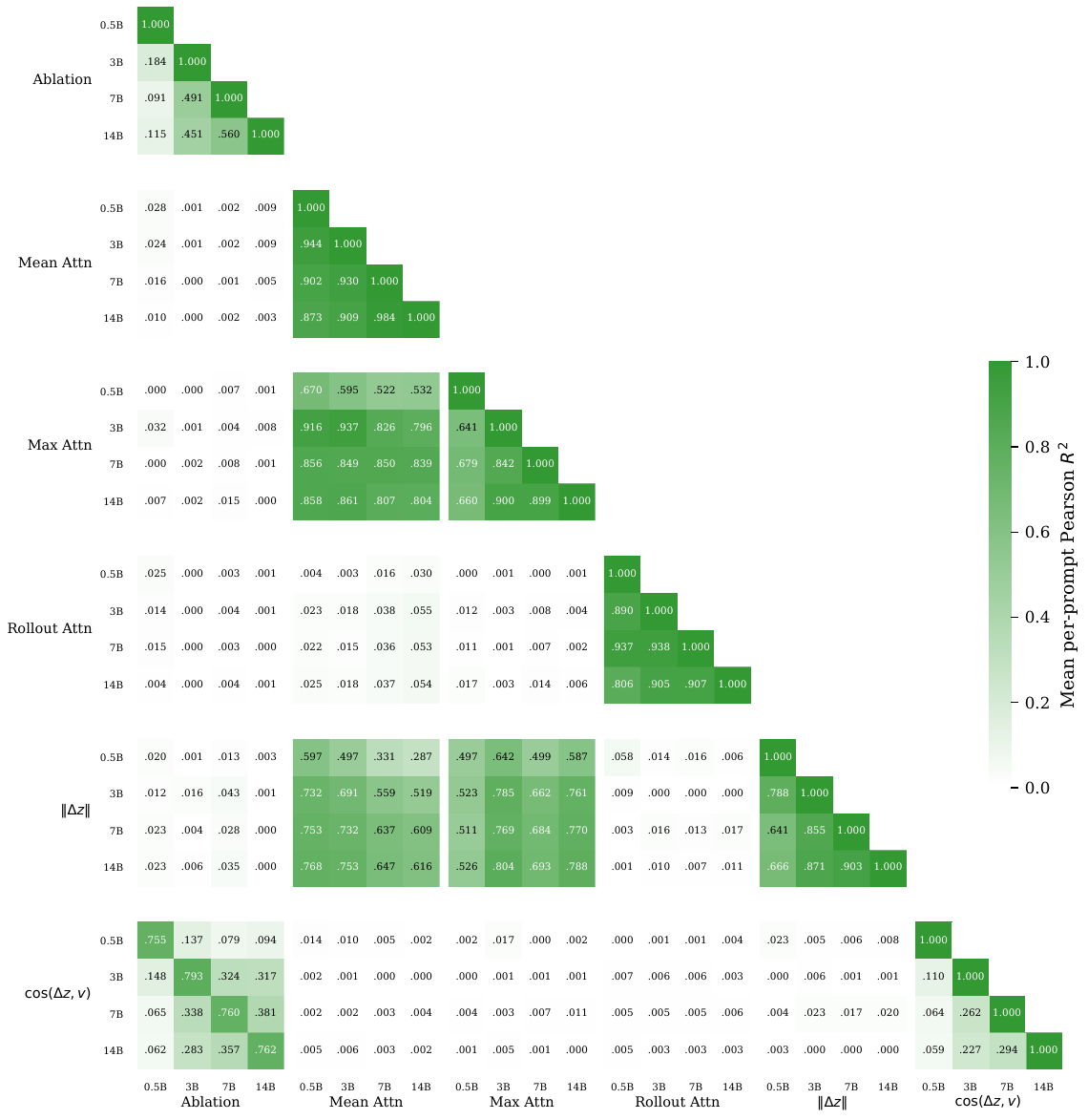}
    \caption{
        Fidelity metrics across a variety of measurements for Qwen 2.5 models with 0.5B, 3B, 7B, and 14B parameters.
        We measure attributions under ablation; mean-, max-, and rollout-pooled attention scores; perturbation norms; and perturbation alignment---all on BoolQ at the sentence level.
        We find that ablation scores are generally hard to predict, even using surrogate models of different sizes; predicting ablation scores using any attention-based metric is unreliable.
    }
    \label{fig:sentence_heatmap}
\end{figure}

A detailed heatmap of attention scoring methods and their agreement with $\Fattr$ as cross-predictors is given in \cref{fig:sentence_heatmap}.

\subsection{Output log-odds scaling in Qwen}
\label{sec:scaling}
\begin{figure}[ht]
    \centering
    \includegraphics[width=0.8\linewidth]{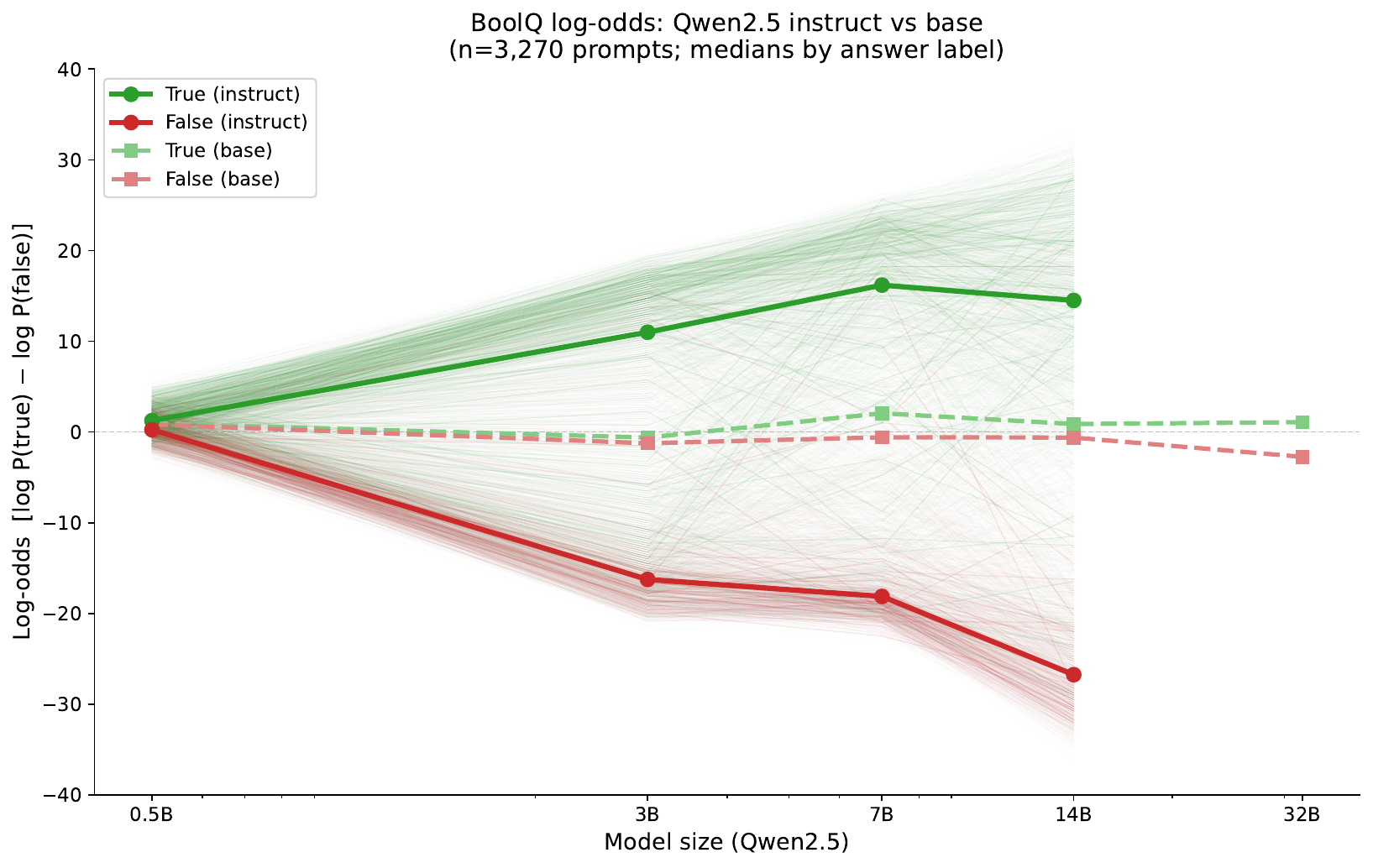}
    \caption{
        Median log-odds by ground-truth label for instruct (solid circles) and base (dashed squares) models, with individual prompt trajectories shown as translucent lines.
        Instruction-tuned models exhibit increasing class separation with scale, while base models remain near zero regardless of size.
        This suggests that the prediction signal measured by $F_{\mathrm{pred}}$ is largely a product of instruction tuning rather than pre-training alone.
    }
    \label{fig:qwen_logodds_scaling}
\end{figure}

We plot the log-odds of 100 randomly-selected prompts over different Qwen model sizes, as well as the true- and false-class medians, in \cref{fig:qwen_logodds_scaling}.
We repeat this process for base and instruct models, and find that instruct fine-tuning appears to drive log-odds differentiation in models.

\subsection{Base log-odds distribution}
\begin{figure}[ht]
    \centering
    \includegraphics[width=\linewidth]{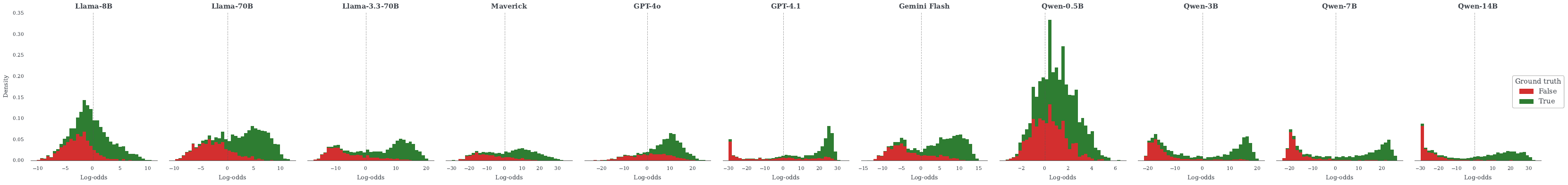}
    \caption{
        \textbf{Distribution of model log-odds on BoolQ by ground-truth label.}
        Each panel shows stacked histograms of log-odds for a single model, with
        green = \textsc{True} and red = \textsc{False}.
        Models above 3B parameters produce bimodal distributions with clear class separation.
        Qwen-0.5B exhibits substantial overlap between classes, consistent with its low prediction fidelity.
        The scale of log-odds varies widely across models (e.g., $|\ell| > 30$ for GPT-4.1 vs.\ $|\ell| < 10$ for Llama-8B), motivating the use of rank correlation.
    }
    \label{fig:histograms}
\end{figure}

In \cref{fig:histograms}, we plot histograms of base log-odds distributions for different models on the BoolQ dataset.
Histogram bars are subdivided by the proportion of true and false responses (ground truth) in the corresponding bin.
Note that Qwen-0.5B has a highly irregular distribution compared to the rest of the models; this is a likely explanation for why it has extremely low $\Fpred$ with all other models.

\subsection{RACE heatmap}
\begin{figure}[ht]
    \centering
    \includegraphics[width=0.45\linewidth]{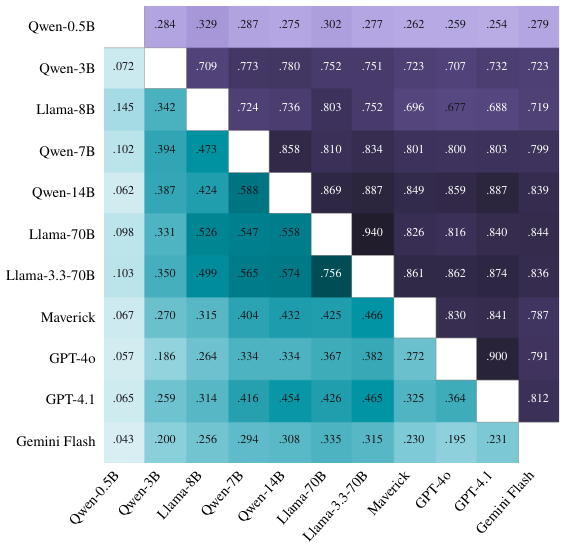}
    \caption{
        \textbf{RACE cross-model Multivariate RV fidelity scores.}
        Pairwise $F_\mathrm{pred}$ and $F_\mathrm{attr}$ heatmaps for eleven models across four families.
    }
    \label{fig:heatmap_race}
\end{figure}

In \cref{fig:heatmap_race}, we show attribution and fidelity for RACE. 
Comparing this figure with \cref{fig:dashboard}'s BoolQ shows generally comparable dynamics across the binary and multivariate classification settings.

\FloatBarrier
\section{Additional per-layer fidelity trajectories}
\label{sec:per_layer}

We extend the surrogate-fidelity framework of \cref{sec:method} from the
final layer to every layer of the residual stream, asking at \emph{which
depth} the cross-model agreement on each fidelity level emerges. Throughout,
we use the \textit{logit lens}~\citep{nostalgebraist2020logitlens} convention
introduced in \cref{sec:preliminaries}: for every $l \in \{0, 1, \ldots, L\}$
we read $\ell^{(l)}(x) := \operatorname{RMSNorm}(h^{(l)}) \cdot v$ where index
$0$ is the embedding-layer projection $\operatorname{RMSNorm}(h^{(0)}) \cdot v$ and
indices $1, \ldots, L$ are the post-block residuals.\footnote{We restrict
the per-layer extension to single-token positive/negative labels so that
the linearization $\ell = z \cdot v$ holds exactly
modulo bit-precision noise, allowing $\ell^{(l)}$ to be read directly from
$\operatorname{RMSNorm}(h^{(l)})$ without instantiating $W_U$ at every layer. We
verified this identity on Qwen-0.5B for all $L = 24$ decoder layers: the
maximum discrepancy between $\operatorname{RMSNorm}(h^{(l)}) \cdot v$ and the explicit
$W_U[\textsc{pos}] - W_U[\textsc{neg}]$ projection of $\operatorname{RMSNorm}(h^{(l)})$
is $9.6 \times 10^{-2}$ versus a five-bf16-ULP threshold of $2.4 \times
10^{-1}$. The analogous DLA-decomposition identity from \cref{eq:dla} holds
to $8.6 \times 10^{-2}$.} Per-layer arrays therefore have length $L + 1$.
We use the open-weight subset of the model pool (Qwen-2.5 \{0.5B, 3B, 7B,
14B\}-Instruct and Llama-3.1-8B-Instruct), the BoolQ validation split
($n = 500$ prompts), and the same sentence-level segmenter as
\cref{sec:experiments}, yielding $2{,}105$ $(\textsc{prompt},
\textsc{segment})$ pairs that all five models share.

\subsection{Trajectories of $\ell^{(l)}$ and the lens distribution}
\label{sec:per_layer:trajectories}

\begin{figure}[ht]
    \centering
    \begin{subfigure}[t]{.49\linewidth}
        \centering
        \includegraphics[width=\linewidth]{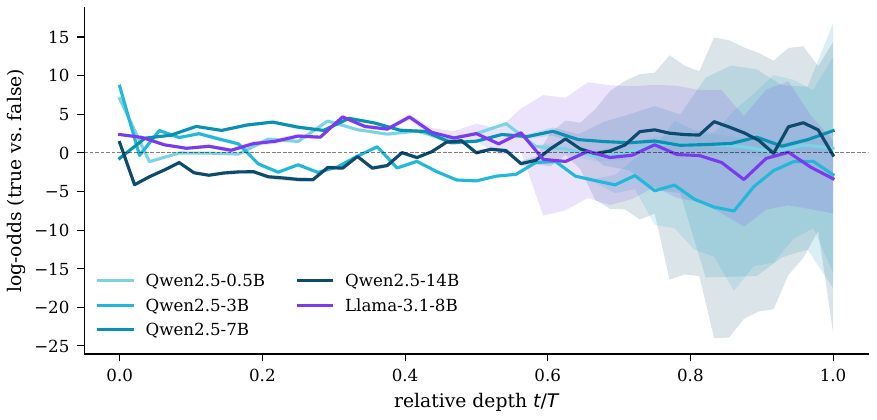}
        \caption{Median $\ell^{(l)}$ (with IQR ribbons) per relative depth
    $l / L$ across $n = 500$ BoolQ prompts. Trajectories converge through
    relative depths $0.1$--$0.5$ and re-diverge in the back half.}
        \label{fig:per_layer:trajectories}
    \end{subfigure}%
    \hfill
    \begin{subfigure}[t]{.49\linewidth}
        \centering
        \includegraphics[width=\linewidth]{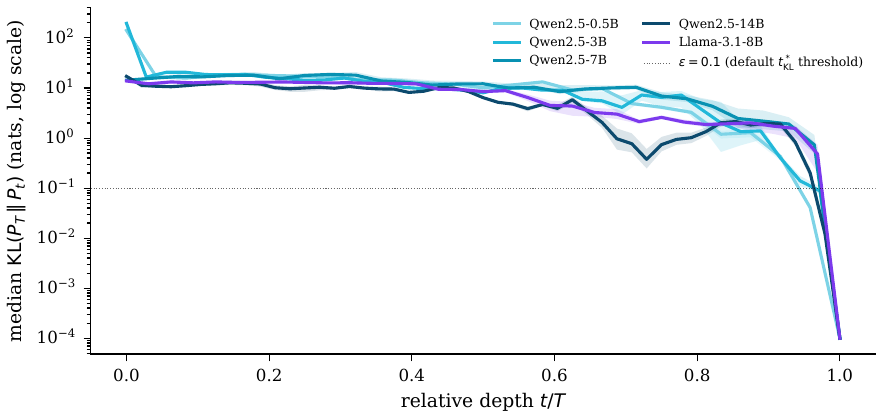}
        \caption{Median $\mathrm{KL}\big(P^{(L)} \,\|\, P^{(l)}\big)$ per
    relative depth, where $P^{(l)} = \mathrm{softmax}(W_U^\top
    \operatorname{RMSNorm}(h^{(l)}))$ is the lens distribution at layer $l$. The
    final-layer distribution is reached only in the last 1--3 layers
    under any reasonable threshold (the dotted line marks $\varepsilon
    = 0.1$~nats).}
        \label{fig:per_layer:kl}
    \end{subfigure}
    \caption{Per-layer trajectories and KL.}
\end{figure}

\begin{figure}[ht]
    \centering
    \includegraphics[width=.6\linewidth]{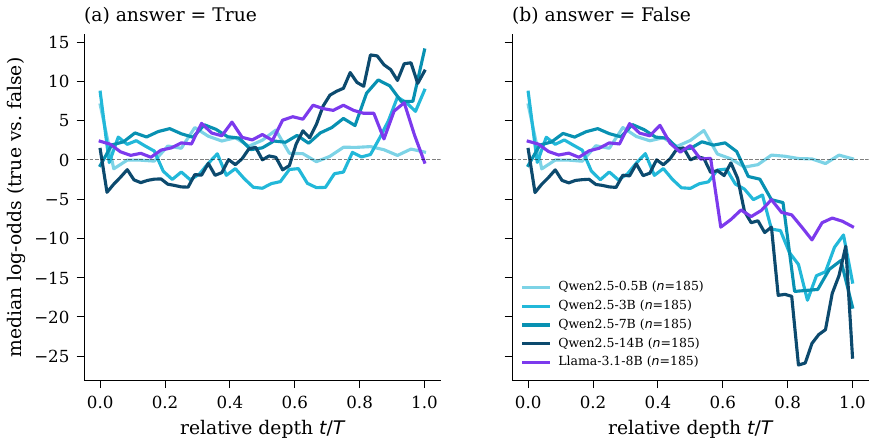}
    \caption{Same as \cref{fig:per_layer:trajectories} but split by
    ground-truth answer. The expected sign separation appears almost
    entirely in the second half of depth and is cleanest for Llama-3.1-8B
    and Qwen-14B.}
    \label{fig:per_layer:trajectories_by_answer}
\end{figure}

\Cref{fig:per_layer:trajectories} plots the per-layer trajectories of
$\ell^{(l)}$. All five models start with a substantial embedding-layer
bias whose sign and magnitude depend on the model's static embedding of
the assistant-marker token at the prediction position; readers should
treat the layer-$0$ slot as a per-tokenizer baseline rather than a
content-dependent prediction. \Cref{fig:per_layer:trajectories_by_answer}
splits the same trajectories by the ground-truth answer: the expected
sign separation between the True- and False-prompt sub-populations
emerges almost entirely in the back half of depth.
\Cref{fig:per_layer:kl} summarizes the same data as the per-layer
$\mathrm{KL}\big(P^{(L)} \,\|\, P^{(l)}\big)$ — the lens distribution
falls within $0.1$~nats of the final-layer distribution only in the
last 1--3 layers for every model.

\subsection{Per-layer attribution decomposition}
\label{sec:per_layer:attribution}

\begin{figure}[ht]
    \centering
    \begin{subfigure}[t]{.49\linewidth}
        \centering
        \includegraphics[width=\linewidth]{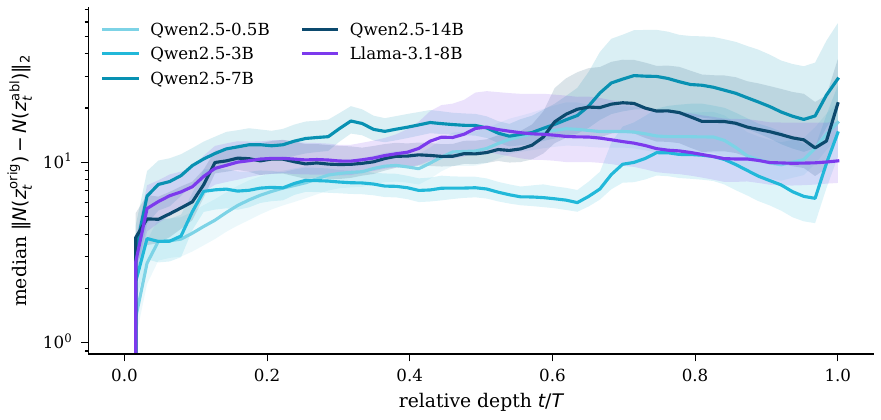}
        \caption{Median $\|\Delta z^{(l)}\|$ per relative depth. The
        structural footprint of an ablation appears almost immediately
        (depth $\lesssim 0.1$) and stays roughly flat thereafter.}
        \label{fig:per_layer:dnorm}
    \end{subfigure}%
    \hfill
    \begin{subfigure}[t]{.49\linewidth}
        \centering
        \includegraphics[width=\linewidth]{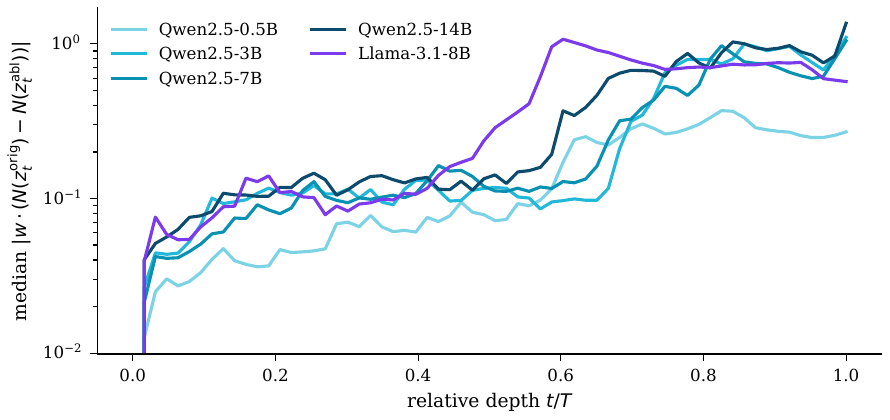}
        \caption{Median $|\Delta\ell^{(l)}|$ per relative depth. The
        readout-aligned component is small (${\sim}0.05$--$0.1$) for the
        first half of depth and grows by 1--2 orders of magnitude in the
        last 30\% of layers.}
        \label{fig:per_layer:wdotdz}
    \end{subfigure}
    \caption{Per-layer structural perturbation norm and readout-aligned attribution magnitude.}
    \label{fig:per_layer:dnorm_and_wdotdz}
\end{figure}

\begin{figure}[ht]
    \centering
    \begin{subfigure}[t]{.49\linewidth}
        \centering
        \includegraphics[width=\linewidth]{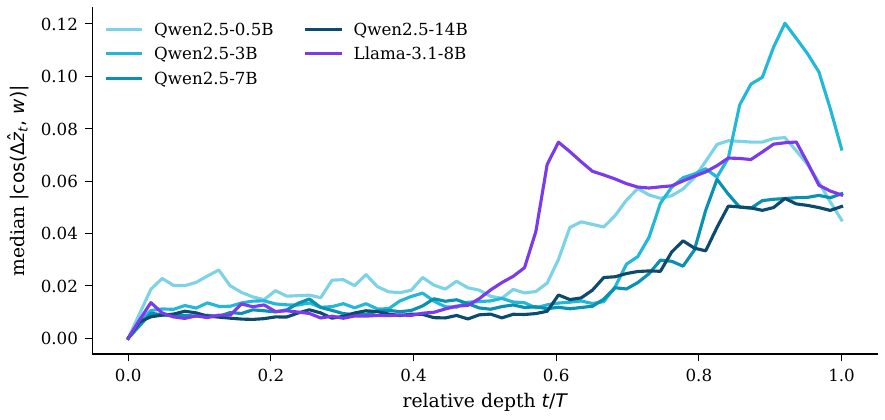}
        \caption{Median $|\cos(\Delta z^{(l)}, v)|$ per relative depth.
        Comparing with \cref{fig:per_layer:dnorm,fig:per_layer:wdotdz} shows
        that the late rise in $|\Delta\ell^{(l)}|$ is driven by the rising
        $|\cos|$ factor of \cref{eq:factorize_attribution}, not by larger
        perturbations.}
        \label{fig:per_layer:cos}
    \end{subfigure}%
    \hfill
    \begin{subfigure}[t]{.49\linewidth}
        \centering
        \includegraphics[width=\linewidth]{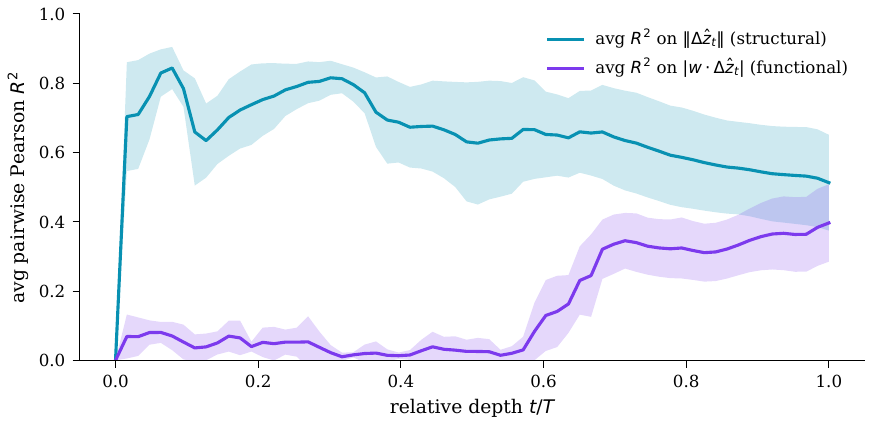}
        \caption{Pair-averaged $\Fmag^{(l)}$ (cyan, structural) and
        $\Fattr^{(l)}$ (purple, functional) versus relative depth, with
        \,$\pm 1\sigma$ ribbons across the $10$ model pairs.}
        \label{fig:per_layer:pair_average}
    \end{subfigure}
    \caption{Per-layer directional alignment and pair-averaged attribution magnitudes.}
    \label{fig:per_layer:cos_and_pair_average}
\end{figure}

Following the factorization in \cref{eq:factorize_attribution} we record
three per-$(\textsc{prompt}, \textsc{segment}, \textsc{layer})$ scalars
under the same leave-one-out segment ablation as \cref{sec:method}:
$\|\Delta z^{(l)}\|$, $\Delta\ell^{(l)} = \Delta z^{(l)} \cdot v$, and
$\cos(\Delta z^{(l)}, v)$, where $\Delta z^{(l)} :=
\operatorname{RMSNorm}(h^{(l)}(x)) - \operatorname{RMSNorm}(h^{(l)}(\tilde{x}))$.
\Cref{fig:per_layer:dnorm,fig:per_layer:wdotdz,fig:per_layer:cos} show
the per-model medians.

\subsection{Cross-model fidelity at depth}
\label{sec:per_layer:fidelity}

We compute the per-layer extensions $\Fmag^{(l)}$ and $\Fattr^{(l)}$ of
the metrics defined in \cref{eq:f_mag,eq:f_attr} on the $2{,}105$
$(\textsc{prompt}, \textsc{segment})$ pairs shared by all five models.
Per-model trajectories are resampled to a common length-$64$ grid in
relative depth $l / L \in [0, 1]$ via linear interpolation in $l / L$
space; pairwise Pearson $r^2$ (the coefficient of determination) is
computed at each of the $64$ grid points, i.e. the squared Pearson
correlation between the two models' per-segment attribution vectors at
that depth.

\begin{figure}[ht]
    \centering
    \begin{subfigure}[t]{.49\linewidth}
        \centering
        \includegraphics[width=\linewidth]{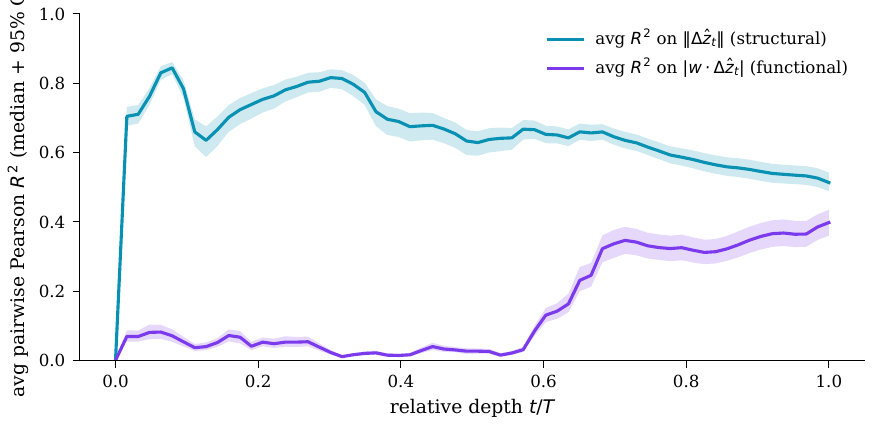}
        \caption{Bootstrap-CI version of \cref{fig:per_layer:pair_average}
        ($B = 500$ resamples of $(\textsc{prompt}, \textsc{segment})$ pairs
        with replacement; the $95\%$ CI is the empirical $2.5$/$97.5$
        percentile of the pair-averaged $r^2$ at each depth bin). CIs are
        $\leq \pm 0.05$ in $r^2$ at every depth.}
        \label{fig:per_layer:bootstrap}
    \end{subfigure}%
    \hfill
    \begin{subfigure}[t]{.49\linewidth}
        \centering
        \includegraphics[width=\linewidth]{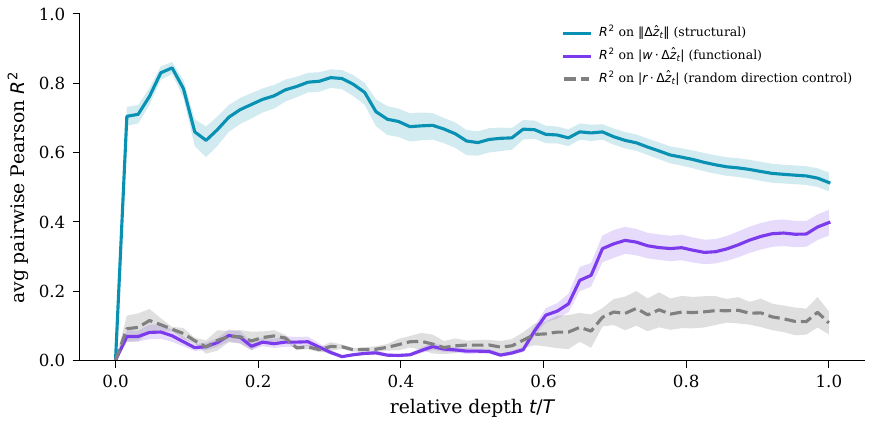}
        \caption{The structural and functional pair-averaged curves of
        \cref{fig:per_layer:bootstrap} with the random-direction control
        overlay (gray dashed). The control rises late too, plateauing at
        $r^2 \approx 0.13$--$0.15$ before ending at $r^2 \approx 0.11$
        at the readout. The $v$-specific premium over the random baseline
        at the readout is therefore $\approx 0.29$ in $r^2$.}
        \label{fig:per_layer:random_dir}
    \end{subfigure}
    \caption{Per-layer correlation analysis with bootstrap confidence intervals and random-direction control.}
    \label{fig:per_layer:bootstrap_and_random}
\end{figure}

\Cref{fig:per_layer:pair_average,fig:per_layer:bootstrap} aggregate over
the $10$ model pairs. The headline pattern at this scale is a clean
\emph{depth dissociation}:
\begin{itemize}[leftmargin=*,nosep]
    \item Structural agreement $\Fmag^{(l)}$ is high (${\sim}0.65$--$0.85$)
    across most depths past the embedding slot, then tapers to
    ${\sim}0.51$ at the readout.
    \item Functional agreement $\Fattr^{(l)}$ is low (${\lesssim}0.08$) for
    the first $\sim\!55\%$ of depth, rises sharply between depths
    $0.55$ and $0.7$, and reaches ${\sim}0.40$ at the readout.
    \item The final-layer $\Fattr^{(L)} \approx 0.40$ matches the
    main-paper $\Fattr$ baseline (Table~\ref{tab:f-summary}, BoolQ
    Open$\to$Open) to within sampling noise.
\end{itemize}

\paragraph{Random-direction control.}
The dissociation in \cref{fig:per_layer:bootstrap} could in principle
be a generic property of late-layer perturbations: any fixed direction
in residual-stream space might exhibit a low-then-high cross-model
agreement curve. To test this, we replace $v$ in $\Fattr^{(l)}$'s
definition with a per-model random unit direction $r$ matched in norm
to $\|v\|$ (averaging over $K = 5$ random draws per model on a
$200$-prompt subsample). \Cref{fig:per_layer:random_dir} overlays the
random-direction curve on the structural and functional curves of
\cref{fig:per_layer:bootstrap}.

The random-direction control rises late as well, plateauing at
$r^2 \approx 0.13$--$0.15$ before ending at $r^2 \approx 0.11$ at the
readout — a non-negligible but minority fraction of the headline
$\Fattr^{(L)} \approx 0.40$. The $v$-specific premium over the random
baseline at the readout layer is therefore $\approx 0.29$ in $r^2$.
We read this as a partial qualification of the framework's main-text
claim that low $\Fattr$ reflects models ``disagreeing on which inputs
drive their predictions'': of the cross-model agreement that
$\Fattr$ measures at the readout, roughly three quarters is
$v$-specific (i.e., genuinely about the readout direction) and roughly
one quarter is direction-agnostic late-stage residual-stream
consistency.

\subsection{Tuned lens results}
\begin{figure}[ht]
    \centering
    \includegraphics[width=0.8\linewidth]{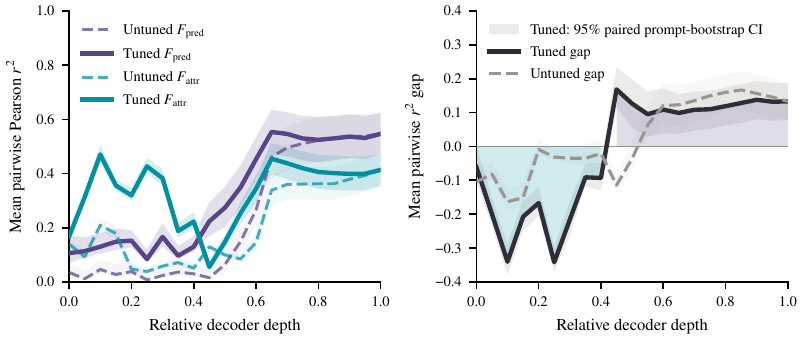}
    \caption{Tuned lens sensitivity analysis, compared to an untuned logit lens. }
    \label{fig:tuned_lens}
\end{figure}

As an exploratory sensitivity analysis, we replace the direct logit lens with identity-initialized affine tuned lenses trained on a disjoint BoolQ training subset by minimizing full-vocabulary $KL(p_\text{final} \| p_\text{lens})$, as in \citet{belrose2025elicitinglatentpredictionstransformers}. 
We plot the results of this experiment in \cref{fig:tuned_lens}:
Solid curves show the tuned lens and dashed curves the direct lens on the same 100 held-out prompts (569 user-message segments), with 1,000 prompt-cluster bootstrap resamples. The result is exploratory because it uses one seed, a sparse depth grid, and model-specific training budgets; however, it does not suggest that this design choice has significant impact on our results.

\subsection{Caveats and scope}
\label{sec:per_layer:caveats}

The per-layer extension inherits the assumptions of
\cref{sec:preliminaries,sec:method} but adds two of its own:

We use single-token
$\textsc{pos}$/$\textsc{neg}$ labels so the linearization $\ell^{(l)}
= \operatorname{RMSNorm}(h^{(l)}) \cdot v$ holds exactly modulo bf16 noise; under
the multi-token aggregation used for the main-paper $\Fpred$/$\Fattr$
the projection-based identity becomes a softmax-weighted approximation,
which changes both calibration and absolute scales.

Our regex sentence segmenter produces fewer segments per BoolQ
passage than the production sentence-tokenizer used in
\cref{sec:experiments}; the $\Fattr^{(L)}$ baseline we measure here
($\approx 0.40$) is consistent with the main-text Open$\to$Open
$\Fattr$ baseline within sampling noise, but absolute counts of
$(\textsc{prompt}, \textsc{segment})$ pairs are not directly
comparable. None of these caveats affects the qualitative shape of
the depth dissociation, which is robust under all controls in
\cref{fig:per_layer:bootstrap,fig:per_layer:random_dir}.

\FloatBarrier
\section{System-prompt perturbation as a secondary behavioral fidelity signal}
\label{sec:system-prompt-perturbation}
    The behavioral fidelity framework introduced in Section \ref{sec:method} measures attribution via input ablation: removing a segment of the prompt and observing the change in log-odds.
    We now consider an orthogonal causal intervention: mutate the system prompt, and ask whether models that respond similarly to context perturbation also exhibit greater surrogate fidelity.
    Where ablation measures sensitivity to the \emph{absence} of information, system-prompt perturbation measures sensitivity to its \emph{framing}.

    Concretely, we use GEPA \citep{agrawal2026gepareflectivepromptevolution}, a state-of-the-art prompt optimization algorithm, both as a method for generating a diverse set of system-prompt candidates and as a tool for per-model optimization.
    This enables us to assess two strict tests of context-level fidelity:
    (i) \emph{perturbation-response similarity}: do models whose predictions shift in correlated ways under the same system-prompt variation constitute better surrogates for one another?
    (ii) \emph{prompt transferability}: does a system prompt optimized with respect to donor model $M_D$ improve performance when applied to recipient model $M_R$, and does the degree of transfer serve as a directional surrogacy measure in contrast to the similarity metrics of Section \ref{sec:method}?

\paragraph{Experimental protocol.}

We evaluate the same eleven models, spanning four families (Qwen, Llama, GPT, Gemini), described in \cref{sec:experiments} on the ANLI R3 test split ($n=400$), chosen for its difficulty to ensure headroom for prompt-driven gains and reason labels required by GEPA's feedback mechanism. We construct three prompt conditions: (i)~\emph{candidates} -- ten system prompts selected for maximal tree-path diversity from a single GEPA run targeting GPT-4.1 (GPT-5.2 as reflector), applied uniformly to all models as shared context-level perturbations; (ii)~\emph{per-model best} -- the top-performing prompt from an independent GEPA run targeting each model individually; and (iii)~the \emph{baseline} system prompt used as the control. All GEPA runs use 20 optimization steps with minibatches of 20 from the ANLI R3 development split.

\subsection{Perturbation-response fidelity}
\label{sec:perturbation-response-fidelity}
We ask whether replacing the system prompt with a GEPA-generated alternative drives two models' predictions into alignment, and whether this alignment extends to \emph{how} those predictions shift.
For each prompt condition and model, we compute $\Fpred$ (Pearson $r^2$ of log-odds) and $\Fattr$ (Pearson $r^2$ of $\Delta\ell$ relative to the baseline).
Table~\ref{tab:prompt-perturbation-fidelity} reports the results.

\begin{table}[ht]
    \caption{
        \textbf{Cross-model fidelity under system-prompt perturbation on ANLI R3.}
        $n=400$ prompts, 11 models across 4 families.
        Each cell reports the mean pairwise Pearson $r^2$ over the relevant model-pair set.
        $\Fpred$ measures agreement on log-odds under a shared prompt condition; $\Fattr$ measures agreement on $\Delta\ell$ induced by replacing the baseline prompt.
        \emph{Top}: Prediction fidelity increases under optimized prompts (baseline $\bar{r}^2 = .385 \to$ candidates $\bar{r}^2 = .480$), indicating that structured system prompts drive cross-model output convergence.
        \emph{Bottom}: Attribution fidelity, defined by the shift in log-odds from perturbing the baseline system prompt, is uniformly low and does not improve with prompt specificity (candidates $\bar{r}^2 = .092 \to$ cross-model best $\bar{r}^2 = .044$).
    }
    \label{tab:prompt-perturbation-fidelity}
    \centering
    \small
    \begin{tabular}{ll ccccccc}
        \toprule
        & Condition
        & All
        & W.\,Qwen
        & W.\,Qwen\textsuperscript{-}
        & W.\,Llama
        & W.\,Llama\textsuperscript{-}
        & W.\,GPT
        & Cross \\
        \midrule
        $\Fpred$
        & Baseline                       & \textbf{.385} & .135 & .278 & .600 & .716 & .838 & .362 \\
        & Candidates                     & \textbf{.480} & .305 & .430 & .602 & .730 & .840 & .469 \\
        & Cross-model best               & \textbf{.478} & .297 & .541 & .610 & .816 & .839 & .467 \\
        & Per-model best\textsuperscript{\S} & \textbf{.479} & .303 & .596 & .620 & .808 & .863 & .468 \\
        \midrule
        $\Fattr$
        & Candidates                     & \textbf{.092} & .030 & .077 & .204 & .213 & .095 & .080 \\
        & Cross-model best               & \textbf{.044} & .023 & .062 & .113 & .163 & .000 & .038 \\
        & Per-model best\textsuperscript{\S} & \textbf{.069} & .058 & .058 & .089 & .035 & .033 & .071 \\
        \toprule
    \end{tabular}

    \vspace{4pt}
    {\footnotesize
    W.\,Qwen\textsuperscript{-} and W.\,Llama\textsuperscript{-} exclude Qwen-0.5B and Llama-8B respectively.

    \textsuperscript{\S}\,Non-uniform perturbation: each model evaluated under its own optimized prompt.
    }
\end{table}

The prediction-level finding is clear: replacing the baseline system prompt with a more structured alternative increases cross-model prediction agreement.
Overall agreement rises by about $25\%$ relative to baseline, driven mainly by Qwen models and cross-family pairs; Llama and GPT within-family agreement, which is already high in the baseline, does not increase by much.
The gain holds across all family groupings and is insensitive to prompt source; even per-model best prompts, where each model is under its own independently optimized prompt, act as a non-uniform perturbation, suggesting that prompt optimization drives models toward a shared output regime independent of perturbation direction.

\begin{figure}[t]
    \centering
    \includegraphics[width=0.6\linewidth]{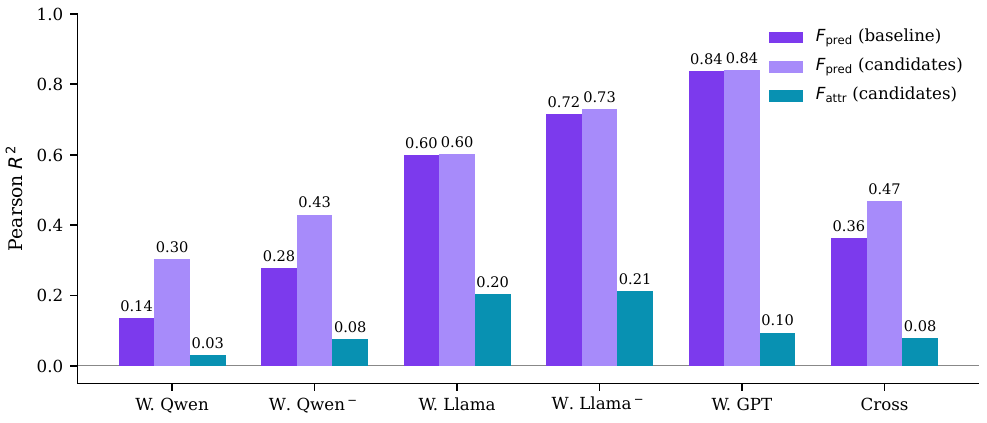}
    \caption{
        \textbf{Prediction and attribution fidelity under system-prompt perturbation on ANLI R3.}
        Pearson $r^2$ grouped by model-pair family.
        {\color[HTML]{7C3CE7} \textbf{Dark purple:}} $\Fpred$ under the baseline system prompt.
        {\color[HTML]{A78BF5} \textbf{Light purple:}} $\Fpred$ under diverse GEPA-generated candidates.
        {\color[HTML]{0791AF} \textbf{Cyan:}} $\Fattr$ under the same candidates, computed as the Pearson $r^2$ of $\Delta\ell$ induced by replacing the baseline prompt.
        Structured system prompts drive output convergence while attribution fidelity remains uniformly low, the output--behavior gap persists under prompt optimization.
    }
    \label{fig:prompt-perturbation-fidelity}
\end{figure}

Attribution fidelity tells a sharply different story.
Mean pairwise $\Fattr$ under diverse candidates is $r^2=.092$.
The like-for-like full-dialog entailment-minus-contradiction ablation diagnostic has median $r^2=.145$; the canonical all-margin ANLI-R3 result is centered RV =.291 and is not directly comparable. 
Under cross-model best prompts, attribution fidelity drops further to 
.044.
Despite near-perfect agreement on log-odds, prompts that shift one model's per item predictions do not reliably shift the other's, exemplifying the critical failure mode identified in Section~\ref{sec:attribution-fidelity}: high output fidelity masking divergent causal structure.

\subsection{Cross-model prompt transfer}
\label{sec:cross-model-prompt-transfer}
The bilateral fidelity metrics in Section~\ref{sec:perturbation-response-fidelity} treat prompt perturbation as a shared intervention.
 We now ask a directed question: can a prompt optimized for one model transfer its behavioral gains to another, and does the degree of transfer serve as a directional surrogacy measure that the symmetric metrics of Section \ref{sec:method} cannot provide?

\begin{figure}[!b]
    \centering
    \begin{minipage}{0.49\linewidth}
        \centering
        \includegraphics[width=\linewidth]{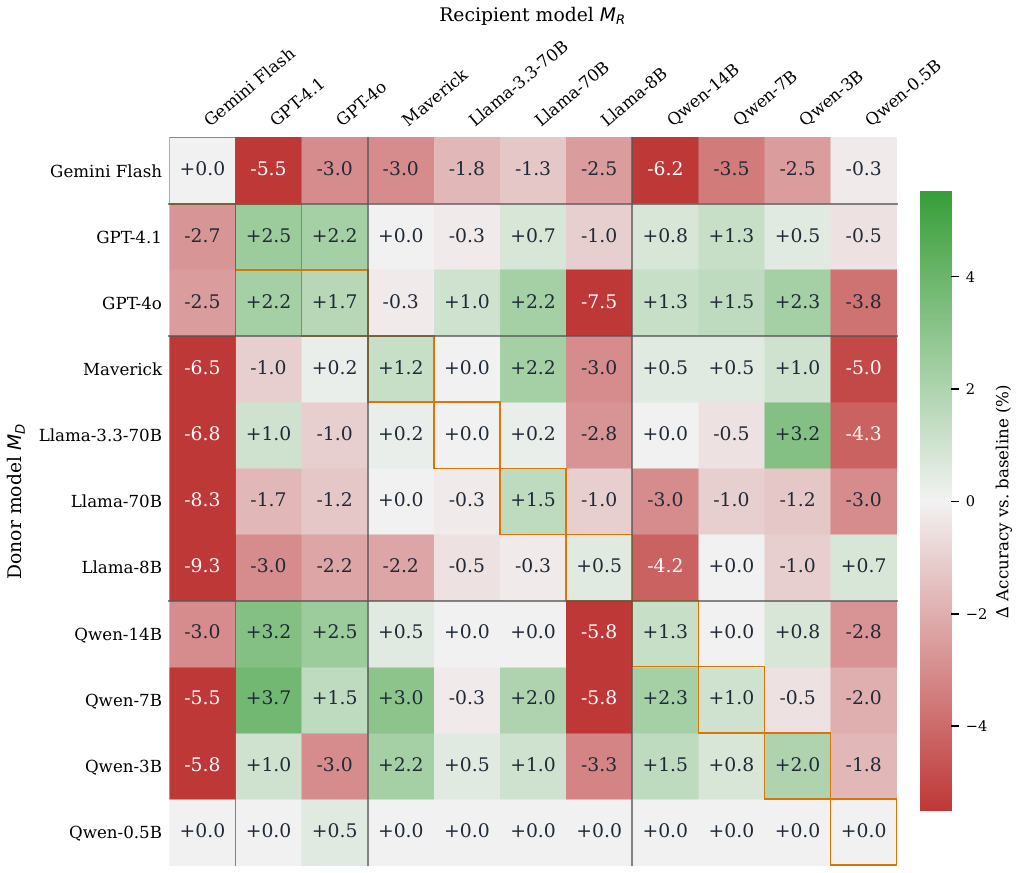}
    \end{minipage}\hfill
    \begin{minipage}{0.49\linewidth}
        \centering
        \includegraphics[width=\linewidth]{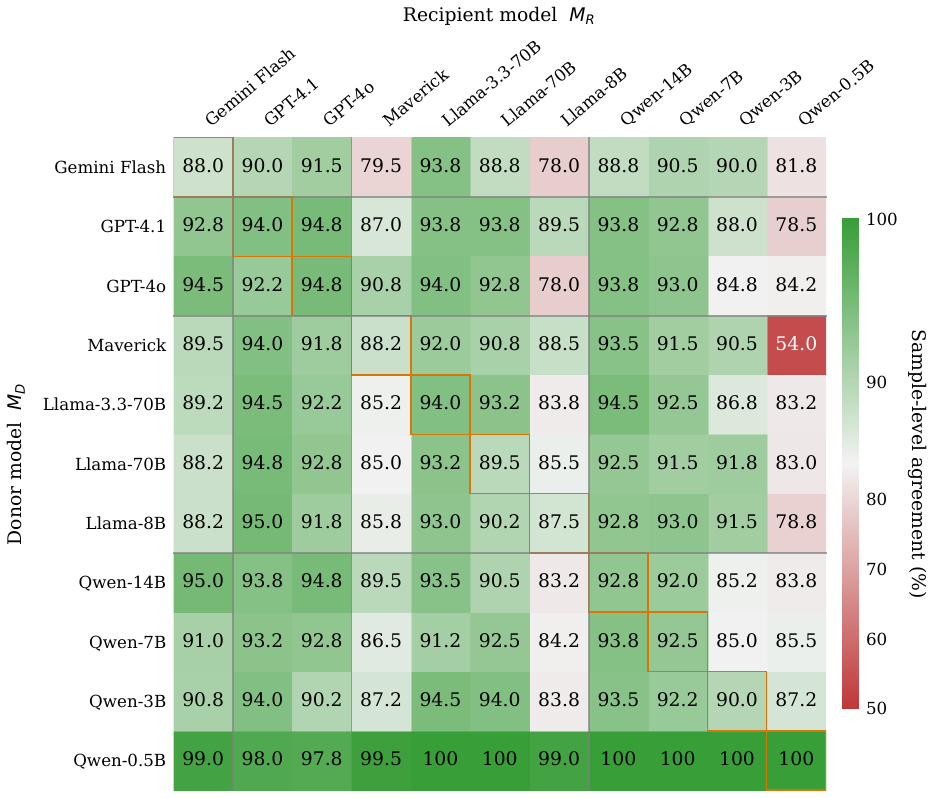}
    \end{minipage}
    \caption{
        \textbf{Cross-model system-prompt transfer on ANLI R3} ($n=400$).
        \emph{Left}: $\Delta$accuracy when recipient model $M_R$ is evaluated under a system prompt optimized for donor model $M_D$, relative to $M_R$'s baseline performance.
        Cross-model transfer is predominantly neutral or negative.
        \emph{Right}: sample-level prediction agreement between $M_R$ under $M_D$'s optimized prompt and $M_R$'s own baseline (off-diagonal mean 90.4\%). Optimized prompts largely preserve per-sample decisions despite shifting aggregate accuracy.
    }
    \label{fig:prompt-transfer}
\end{figure}

For each donor--recipient pair $(M_D, M_R)$, we apply the system prompt optimized for $M_D$ to $M_R$ and measure the accuracy change relative to $M_R$'s baseline.
Figure~\ref{fig:prompt-transfer} (left panel) reports the full $11 \times 11$ transfer matrix.
Cross-model transfer is predominantly neutral or negative (mean off-diagonal $\Delta$accuracy $= -1.4\%$): prompts that adapt to one model's failure patterns do not generally remedy another's.
Crucially, the matrix is asymmetric, transferring $M_D$'s prompt to $M_R$ yields a different outcome than the reverse. In line with the small accuracy deltas, sample-level agreement between $M_R$'s predictions under $M_D$'s optimized prompt and $M_R$'s own baseline is high (right panel; off-diagonal mean 90.4\%).
Optimized prompts largely preserve which samples each model answers correctly, perturbing a minority of per-item decisions.
The principal exception is Maverick $\to$ Qwen-0.5B (54.0\%), consistent with Qwen-0.5B's position as a general outlier (Section~\ref{sec:prediction-fidelity}).

\subsection{Voice transplant: prompt as directional bias vector}
\label{sec:voice-transplant}
$\Fpred$ and $\Fattr$ measure bilateral similarity: are two models alike?
They do not measure whether one model's behavior can be steered toward another's.
We introduce the \emph{voice-transplant correlation} to capture this asymmetric notion of surrogacy.
For each ordered pair $(M_D, M_R)$, we define \begin{equation} \mathrm{VT}(M_D \to M_R) = r\!\left(\ell_{M_D}^{\mathrm{baseline}} - \ell_{M_R}^{\mathrm{baseline}},\  \Delta\ell_{M_R}^{\mathrm{best}_{M_D}}\right) \end{equation} where $\ell_{M_D}^{\mathrm{baseline}} - \ell_{M_R}^{\mathrm{baseline}}$ is the natural difference in log-odds output between the donor and recipient under the baseline prompt, and $\Delta\ell_{M_R}^{\mathrm{best}_{M_D}}$ is the shift in the recipient's log-odds induced by applying the donor's optimized prompt.
Positive VT indicates the recipient shifts toward the donor; negative VT indicates the recipient shifts away.

\begin{figure}[t]
    \centering
    \includegraphics[width=0.65\linewidth]{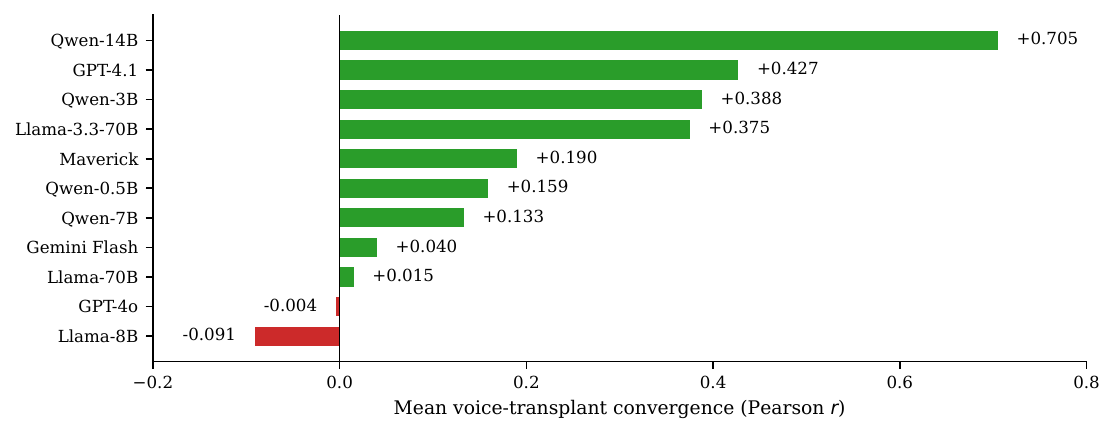}
    \caption{
       \textbf{Recipient susceptibility to voice transplant on ANLI R3.} For each ordered pair $(M_D, M_R)$, we define $\mathrm{VT}(M_D \to M_R) =$ Pearson $r$ on the gap in donor and recipient baseline log-odds $(\ell_{M_D} - \ell_{M_R})$ and the recipient's log-odds shift under the donor's optimized prompt. Bars show the mean VT across all non-self donors; positive values indicate the recipient shifts toward the donor, negative values indicate a shift away. Qwen-14B ($r = +0.71$) consistently converges toward its donors, while Llama-8B ($r = -0.09$) resists.
    }
    \label{fig:voice-transplant}
\end{figure}

Figure~\ref{fig:voice-transplant} reports mean VT received across all non-self donors, ranked by susceptibility.
Qwen-14B achieves the highest mean VT ($r = +0.71$): applying any other model's optimized prompt consistently shifts its log-odds toward the donor's baseline, indicating strong behavioral malleability.
Most models demonstrate some degree of positive convergence, GPT-4.1 ($r = +0.43$), Qwen-3B ($r = +0.39$), Llama-3.3-70B ($r = +0.38$), spanning model families and architectures, suggesting that prompt-based alignment is broadly viable.
At the other end, Llama-8B ($r = -0.09$) and GPT-4o ($r = -0.004$) show negligible or negative convergence: applying another model's optimized prompt does not reliably close the log-odds gap. 

\subsection{Summary}
\label{sec:summary}
System-prompt perturbation replicates the central finding under a complementary form of causal intervention.
Perturbing the instructional context drives cross-model output convergence while attribution fidelity remains uniformly low, independent of if the perturbation is a diverse set of candidates, or optimized for each model itself. The dissociation is starkest for GPT-4o and GPT-4.1, which achieve the highest prediction fidelity in our evaluation (0.840) yet near-zero attribution fidelity under every prompt condition tested.
Cross-model prompt transfer shows limited success in our evaluation, though the degree and direction of transfer vary across pairs, indicating that prompt-level surrogacy is inherently directional.
Finally, we propose voice-transplant (VT) susceptibility as a lightweight, API-accessible measure of a model's impressionability, its capacity to be behaviorally steered toward a target. This provides MI practitioners a preliminary diagnostic for identifying which surrogates are amenable to prompt-based alignment before committing to mechanistic analysis.

\FloatBarrier
\section{Connection to representational similarity}
\label{sec:cka}

  The progression from binary log-odds to multi-class pairwise log-odds suggests a deeper perspective: our fidelity metrics are \textit{task-relevant projections} of a full representational similarity measure, while accessible through log-probabilities alone.

  \paragraph{From scalars to vectors to representations.}
  In the binary case, prediction fidelity compares the scalar log-odds $\ell(x) = z \cdot v$ across prompts, where $v = u_+ - u_-$ is the log-odds direction. This is a projection of the
  post-normalization residual stream $z \in \mathbb{R}^d$ onto a single direction in representation space. In the multi-class case \cref{sec:log_odds_details}, the pairwise log-odds vector $\boldsymbol{\ell}(x) \in
  \mathbb{R}^{\binom{K}{2}}$ corresponds to projecting $z$ onto $\binom{K}{2}$ directions simultaneously:
  \begin{equation}
      \ell_{ab}(x) = z \cdot (u_a - u_b), \qquad (a, b) \in \mathcal{C},
      \label{eq:projection}
  \end{equation}
  where $u_a, u_b$ are columns of the unembedding matrix $W_U$. Stacking these projections, the full pairwise log-odds vector is $\boldsymbol{\ell}(x) = Pz$, where $P \in \mathbb{R}^{\binom{K}{2} \times d}$.

  The same applies to attributions: the vector-valued ablation response $\boldsymbol{\Delta\ell}_i(x) = P\Delta z_i$ (\cref{sec:log_odds_details}) is the projection of the representation perturbation $\Delta
  z_i = z - z'_i$ onto the same task-relevant subspace.

  At the limit, one could skip the projection entirely and compare the full residual-stream matrices $\mathbf{Z}_M \in \mathbb{R}^{n \times d}$ directly. This requires white-box access to model internals, but
  yields a complete picture of representational similarity.

  \paragraph{Centered Kernel Alignment.}
  The natural measure for comparing full representation matrices is \textit{Centered Kernel Alignment} (CKA)~\citep{kornblith2019similarity}, which has become the standard for comparing neural network
  representations across architectures. For centered data matrices $\mathbf{X}, \mathbf{Y} \in \mathbb{R}^{n \times d}$, linear CKA is defined as:
  \begin{equation}
      \text{CKA}(\mathbf{X}, \mathbf{Y}) = \frac{\operatorname{tr}\!\left(\mathbf{X}\mathbf{X}^\top \mathbf{Y}\mathbf{Y}^\top\right)}{\sqrt{\operatorname{tr}\!\left((\mathbf{X}\mathbf{X}^\top)^2\right) \cdot
  \operatorname{tr}\!\left((\mathbf{Y}\mathbf{Y}^\top)^2\right)}}.
      \label{eq:cka}
  \end{equation}
  This is the cosine similarity between the two models' Gram matrices $\mathbf{X}\mathbf{X}^\top$ and $\mathbf{Y}\mathbf{Y}^\top$ in the Hilbert--Schmidt inner product space. Each Gram matrix entry
  $(\mathbf{X}\mathbf{X}^\top)_{ij} = x_i \cdot x_j$ captures the dot-product similarity between two prompts' representations within a model; CKA asks whether these pairwise similarity structures agree across
  models.

\begin{figure}[ht]
    \centering
    \begin{minipage}{0.3\linewidth}
        \centering
        \includegraphics[width=\linewidth]{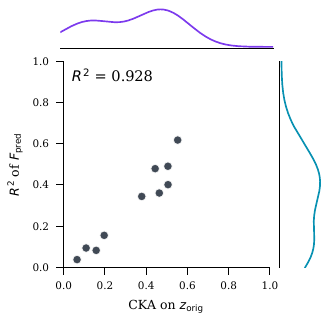}
    \end{minipage}\hspace{1em}
    \begin{minipage}{0.3\linewidth}
        \centering
        \includegraphics[width=\linewidth]{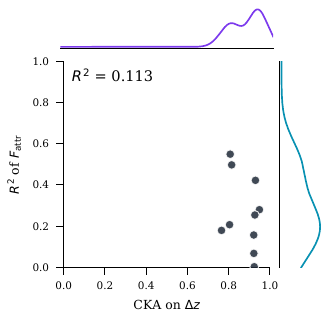}
    \end{minipage}
    \caption{
        \textbf{Correlation between representation CKA and $r^2$ Fidelity on BoolQ.} The prediction fidelity and CKA are perfectly aligned, but surprisingly the attribution fidelity says the opposite.
    }
    \label{fig:cka_vs_fidelity}
\end{figure}

  Linear CKA is mathematically equivalent to the RV coefficient. When applied to scalar data ($d = 1$), it reduces to the squared Pearson correlation $r^2$ --- which is
  the binary prediction fidelity metric $F_\text{pred}$ of \cref{eq:f_pred}. The RV coefficient (\cref{eq:rv}) applied to multi-class pairwise log-odds is therefore CKA on the projected representations
  $P\mathbf{Z}$. Our fidelity framework forms a unified hierarchy:
  \begin{equation}
      \underbrace{r^2(\ell_{M_S}, \ell_{M_T})}_{\text{binary } F\text{pred}}
      \subset
      \underbrace{\text{RV}(\boldsymbol{L}_{M_S}, \boldsymbol{L}_{M_T})}_{\text{multi-class } F\text{pred}}
      \subset
      \underbrace{\text{CKA}(\mathbf{Z}_{M_S}, \mathbf{Z}_{M_T})}_{\text{representational similarity}}
      \label{eq:hierarchy}
  \end{equation}
  Each level projects the residual stream onto progressively fewer task-relevant directions: $d$ (full representation), $\binom{K}{2}$ (pairwise log-odds), or $1$ (binary log-odds).

  \paragraph{Task-relevant projections vs.\ random projections.}
  Our projections are not arbitrary --- they use the unembedding directions $(u_a - u_b)$ that the model was \textit{trained} to separate. The training objective (next-token prediction) optimizes the last-layer representation $z$ so that its projection onto unembedding vectors produces the correct probability distribution. These directions therefore capture maximal task-relevant variance in $z$ by construction.

  By contrast, the Johnson--Lindenstrauss lemma~\citep{Johnson1984ExtensionsOL} guarantees that random projections from $\mathbb{R}^d$ to $\mathbb{R}^k$ preserve pairwise dot products up to $(1 \pm \varepsilon)$
  distortion when $k \geq O(\log n / \varepsilon^2)$. For $n = 5{,}000$ prompts and $\varepsilon = 0.1$, this requires $k \approx 850$ random dimensions to approximate the full Gram matrix. Our task-relevant
  projections use only $k = \binom{K}{2}$ directions (6 for a 4-class task) yet capture the subspace that determines the classification output, i.e., the operationally relevant subspace for surrogate fidelity.

  This perspective also clarifies the practical value of our fidelity framework: it provides a \textit{window into representational similarity through black-box API access}. Full CKA requires white-box
  access to hidden states; our projected variant requires only log-probabilities for the related tokens.

\begin{figure}[ht]
    \centering
    \includegraphics[width=0.9\linewidth]{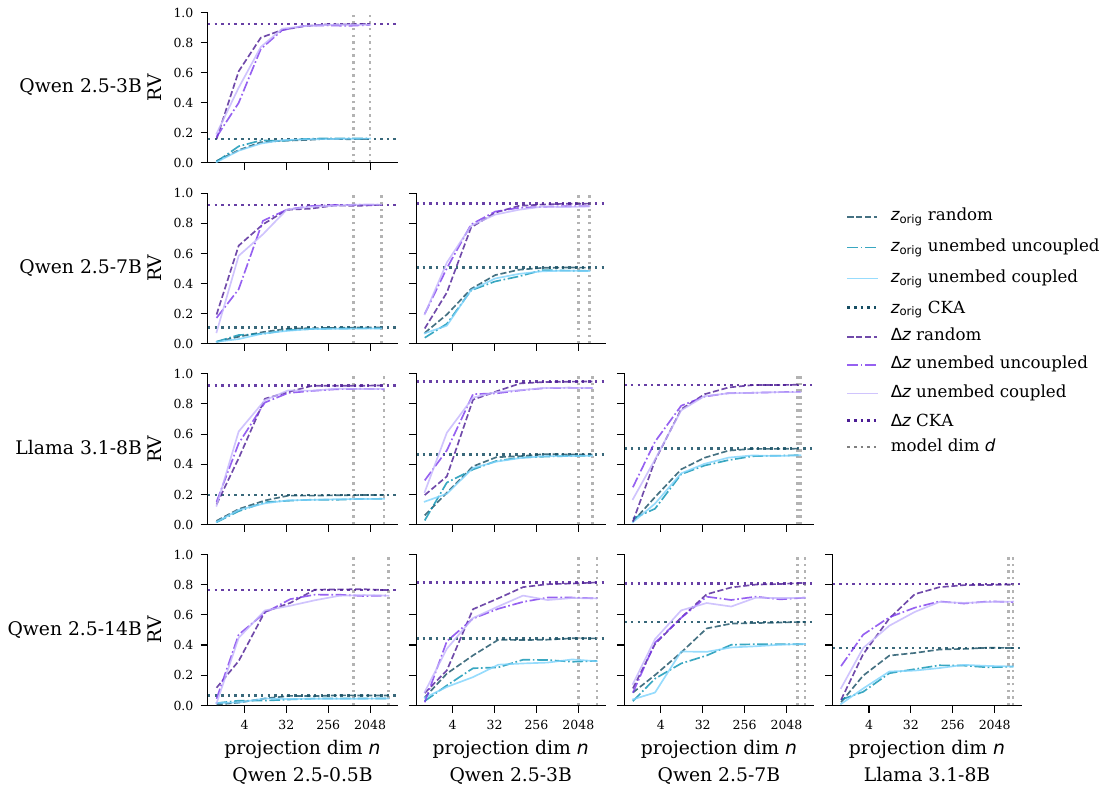}
    \caption{
       \textbf{RV convergence to CKA.} X-axis: projection dimension $n$ (log scale, 1 to $d$). Y-axis: RV on projected data. Three curves (random Gaussian, unembeddings coupled, unembeddings uncoupled) for prediction and attribution respectively converging to horizontal dashed line (representation CKA).
    }
    \label{fig:rv_convergence}
\end{figure}

  \paragraph{Projected vs.\ full representational CKA.}
  To validate that our log-odds fidelity metrics are faithful proxies for representational similarity, we compute both full CKA on the residual stream $z$ and $r^2$ on binary log-odds on BoolQ for all pairs of open-source models (Qwen2.5-{0.5B, 3B, 7B, 14B}-Instruct and Llama-3.1-8B, yielding 10 model pairs). This is done for both prediction and attribution fidelity. \cref{fig:cka_vs_fidelity} shows the correlation between the two measures across model pairs.

  \paragraph{Convergence of random projections.}
  To quantify the efficiency of task-relevant projections, we compare three projection strategies as a function of projection dimension $n$ from $n = 1$ over $n = d$ (the model's latent dimension):
  \begin{enumerate}[nosep]
      \item \textbf{Random Gaussian:} project $z$ onto $n$ i.i.d.\ random unit vectors.
      \item \textbf{Unembeddings coupled:} project $z$ onto $n$ randomly sampled tokens unembeddings (same token for $M_S$ and $M_T$).
      \item \textbf{Unembeddings uncoupled:} project $z$ onto $n$ randomly sampled tokens unembeddings (independent tokens for $M_S$ and $M_T$).
  \end{enumerate}
  For each $n$ and each strategy, we compute RV on the $n$-dimensional projected data and track convergence toward full CKA. \cref{fig:rv_convergence} shows convergence curves of different model pairs. From the plots, we can see coupled unembeddings usually converge a little faster, but to a lower plateau, while only random projection can reach CKA-like levels. This aligns with our expectation that the unembeddings capture a much more relevant subspace, which is effective but limited in covering the whole representation space. 

  Comparing with the $r^2$ fidelity values shown in \cref{fig:dashboard}, we can see the heatmap's numbers are much higher than the corresponding model pairs' curve at the same dimension $d = 1$. This shows our task-relevant log-odds is an effective approach to estimate models'
  representational agreement.

\FloatBarrier
\section{Robustness of the fidelity estimates}
\label{app:robustness}

This appendix examines how the fidelity estimates of \cref{sec:results} depend on the choices behind them: the statistic, the label readout, the treatment of missing values, run-to-run noise, segment length, and which prompts, segments and pairs are pooled. 
Table~\ref{tab:rob-grid} collects the segment-level metrics under each measurement variant, and the sections below discuss its columns in turn. 
Unless stated, values are medians of Pearson $r^2$ over model pairs for BoolQ sentence-level ablations.

\begin{table}[ht]
\caption{BoolQ sentence-level metrics under each measurement variant (medians
over pairs). The first block compares the same signal in two models, over all
55 pairs for $F_{\mathrm{attr}}$ and over the ten pairs of open models
otherwise. The second block uses an open model's signal to predict another
model's attribution (Open$\to$All); rows marked $|\mathrm{attr}|$ predict
unsigned attribution. \emph{Prompt-centered} subtracts each prompt's mean from
both signals. \emph{Per-prompt} $\rho$ is the mean of Spearman correlations
computed within prompts. \emph{Length-controlled} residualizes the open
model's signal on the log word count of the segment. Columns of $\rho$ and
$r^2$ are on different scales. A dash marks a variant that is undefined or was
not computed.}
\label{tab:rob-grid}
\centering
\small
\resizebox{\linewidth}{!}{\begin{tabular}{lccccccc}
\toprule
 & \multicolumn{5}{c}{All segments} & \multicolumn{2}{c}{Passage sentences} \\
\cmidrule(lr){2-6}\cmidrule(lr){7-8}
 & & Prompt- & & Per-prompt & Length- & & Length- \\
Metric & $r^2$ & centered $r^2$ & $\rho$ & $\rho$ & controlled $r^2$ & $r^2$ & controlled $r^2$ \\
\midrule
\multicolumn{8}{l}{\emph{Same signal in two models}} \\
$F_{\mathrm{attr}}$ & .451 & .473 & .392 & .406 & -- & .404 & -- \\
$F_{\mathrm{mag}}$ & .843 & .850 & .890 & .858 & .845 & .599 & -- \\
$F_{\mathrm{align}}$ & .201 & .207 & .355 & .383 & -- & .137 & -- \\
$F_{\mathrm{attn}}^{\mathrm{mean}}$ & .904 & .895 & .951 & .915 & .904 & -- & -- \\
$F_{\mathrm{attn}}^{\mathrm{max}}$ & .859 & .851 & .919 & .885 & .865 & -- & -- \\
$F_{\mathrm{attn}}^{\mathrm{rollout}}$ & .861 & .880 & .954 & .950 & .882 & -- & -- \\
\midrule
\multicolumn{8}{l}{\emph{An open model's signal against another model's attribution}} \\
$F_{\mathrm{mag}\to|\mathrm{attr}|}$ & .108 & .092 & .489 & .385 & .109 & .425 & .398 \\
$F_{\mathrm{align}\to\mathrm{attr}}$ & .257 & .259 & .359 & .381 & -- & .183 & -- \\
$F_{\mathrm{attn}\to\mathrm{attr}}^{\mathrm{mean}}$ & .002 & .004 & $-$.015 & $-$.039 & .002 & .029 & .023 \\
$F_{\mathrm{attn}\to\mathrm{attr}}^{\mathrm{max}}$ & .002 & .003 & $-$.021 & $-$.053 & .002 & .028 & .023 \\
$F_{\mathrm{attn}\to\mathrm{attr}}^{\mathrm{rollout}}$ & .003 & .005 & $-$.044 & $-$.023 & .002 & .003 & .001 \\
$F_{\mathrm{attn}\to|\mathrm{attr}|}^{\mathrm{mean}}$ & .199 & -- & -- & -- & .199 & .136 & .114 \\
$F_{\mathrm{attn}\to|\mathrm{attr}|}^{\mathrm{max}}$ & .118 & -- & -- & -- & .119 & .121 & .090 \\
$F_{\mathrm{attn}\to|\mathrm{attr}|}^{\mathrm{rollout}}$ & .040 & -- & -- & -- & .046 & .022 & .010 \\
\bottomrule
\end{tabular}
}
\end{table}

\subsection{Metric calibration}
\label{app:rob-calibration}

\begin{figure}[ht]
\centering
\includegraphics[width=0.85\linewidth]{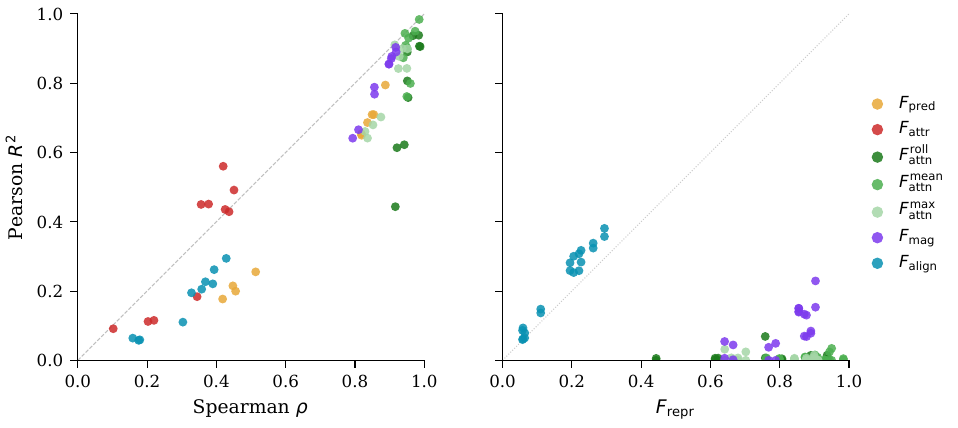}
\caption{\textbf{Left:} a scatterplot of Spearman $\rho$ against Pearson $r^2$.
Note that only $F_{\mathrm{attr}}$ has $r^2 > \rho$. \textbf{Right:} a
scatterplot of $F_{\mathrm{mech}}$ versus $F_{\mathrm{cross}}$ for different
mechanistic signals. Note that $F_{\mathrm{pred}}$ and
$F_{\mathrm{attr}}$ are excluded here, and that only $F_{\mathrm{align}}$ has
better cross- than self-prediction.}
\label{fig:metric-calibration}
\end{figure}

In Figure~\ref{fig:metric-calibration}, we compare Spearman $\rho$ and $r^2$,
as well as $F_{\mathrm{mech}}$ and $F_{\mathrm{cross}}$ scores, in open models.
Each point represents one pair of distinct open models.

\paragraph{Pearson and Spearman, pooled and within prompts.}
Unlike $F_{\mathrm{pred}}$, where $r^2$ closely tracks Spearman $\rho$, $r^2$
values for $F_{\mathrm{attr}}$ are slightly higher than Spearman $\rho$
(Figure~\ref{fig:metric-calibration}, left), as they are significantly more
stable in the presence of near-zero attributions.
The repeat runs of Section~\ref{app:rob-noise} support this reading: two runs
of GPT-4o agree with $r = .953$ but $\rho = .869$, because the ordering among
many near-zero attributions is not reproducible.

The first four columns of Table~\ref{tab:rob-grid} extend the comparison to all
pairs and add two within-prompt statistics. Centering within prompts moves no
$r^2$ median by more than .05, so pooled $r^2$ is not driven by differences in
scale between prompts. Rank correlation gives the same picture with one
exception: for perturbation magnitude against unsigned attribution, pooled
$\rho$ (.489) is far higher than pooled $r^2$ (.108) suggests. Section~\ref{app:rob-passage} traces the low
$r^2$ to system-prompt segments.

\paragraph{Self- and cross-prediction.}
The right panel of Figure~\ref{fig:metric-calibration} shows that agreement
between two models on a mechanistic signal ($F_{\mathrm{mech}}$) does
not imply that one model's signal predicts the other's attribution
($F_{\mathrm{cross}}$). For perturbation magnitude, the low cross-prediction is
specific to system-prompt segments (Section~\ref{app:rob-passage}).

\begin{figure}[ht]
\centering
\includegraphics[width=0.75\linewidth]{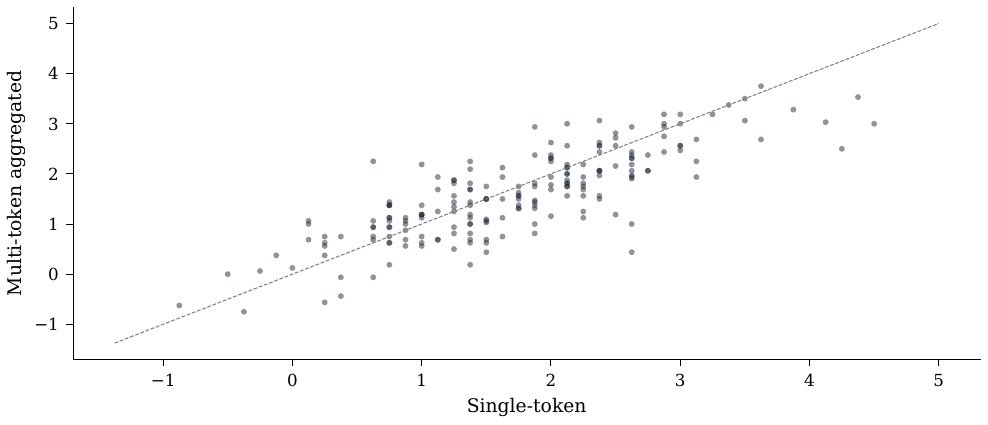}
\caption{The correlation between the log-sum-exp of a set of tokens and a
single-token variant is high, likely due to tight clustering between
semantically similar tokens in unembedding space. Across the open models used
in this paper, the average unembedding vector cosine similarity is 0.648 for
true tokens, 0.647 for false tokens, and 0.515 between true and false tokens.
The high cosine similarity between true and false tokens implies that the
true-vs-false direction captured in the log-odds is only a small fraction of
what is represented in the LLM's last layer.}
\label{fig:logsumexp}
\end{figure}

\paragraph{Label variants.}
Figure~\ref{fig:logsumexp} shows that the log-sum-exp over variants of the
``True'' token in BoolQ is highly correlated with the individual ``true'' vs
``false'' direction. In practice, only a single difference of unembeddings is
linear; the log-sum-exp breaks this nonlinearity. Because these quantities are
so similar to one another, we do not expect significant issues from this
approximation.

\subsection{Missing values}
\label{app:rob-missing}

Hosted models return the top $K = 19$ log-probabilities of the first generated
token, and a response can omit every variant of a label. We handle this in two
ways. For the binary benchmarks (BoolQ, WinoGrande) we use pairwise complete
cases: a prompt or segment enters a pair's statistic only when the log-odds are
defined for both models, before and after the deletion. For the multi-class
benchmarks (ANLI, RACE) we impute a floor (\cref{eq:imputed_floor}). 
How often a hosted model omits a label score depends on the benchmark:
0.2--9.4\% of label scores on BoolQ, 9.1--24.0\% on WinoGrande, 0.3--32.9\% on
RACE and 13.5--51.0\% on ANLI, across the six hosted models. Open models omit
none.

\paragraph{Binary benchmarks: complete cases.}
As a sensitivity analysis we recomputed BoolQ fidelity with a floor in place
of complete cases, replacing each undefined value with a value just beyond
that model's most extreme finite one. By pair set, the $F_{\mathrm{pred}}$
median moves by at most .015. $F_{\mathrm{attr}}$ for hosted pairs is not
robust to the choice: the floor lowers the open--hosted median from .404 to
.165 and the hosted--hosted median from .521 to .133. This is an artifact of
the imputation. A label drops out of the top $K$ when the model is nearly
certain of the other one, and the floor assigns every such segment the model's
most extreme attribution regardless of the true effect. For GPT-4.1, 15\% of
segments are imputed, at $\pm 59$ nats against a 99th percentile of 42 nats
among observed $|\Delta\ell|$, and they account for 81\% of the variance of the
imputed signal (GPT-4o: 8.5\% of segments, 70\% of the variance). Open--open
pairs have no missing labels and give .399 under either policy, close to the
open--hosted complete-case value. The cost of complete cases is coverage: the
median share of segments that are jointly observed is .95 for open--hosted
pairs and .75 for pairs involving GPT-4.1, and the excluded segments are those
on which a hosted model is most confident.

\begin{figure}[ht]
\centering
\includegraphics[width=0.85\linewidth]{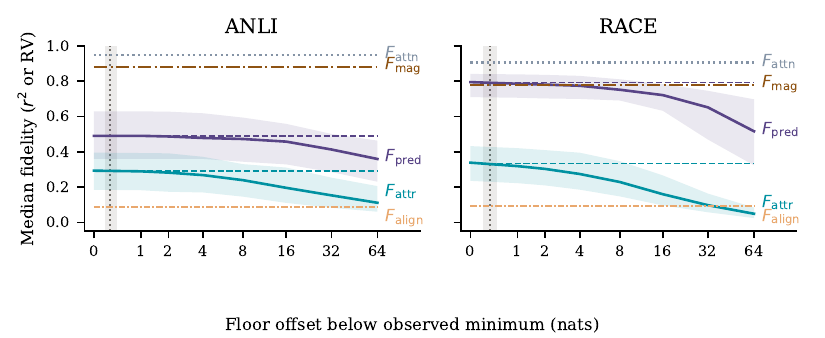}
\caption{Sensitivity to choice of log-probs floor for ANLI and RACE benchmarks.
We see that, while the absolute magnitude of $F_{\mathrm{pred}}$ and
$F_{\mathrm{attr}}$ is sensitive to this metric, the overall signal fidelity
responds identically across all signals. We include mechanistic
signals as horizontal lines, and mark the actual imputed log-probs value we
used in this paper as a vertical dotted line.}
\label{fig:floor-sensitivity}
\end{figure}

\paragraph{Multi-class benchmarks: floor sensitivity.}
In multiclass settings, it may be the case that minority classes are not in the
top $k$ tokens, and therefore have no measured log-probability available. In
our paper, we impute these values by setting them slightly lower than the
minimum observed log-probabilities, but it is reasonable to wonder what the
effect of this is on reported fidelity values. We sweep the imputation value as
a threshold in Figure~\ref{fig:floor-sensitivity} and find that, while the
absolute $F_{\mathrm{attr}}$ and $F_{\mathrm{pred}}$ values are somewhat
sensitive to these values, the overall ordering and dynamics are not.
The floor matters more on these benchmarks than on BoolQ because more is
imputed. Absolute multi-class fidelity values for pairs involving hosted models
should be read with this in mind.

\subsection{Noise floor and ceiling}
\label{app:rob-noise}

Attribution fidelity is bounded on both sides. From below, two models can
agree without either responding to what a segment says. From above, a hosted
model does not fully agree with its own repeat run. We estimate both bounds.

\paragraph{Floor.}
We shuffle one model's attributions among segments that share a content-free
property and recompute $r^2$, with 100 shuffles per pair
(Table~\ref{tab:rob-floor}). An unstructured shuffle leaves nothing. So does a
shuffle among segments with the same role (system, passage or question) and
position: role alone says nothing about a signed attribution, because the
direction of a segment's effect depends on the answer. Matching also on the
answer each model gave leaves .175 over all segments, and up to
.266 for a single pair. This is the agreement available from a
segment's role and the two models' answers, without reading the segment, and
it is about a third of the observed .451. It comes from the question segment,
whose deletion moves both models away from the answer they gave: with question
segments excluded the same shuffle leaves .017. On passage sentences it
also leaves .017, and at most .033 for any pair, against an observed
.404. In every pair and under every shuffle, the observed value exceeds all
100 shuffled values.

\begin{table}[ht]
\caption{Floor on BoolQ sentence-level $F_{\mathrm{attr}}$. One model's
attributions are shuffled among segments of the stated kind and $r^2$ is
recomputed; entries are the median and maximum over the 55 pairs of the mean
over 100 shuffles. The first row gives the observed values.}
\label{tab:rob-floor}
\centering
\small
\begin{tabular}{lcccc}
\toprule
 & \multicolumn{2}{c}{All segments} & \multicolumn{2}{c}{Passage sentences} \\
\cmidrule(lr){2-3}\cmidrule(lr){4-5}
Attributions shuffled among segments of & Median & Max & Median & Max \\
\midrule
Observed (no shuffle) & .451 & .826 & .404 & .853 \\
\midrule
Any segment & .000 & .000 & .000 & .000 \\
Same prompt & .025 & .050 & .040 & .096 \\
Same role and position & .000 & .004 & .000 & .004 \\
\quad and same gold answer & .086 & .133 & .010 & .016 \\
\quad and same answer from each model & .175 & .266 & .017 & .033 \\
\bottomrule
\end{tabular}

\end{table}

The same shuffles applied to the mechanistic signals of the open
models show where segment role matters most. For perturbation magnitude, which
is unsigned, matching on role and position alone leaves .681 of the
observed .843 over all segments, and .046 of the observed .599 on
passage sentences (Section~\ref{app:rob-passage}).

\paragraph{Ceiling.}
We re-ran every model five times on the full BoolQ sentence-level configuration
(3,270 prompts, 27,516 segments) with the published protocol, and correlated
each signal with itself across runs (Table~\ref{tab:rob-noise}). Open models
reproduce the published results exactly. Among hosted models, noise is
concentrated in the segment attributions of the two GPT models.

\begin{table}[ht]
\caption{Run-to-run repeatability on BoolQ sentence ablations. Columns 2--5 are
means over the ten pairs of repeat runs; the last column compares each repeat
with the published run, collected about two months earlier, and so also
reflects changes in the served model.}
\label{tab:rob-noise}
\centering
\small
\begin{tabular}{lccccc}
\toprule
 & $F_{\mathrm{pred}}$ & \multicolumn{3}{c}{$F_{\mathrm{attr}}$, run against run} & $F_{\mathrm{attr}}$ \\
\cmidrule(lr){3-5}
Model & $r$ & $r$ & $\rho$ & Min.\ coverage & $r$, run vs.\ published \\
\midrule
Open models (all five) & 1.000 & 1.000 & 1.000 & 1.00 & 1.000 \\
Llama-3.1-70B & 1.000 & 0.999 & 0.992 & 0.95 & 0.999 \\
Llama-3.3-70B & 1.000 & 1.000 & 0.993 & 0.95 & 0.999 \\
Llama-4 Maverick & 0.999 & 0.996 & 0.983 & 0.97 & 0.996 \\
GPT-4o & 0.992 & 0.953 & 0.869 & 0.91 & 0.891 \\
GPT-4.1 & 0.996 & 0.958 & 0.858 & 0.73 & 0.935 \\
Gemini 2.5 Flash Lite & 1.000 & 1.000 & 1.000 & 1.00 & 0.925 \\
\bottomrule
\end{tabular}

\end{table}

Write a model's measured attribution as a reproducible signal plus independent
noise, and let $\rho_M$ be the correlation between two runs of model $M$. Then
the squared correlation between any noise-free predictor and one run of $M$ is
at most $\rho_M$, and for two noisy models
\begin{equation}
  F_{\mathrm{attr}}(M_S, M_T) \;\le\; \rho_{M_S}\,\rho_{M_T}.
  \label{eq:noise-ceiling}
\end{equation}
The lowest ceiling among our 55 pairs is $.953 \times .958 = .913$ (GPT-4o with
GPT-4.1); for an open surrogate and GPT-4o it is .953. The median
$F_{\mathrm{attr}}$ of .451 becomes .455 after dividing each pair by its
ceiling. Run-to-run noise therefore explains almost none of the gap between
attribution fidelity and one.

\paragraph{Stability of the cross-model estimates.}
Recomputing every BoolQ $F_{\mathrm{pred}}$ and $F_{\mathrm{attr}}$ point
estimate on each repeat gives a standard deviation of at most .007 across
repeats. Against the published run the estimates shift more, because hosted
endpoints changed between the two collection dates: Gemini returned identical
outputs in all five repeats, yet these correlate with the published run at only
$r = .925$. Pairs involving Gemini have $F_{\mathrm{attr}}$ up to .058 higher
on the repeats. The all-pairs median is .451 on the published run and .456 on
the repeats.

\subsection{Segment length}
\label{app:rob-length}

Longer segments receive more attention and move the residual stream further,
so length could inflate agreement on mechanistic signals. We
residualize each open model's $\|\Delta z\|$ and attention scores on the log
word count of the segment (Table~\ref{tab:rob-grid}, length-controlled
columns). Over all segments, length explains at most .11 of the variance of any
signal, and residualizing moves no median by more than .021. Length matters
more within passages, where it explains .05--.30 of the variance and the
control lowers the cross-level medians by up to .031; where
Sections~\ref{app:rob-passage} and~\ref{app:rob-unsigned} quote passage-only
values they give the controlled value alongside.

\subsection{Labels and correctness}
\label{app:rob-labels}

$F_{\mathrm{pred}}$ pools prompts whose gold answers differ, so part of it
reflects agreement on the answer and not on confidence. Attributions are
differences of log-odds, in which a label main effect cancels; what remains of
the label is its interaction with segment role (Section~\ref{app:rob-noise}).
The two metrics are therefore not on a common scale. Table~\ref{tab:rob-label}
recomputes both within one gold label, among prompts both models answer
correctly, and within one label among those.

\begin{table}[ht]
\caption{BoolQ sentence-level fidelity on subsets of prompts (median over
pairs). ``Both correct'' requires $\mathrm{sign}(\ell)$ to match the gold label
for both models. The last column counts pairs, out of 55, in which
$F_{\mathrm{pred}}$ exceeds $F_{\mathrm{attr}}$.}
\label{tab:rob-label}
\centering
\small
\begin{tabular}{lcccccc}
\toprule
 & & \multicolumn{2}{c}{Open$\to$Closed} & \multicolumn{2}{c}{All pairs} & Pairs with \\
\cmidrule(lr){3-4}\cmidrule(lr){5-6}
Prompts & Median $n$ & $F_{\mathrm{pred}}$ & $F_{\mathrm{attr}}$ & $F_{\mathrm{pred}}$ & $F_{\mathrm{attr}}$ & $F_{\mathrm{pred}} > F_{\mathrm{attr}}$ \\
\midrule
All prompts & 3{,}267 & .669 & .404 & .718 & .451 & 55/55 \\
\midrule
Gold True & 2{,}033 & .429 & .388 & .498 & .428 & 52/55 \\
Gold False & 1{,}236 & .333 & .320 & .407 & .429 & 19/55 \\
\midrule
Both correct & 2{,}356 & .881 & .531 & .883 & .586 & 55/55 \\
Both correct, gold True & 1{,}486 & .188 & .533 & .214 & .559 & 3/55 \\
Both correct, gold False & 1{,}016 & .150 & .489 & .197 & .508 & 2/55 \\
\bottomrule
\end{tabular}

\end{table}

Within one label, $F_{\mathrm{pred}}$ falls from .718 to .498 and .407 while
$F_{\mathrm{attr}}$ is unchanged. Restricting to prompts both models answer
correctly raises both in pooled form, because the subset is two well-separated
clusters. Within one label among those prompts, $F_{\mathrm{pred}}$ is about .2
and $F_{\mathrm{attr}}$ is .51--.56; $F_{\mathrm{pred}}$ exceeds
$F_{\mathrm{attr}}$ in 2 or 3 of 55 pairs. Pooled prediction fidelity thus
mostly measures agreement on the answer. It remains useful for choosing among
surrogates: across the 55 pairs, pooled $F_{\mathrm{pred}}$ and
$F_{\mathrm{attr}}$ have a rank correlation of .94.

\subsection{Passage-only ablations}
\label{app:rob-passage}

Deleting the question or a system-prompt segment is a different kind of
intervention from deleting a passage sentence: it removes the task or the
query, not a piece of evidence. The last two columns of
Table~\ref{tab:rob-grid} repeat the comparisons on passage sentences alone.

Attribution fidelity changes little, from .451 to .404 over all pairs and from
.404 to .365 for open-to-closed pairs, but what it measures changes. The floor
of Section~\ref{app:rob-noise} is .175 over all segments and .017 on
passage sentences, so the passage-only value is the cleaner measure of
agreement about content. One cross-level result changes substantially. On
passage sentences, perturbation magnitude explains .425 of the size of another
model's attribution (.398 after length control), against .108 over all
segments. The all-segment value is depressed by system-prompt segments, whose
deletion moves the residual stream a great deal and the answer very little.
Alignment predicts signed attribution somewhat less well on passage sentences
(.183 against .257). Attention remains uninformative about signed attribution.

\begin{figure}[ht]
\centering
\includegraphics[width=0.42\linewidth]{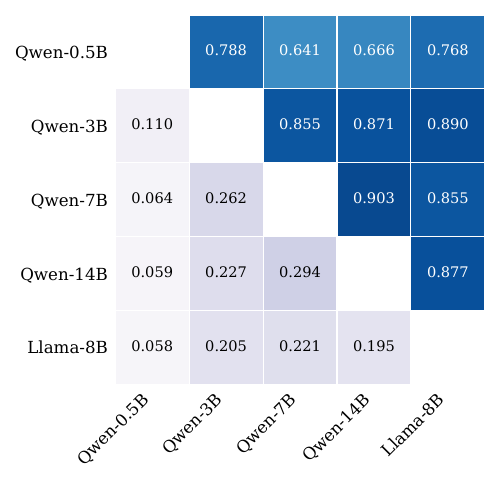}
\caption{Mechanistic fidelity decomposition on BoolQ over all segments.
Upper triangle: $F_{\mathrm{mag}}$ (Pearson $r^2$ of perturbation magnitudes
$\|\Delta z\|$). Lower triangle: $F_{\mathrm{align}}$ (Pearson $r^2$ of readout
alignments $\gamma$).}
\label{fig:mag-align}
\end{figure}

\paragraph{Magnitude and alignment between open models.}
As shown in \cref{eq:factorize_attribution}, the attribution
$\Delta\ell = \|\Delta z\|\,\|v\|\cos(\Delta z, v)$ decomposes into a
perturbation magnitude and an alignment term. The mechanistic metrics
$F_{\mathrm{mag}}$ and $F_{\mathrm{align}}$ (Equations \ref{eq:f_mag} and \ref{eq:f_align}) measure
cross-model agreement on each factor separately.
Figure~\ref{fig:mag-align} shows a $5 \times 5$ split heatmap of
$F_{\mathrm{mag}}$ and $F_{\mathrm{align}}$ across the five open models on
BoolQ.
Over all segments, agreement on magnitude is high (median .843) and agreement
on alignment is low (.201). Most of the former is segment role: a shuffle among
segments with the same role and position leaves .681
(Section~\ref{app:rob-noise}), because deleting a system-prompt segment moves
every model's residual stream a great deal. On passage sentences
$F_{\mathrm{mag}}$ is .599, with a floor of .046, and
$F_{\mathrm{align}}$ is .137. Agreement on how much a passage sentence
moves the representation is therefore real but more moderate than the
all-segment value suggests. For the same pairs $F_{\mathrm{attr}}$ is .394 on
passage sentences, well above $F_{\mathrm{align}}$: two models' attributions
agree better than their alignment terms do, so low agreement on alignment does
not by itself limit attribution fidelity.

\paragraph{Word level.}
Attribution fidelity is much lower for single-word deletions: the all-pairs
median is .168, against .451 at the sentence level. The drop is not measurement
noise. It is as large among open--open pairs, which are deterministic (.165
against .399), as among open--hosted pairs (.130 against .404).

\subsection{Unsigned attribution}
\label{app:rob-unsigned}

Attention scores are non-negative, whereas $\Delta\ell$ is signed, so a low
correlation between them cannot rule out that attention tracks how much a
segment matters. The last three rows of Table~\ref{tab:rob-grid} repeat the
cross-level comparison against $|\Delta\ell|$.

Attention carries almost no information about the direction of a segment's
effect (at most .005 over all segments and .03 on passage sentences) and a
modest amount about its size. Over all segments, mean attention
explains .20 of $|\Delta\ell|$, but part of this is segment role: the question
segment has both the highest attention and the largest $|\Delta\ell|$. On
passage sentences, mean and max attention explain .12--.14, and .09--.11 after
length control; rollout explains .02. For comparison, on the same sentences the
surrogate's perturbation magnitude explains .425, and .398 after length
control.

\subsection{Fidelity by pair type}
\label{app:rob-pairtype}

Table~\ref{tab:rob-pairtype} splits the medians of \cref{tab:cross_benchmark_results} by whether the two
models share a family and whether the pair is open--closed.
Families are Qwen (four open models), Llama (Llama-3.1-8B open; three hosted),
GPT (two) and Gemini (one).

\begin{table}[ht]
\caption{\Cref{tab:cross_benchmark_results} by pair type (median over pairs; Pearson $r^2$ for BoolQ and
WinoGrande, centered RV for ANLI and RACE). Only three same-family open--closed
pairs exist, all between Llama-3.1-8B and a hosted Llama.}
\label{tab:rob-pairtype}
\centering
\small
\resizebox{\linewidth}{!}{\begin{tabular}{llcccccccc}
\toprule
Metric & Pair type & Pairs & BoolQ & ANLI R1 & ANLI R2 & ANLI R3 & WinoGrande & BoolQ word & RACE RV \\
\midrule
$F_{\mathrm{pred}}$ & All & 55 & .718 & .531 & .443 & .520 & .224 & .719 & .791 \\
 & Same family & 13 & .710 & .477 & .429 & .492 & .150 & .710 & .780 \\
 & Cross family & 42 & .726 & .534 & .456 & .530 & .237 & .732 & .795 \\
 & Open--closed & 30 & .669 & .457 & .410 & .475 & .198 & .672 & .751 \\
 & \quad same family & 3 & .710 & .476 & .429 & .492 & .039 & .710 & .752 \\
 & \quad cross family & 27 & .656 & .445 & .367 & .443 & .203 & .665 & .751 \\
\midrule
$F_{\mathrm{attr}}$ & All & 55 & .451 & .308 & .265 & .291 & .152 & .168 & .334 \\
 & Same family & 13 & .489 & .371 & .304 & .326 & .077 & .258 & .394 \\
 & Cross family & 42 & .439 & .307 & .247 & .259 & .167 & .161 & .320 \\
 & Open--closed & 30 & .404 & .239 & .203 & .223 & .128 & .130 & .314 \\
 & \quad same family & 3 & .450 & .276 & .265 & .291 & .025 & .280 & .499 \\
 & \quad cross family & 27 & .393 & .236 & .189 & .207 & .142 & .128 & .308 \\
\bottomrule
\end{tabular}
}
\end{table}

On BoolQ, $F_{\mathrm{pred}}$ is similar within and across families (.710 and
.726). $F_{\mathrm{attr}}$ is higher within a family (.489 against .439), as on
every benchmark except WinoGrande. Open--closed pairs fall below the all-pairs
median on every benchmark for both metrics; on ANLI and RACE part of that gap
may reflect imputed label scores (Section~\ref{app:rob-missing}).

\FloatBarrier
\section{Case study: where does an open model explain a closed one?}
\label{app:case-study}

The fidelity metrics reported in \cref{tab:f-summary} aggregate over all prompts in a
benchmark. However, prompts vary widely in difficulty: on BoolQ, model
log-odds range from $|\ell| < 0.5$ (near-chance) to $|\ell| > 30$
(near-certain).
Here we show that an
open model's own attributions indicate on which BoolQ prompts it agrees with a
closed model, and that they do so better than either model's confidence. All analyses reuse the published results, use every prompt, and
fit no parameter.

\paragraph{What is measured.}
Let $s_i = \Delta\ell_i \cdot \mathrm{sign}(\ell)$ be the support of passage
sentence $i$ for a model's own answer, and call $\max_i s_i$ the model's
\emph{support} on that prompt: how strongly its answer rests on one sentence.
Support is computed from the open surrogate alone. We compare the surrogate
with a target on passage sentences only, by the pooled $r^2$ of $\Delta\ell$
and by how often the two have the same top sentence by $|\Delta\ell|$. Chance
is $1/N$ for $N$ passage sentences. \emph{Word overlap} is a baseline that uses
no model and picks the sentence sharing the most content words with the
question.

\begin{figure}[ht]
\centering
\includegraphics[width=\linewidth]{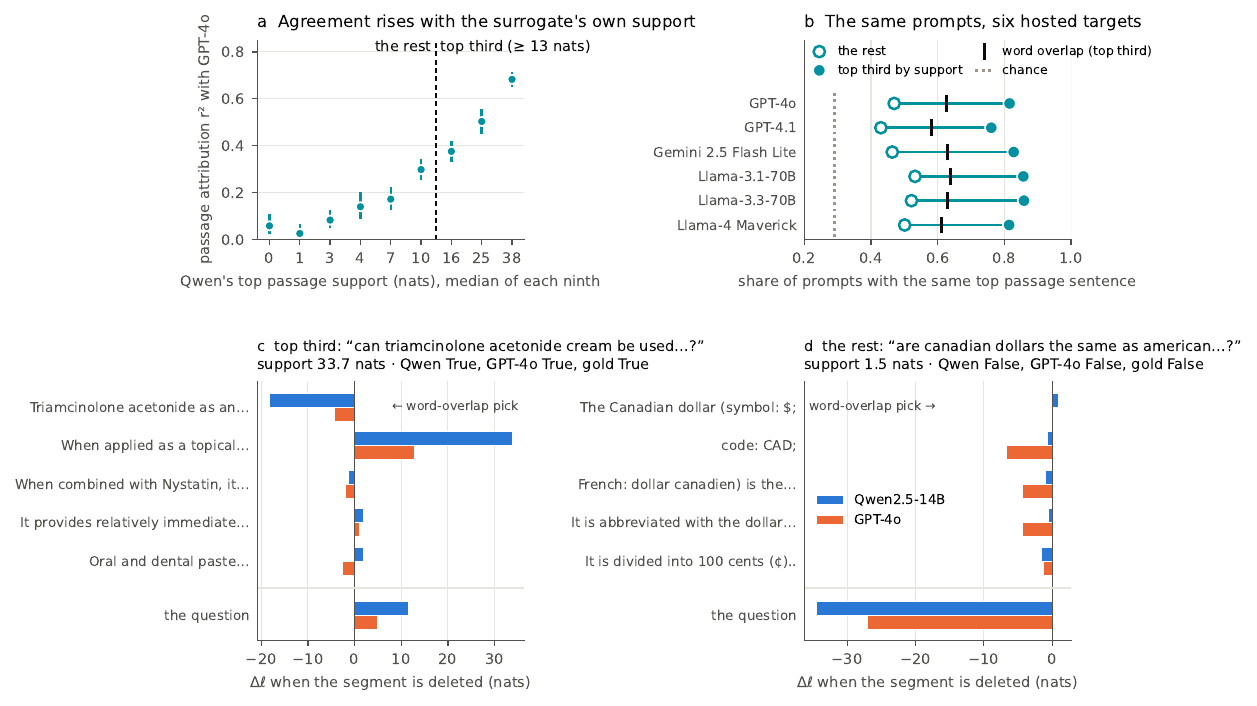}
\caption{(a) Passage $r^2$ between Qwen2.5-14B and GPT-4o by ninth of Qwen's
own support, with 95\% bootstrap intervals; the dashed line marks the top
third. (b) Share of prompts on which Qwen and each hosted target have the same
top passage sentence, for the top third by support (dots) and the rest
(rings), with word overlap on the top third (ticks) and chance (dotted). (c) A
prompt in the top third: both models rest on a sentence that word overlap
misses. (d) A prompt outside it, selected by rule: both answers are
carried by the question, and no passage sentence is decisive for Qwen.}
\label{fig:cs-screen}
\end{figure}

\paragraph{Agreement rises with the surrogate's support.}
For Qwen2.5-14B and GPT-4o, passage $r^2$ rises from below .1 in the lowest
three ninths of Qwen's support to .68 in the highest, and the share of prompts
with the same top sentence from .41 to .93
(Figure~\ref{fig:cs-screen}a). Splitting at the top third of support summarizes
this: Qwen names GPT-4o's top passage sentence on 82\% of those prompts and
47\% of the rest, with passage $r^2$ of .57 and .14. Word overlap reaches 63\%
on the top third, so part of the agreement there is lexical. The surrogate adds
graded, signed attributions that a lexical baseline does not provide: after
centering within prompts, its $|\Delta\ell|$ explains .61 of GPT-4o's, against
.26 for the overlap count.

\paragraph{It holds across targets and for a second surrogate.}
Because support does not involve the target, the same prompts can be scored
against every hosted model (Table~\ref{tab:cs-transfer},
Figure~\ref{fig:cs-screen}b). For Qwen2.5-14B the top-sentence match is
76--86\% on the top third against 43--53\% on the rest. For Llama-3.1-8B, a
smaller and less accurate model from another family, it is 70--84\% against
42--48\%. On a surrogate's top third it matches a hosted target about as often
as a second hosted model does: pairs of hosted models agree on the top
sentence 78\% of the time on Qwen's top third (median over 15 pairs) and 49\%
on the rest. The two surrogates' top thirds overlap only partly (Jaccard .46);
where both flag a prompt they choose the same sentence 90\% of the time.

\begin{table}[ht]
\caption{Agreement between an open surrogate and six hosted targets on the top
third of prompts by the surrogate's support and on the rest (all scored BoolQ
prompts; nothing is fitted). Word overlap and chance are for the top third.}
\label{tab:cs-transfer}
\centering
\small
\begin{tabular}{llcccccc}
\toprule
 & & \multicolumn{2}{c}{Passage $r^2$} & \multicolumn{4}{c}{Same top passage sentence} \\
\cmidrule(lr){3-4}\cmidrule(lr){5-8}
Surrogate & Target & Top third & Rest & Top third & Rest & Word overlap & Chance \\
\midrule
Qwen-14B & GPT-4o & 0.57 & 0.14 & 0.82 & 0.47 & 0.63 & 0.29 \\
 & GPT-4.1 & 0.45 & 0.11 & 0.76 & 0.43 & 0.58 & 0.29 \\
 & Gemini 2.5 Flash Lite & 0.60 & 0.15 & 0.83 & 0.46 & 0.63 & 0.29 \\
 & Llama-3.1-70B & 0.61 & 0.21 & 0.86 & 0.53 & 0.64 & 0.29 \\
 & Llama-3.3-70B & 0.63 & 0.22 & 0.86 & 0.52 & 0.63 & 0.29 \\
 & Llama-4 Maverick & 0.59 & 0.17 & 0.81 & 0.50 & 0.61 & 0.29 \\
\midrule
Llama-3.1-8B & GPT-4o & 0.49 & 0.11 & 0.79 & 0.42 & 0.60 & 0.29 \\
 & GPT-4.1 & 0.40 & 0.08 & 0.70 & 0.42 & 0.55 & 0.28 \\
 & Gemini 2.5 Flash Lite & 0.57 & 0.13 & 0.81 & 0.43 & 0.63 & 0.28 \\
 & Llama-3.1-70B & 0.57 & 0.16 & 0.82 & 0.48 & 0.62 & 0.28 \\
 & Llama-3.3-70B & 0.59 & 0.16 & 0.84 & 0.47 & 0.62 & 0.28 \\
 & Llama-4 Maverick & 0.58 & 0.14 & 0.81 & 0.45 & 0.62 & 0.28 \\
\bottomrule
\end{tabular}

\end{table}

\begin{figure}[ht]
\centering
\includegraphics[width=0.6\linewidth]{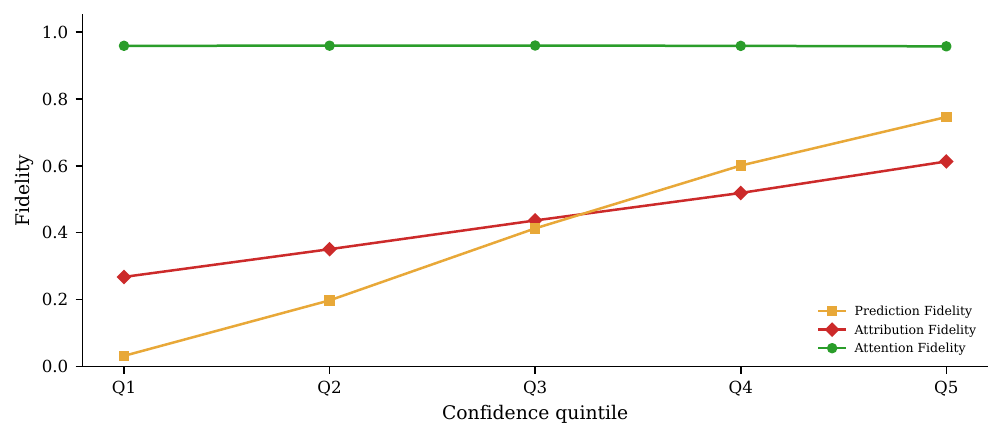}
\caption{$F_{\mathrm{pred}}$, $F_{\mathrm{attr}}$, and $F_{\mathrm{attn}}^{\mathrm{rollout}}$
versus model confidence quintile on BoolQ, averaged over all six Qwen~2.5
instruct pairs. Prediction fidelity rises sharply from Q1 to Q5
($0.03 \to 0.75$). Unlike the pooled $r^2$ used elsewhere in the paper, the
attribution and attention curves are means of within-prompt Pearson $r^2$:
attribution fidelity more than doubles ($0.27 \to 0.61$; pooled, $0.11 \to
0.54$) and attention fidelity is flat ($0.958 \pm 0.001$; pooled,
$0.88 \to 0.92$). The prediction curve pools both gold labels and is subject
to Section~\ref{app:rob-labels}.}
\label{fig:cs-confidence}
\end{figure}

\paragraph{Support is not confidence.}
The obvious alternative screen is confidence, $|\ell|$.
We bin BoolQ prompts into five quintiles by mean $|\ell|$ across the four Qwen
instruct models and compute $F_{\mathrm{pred}}$, $F_{\mathrm{attr}}$, and
$F_{\mathrm{attn}}$ within each bin. Figure~\ref{fig:cs-confidence} shows the
result.
Among these same-family pairs, attribution agreement rises with confidence and
attention agreement does not.
A practitioner may instead want to condition on a single model's
confidence---for instance, the target's (to assess when a surrogate's
explanations are trustworthy for a given target) or the surrogate's (to decide
whether a particular surrogate prediction merits further analysis).
Table~\ref{tab:cs-confidence} compares both with support for closed targets,
each ranking split at its top third.

\begin{table}[ht]
\caption{Three ways of ranking BoolQ prompts, each split at its top third:
agreement between an open surrogate and the hosted targets on passage
sentences (median over six targets). The last column counts the targets for
which the top third has the higher top-sentence match.}
\label{tab:cs-confidence}
\centering
\small
\begin{tabular}{llccccc}
\toprule
 & & \multicolumn{2}{c}{Passage $r^2$} & \multicolumn{2}{c}{Same top sentence} & Targets where \\
\cmidrule(lr){3-4}\cmidrule(lr){5-6}
Surrogate & Prompts ranked by & Top third & Rest & Top third & Rest & top third is higher \\
\midrule
Qwen-14B & Surrogate's support & 0.59 & 0.16 & 0.82 & 0.49 & 6/6 \\
 & Surrogate's confidence $|\ell|$ & 0.52 & 0.43 & 0.57 & 0.61 & 0/6 \\
 & Target's confidence $|\ell|$ & 0.59 & 0.38 & 0.66 & 0.58 & 5/6 \\
\midrule
Llama-3.1-8B & Surrogate's support & 0.57 & 0.14 & 0.81 & 0.44 & 6/6 \\
 & Surrogate's confidence $|\ell|$ & 0.46 & 0.39 & 0.55 & 0.58 & 0/6 \\
 & Target's confidence $|\ell|$ & 0.58 & 0.30 & 0.63 & 0.55 & 5/6 \\
\bottomrule
\end{tabular}

\end{table}

Confidence is related to how strongly the two models' attributions correlate,
but hardly to whether they rest on the same sentence. Ranking by the target's
confidence raises passage $r^2$ on the top third as far as support does (.59
for both, for Qwen), but it leaves much more agreement in the rest (.38
against .16), and it moves the top-sentence match only from .58 to .66.
Ranking by the surrogate's own confidence raises passage $r^2$ modestly (.52
against .43) and does not raise the top-sentence match for any of the twelve
surrogate--target pairs. Support and confidence order the prompts differently:
their rank correlation is $-.29$ for Qwen and $-.03$ for Llama, and their top
thirds overlap little (Jaccard .10 and .15). A model can be confident without its
answer resting on any one passage sentence, as in
Figure~\ref{fig:cs-screen}d.

\paragraph{Where it stops.}
The result concerns input saliency on one benchmark, not mechanism. It is also
mostly about stated answers the two models share. Between 82\% and 86\% of
top-third prompts are ones the surrogate answers True. Within True answers the
split is sharp (for Qwen, 80--90\% against 40--56\%); within False answers it
is weak (53--68\% against 44--52\%). A True answer in BoolQ is usually stated
in one sentence, whereas a False answer often rests on what the passage does
not say, which leave-one-out deletion cannot attribute to any sentence
(Figure~\ref{fig:cs-screen}d).\footnote{The prompt in panel (d) was chosen by rule, not by hand: among
prompts with low surrogate support that both models answer correctly, with
three to five passage sentences and different top sentences for the two
models, it has the median support.} The surrogate does not explain a target's
errors unless it shares them. When a target answers wrongly and Qwen gives the
same wrong answer, the top-sentence match on Qwen's top third is 66--80\%; when
Qwen answers correctly it is 40--49\%, no better than on the rest.

\end{document}